\documentclass{article}

\usepackage[main,final]{neurips_2026}

\usepackage[utf8]{inputenc}
\usepackage[T1]{fontenc}
\usepackage{hyperref}
\usepackage{url}
\usepackage{booktabs}
\usepackage{amsfonts}
\usepackage{amsmath}
\usepackage{amsthm}

\newtheorem{lemma}{Lemma}
\newtheorem{corollary}{Corollary}
\newtheorem{definition}{Definition}
\newtheorem{proposition}{Proposition}

\newtheorem{remark}{Remark}
\usepackage{nicefrac}
\usepackage{microtype}
\usepackage{xcolor}
\usepackage{graphicx}
\usepackage{subcaption}
\usepackage{multirow}
\usepackage{colortbl}
\definecolor{dmcrow}{RGB}{232, 244, 248}
\definecolor{avgcol}{RGB}{252, 246, 226}
\usepackage{enumitem}
\usepackage{algorithm}
\usepackage{algpseudocode}
\usepackage{wrapfig}
\usepackage{placeins}
\usepackage{pgfplots}
\usepgfplotslibrary{fillbetween}
\usepgfplotslibrary{groupplots}
\pgfplotsset{compat=1.17}
\usetikzlibrary{arrows.meta, shapes.geometric}

\newcommand{\ourmeth}{\textup{\textsc{MergeTune}}}
\newcommand{\mergetune}{\textup{\textsc{MergeTune}}}

\newcommand{\CLIP}{CLIP}

\definecolor{tealincre}{HTML}{009688}
\definecolor{graydecli}{HTML}{999999}
\newcommand{\incre}[1]{\textcolor{positivedelta}{#1}}
\newcommand{\decli}[1]{\textcolor{negativedelta}{#1}}

\definecolor{dmcblue}{RGB}{31, 97, 170}
\definecolor{mergetunecolor}{RGB}{180, 60, 60}

\definecolor{bestresult}{RGB}{200, 230, 200}
\definecolor{secondresult}{RGB}{220, 235, 255}
\definecolor{positivedelta}{RGB}{34, 139, 34}
\definecolor{negativedelta}{RGB}{180, 0, 0}
\definecolor{tableheader}{RGB}{240, 240, 240}

\definecolor{speccolor}{RGB}{214, 90, 49}
\definecolor{gencolor}{RGB}{49, 130, 189}
\definecolor{corridorcolor}{RGB}{117, 107, 177}%
\definecolor{baselinecolor}{RGB}{150, 150, 150}%
\definecolor{shiftarrow}{RGB}{50, 160, 50}

\definecolor{unreachable}{RGB}{200, 200, 215}
\definecolor{frontier}{RGB}{80, 50, 140}
\definecolor{mtcolor}{RGB}{0, 128, 128}

\definecolor{sepcolor}{RGB}{128, 0, 128}

\title{Decoupled Mode Connectivity for Base-to-Novel Generalization in Vision-Language Models}

\author{%
  Imad Eddine Marouf$^{1}$\quad Khalid Oublal$^{2}$\quad Enzo Tartaglione$^{1,3}$\quad St\'ephane Lathuili\`ere$^{1,4}$\\
  \normalfont $^{1}$T\'el\'ecom Paris, Institut Polytechnique de Paris, Palaiseau, France\\
  \normalfont $^{2}$\'Ecole Polytechnique, Institut Polytechnique de Paris, Palaiseau, France\\
  \normalfont $^3$CNR-ISASI, Lecce, Italy\quad $^{4}$Inria, Univ. Grenoble Alpes, CNRS, LJK, France
}

\begin{document}

\maketitle

\begin{abstract}
Prompt learning adapts vision-language models such as CLIP by optimizing a short learnable prompt.
The cross-entropy objective pulls this prompt toward base-class specialization, whereas generalization to unseen classes benefits from staying close to CLIP's zero-shot feature space.
Because both objectives act on the same prompt, their gradients conflict at convergence, and in our experiments no single-prompt loss design clearly improves on the cosine baseline.
We propose \textbf{Decoupled Mode Connectivity~(DMC)}, which gives each objective its own prompt.
Starting from a base-trained prompt, DMC trains a specialization prompt and a generalization prompt, and merges their text features with a fixed coefficient at test time, at the inference cost of a single prompt.
A linear mode connectivity~(LMC) corridor in text-feature space keeps the classification loss low for every classifier interpolated between the two endpoints, and a \emph{Visual Anchor} KL term preserves CLIP's zero-shot class-similarity distribution of each image during specialization.
We further argue that such regularizers should satisfy \emph{class-permutation invariance}~(CPI), and show via Fano's inequality, in an idealized model, that violating CPI lower-bounds the drop in novel accuracy by a term proportional to the mutual information between the prompt and the base-class labels.
Across 11 datasets, DMC improves the average harmonic mean of base and novel accuracy over \mergetune{} for both CoOp and KgCoOp, and its interpolation curves pass beyond the single-prompt operating point, shifting the base--novel frontier rather than trading along it.
Code is available at \url{https://github.com/IemProg/DMC}.
\end{abstract}

\section{Introduction}
\label{sec:intro}

Vision-language models such as CLIP~\citep{radford2021learning} acquire broad zero-shot recognition through large-scale pretraining on image-text pairs, and prompt learning~\citep{zhou2022coop, zhou2022cocoop, yao2023visual, khattak2023maple, khattak2023self, yang2024mma} adapts these models to downstream tasks with very few additional parameters. This adaptation comes with a tension between two objectives: supervised training on base classes pushes the learned prompt toward base-class discriminative directions, while generalization to unseen (novel) classes benefits from keeping the prompt close to CLIP's pretrained feature space. Fine-tuning on base classes thus erodes part of CLIP's zero-shot ability to recognize the novel classes, which are absent from the fine-tuning data.

Existing methods address this tension with regularizers that anchor the prompt to the zero-shot model. Cosine penalties pull the learned text features toward CLIP's frozen text features~\citep{yao2023visual, wang2026mergetune}; self-distillation constraints preserve CLIP's predictions during fine-tuning~\citep{khattak2023self, roy2024coprompt}; and linear mode connectivity (LMC) paths enforce smooth transitions between fine-tuned and zero-shot solutions~\citep{wang2026mergetune}. These designs differ in form, but share a common structure: a single prompt optimized under a sum of cross-entropy and a regularization term. We argue that this shared structure, not the choice of regularizer, is the bottleneck. At a converged single prompt, the gradients of the two terms point in opposing directions on all eight dataset--baseline pairs we measure (four datasets, CoOp and KgCoOp; Figure~\ref{fig:gradient_conflict}), so the prompt sits at a compromise where reweighting the terms only moves along a trade-off. Consistently, none of seven single-prompt variants we test, including Fisher-weighted regularization and self-distillation, improves on the cosine baseline by clearly more than retuning its weight (Appendix~\ref{app:loss_modifications}).

We propose \textbf{Decoupled Mode Connectivity (DMC)}, which changes the parameterization instead of the loss. DMC builds on \mergetune{}~\citep{wang2026mergetune}, a continued-training stage that refines a fine-tuned prompt with cross-entropy, a cosine regularizer toward CLIP's zero-shot text features, and an LMC path back to the fine-tuned model. DMC replaces this single refined prompt with two: a specialization prompt $\mathbf{c}_{\mathrm{spec}}$ trained with cross-entropy and a generalization prompt $\mathbf{c}_{\mathrm{gen}}$ trained with the regularizer. Each can then move toward its own optimum, and an LMC corridor in CLIP's text-feature space makes every mixture of the two a good base-class classifier. At test time, DMC merges the two text features once with a fixed coefficient, so inference costs the same as with a single prompt. Unlike post-hoc interpolation such as WiSE-FT~\citep{wortsman2022robust} or training-free merging~\citep{yadav2023ties, yu2024language}, the two prompts are trained jointly so that the path between them has low loss; training-free merging lowers HM in our experiments (Section~\ref{sec:base_novel}). Figure~\ref{fig:overview} illustrates this geometry on Flowers102.

Freed from the cosine regularizer, the specialization prompt still needs a way to retain CLIP's knowledge. We argue that any regularizer used for this should be \emph{class-permutation invariant} (CPI): it should not depend on which classes happen to be in the base set, because it must transfer to classes it never sees. The cosine regularizer is CPI but would pull $\mathbf{c}_{\mathrm{spec}}$ back to the zero-shot features; in an idealized model a CPI violation lower-bounds the drop in novel accuracy. Following this principle, DMC regularizes $\mathbf{c}_{\mathrm{spec}}$ with a label-free \emph{Visual Anchor} (VA), a KL term that preserves CLIP's zero-shot class-similarity distribution of each image.

Our contributions are:
\begin{enumerate}[leftmargin=*, itemsep=2pt, label=\textbf{(\arabic*)}]
\item We introduce \textbf{Decoupled Mode Connectivity (DMC)}, which assigns the specialization and generalization objectives to two prompts coupled by an LMC corridor in CLIP's text-feature space, at the inference cost of a single prompt (Section~\ref{sec:method}).
\item We argue that regularizers should satisfy class-permutation invariance (CPI), show in an idealized model that a CPI violation lower-bounds the drop in novel accuracy (Proposition~\ref{prop:cpi_fano}), and use this principle to design the Visual Anchor.
\item Across 11 datasets, DMC improves the average harmonic mean (HM) of base and novel accuracy over \mergetune{} by $+0.78$ (CoOp) and $+0.33$ (KgCoOp), with gains on 9 and 8 of the 11 datasets, and its interpolation curves pass beyond the single-prompt operating point (Section~\ref{sec:exp}, Figure~\ref{fig:pareto}).
\end{enumerate}
The gains require the two prompts to stay distinct. On the adapter-based MMA~\citep{yang2024mma}, their text features nearly coincide ($\cos \approx 0.99$) and the regularizer and VA no longer help, which delimits the scope of DMC (Appendix~\ref{app:endpoint_collapse}).

\section{Related Work}
\label{sec:related}

\paragraph{Prompt learning for vision-language models.}
Prompt learning adapts frozen vision-language models by optimizing
continuous prompt vectors instead of discrete templates.
CoOp~\citep{zhou2022coop} introduced learnable text prompts for
CLIP, achieving strong few-shot performance on base classes at the
cost of novel-class generalization. CoCoOp~\citep{zhou2022cocoop}
addressed this by conditioning prompts on image features, improving
transferability but adding inference overhead.
MaPLe~\citep{khattak2023maple} extended prompting to both vision
and language branches, improving cross-modal alignment.
PromptSRC~\citep{khattak2023self} introduced self-regularization
constraints that preserve pretrained knowledge during adaptation.
KgCoOp~\citep{yao2023visual} added a cosine penalty between
learned and hand-crafted prompt features. More recently,
CoPrompt~\citep{roy2024coprompt} enforced consistency between
trainable and pretrained predictions. All these methods optimize a
single set of prompts under a combined loss, and face the trade-off
between base specialization and novel generalization that we target.

\paragraph{Regularization for base-to-novel generalization.}
The base-to-novel setting requires the fine-tuned model to
generalize to classes unseen during training. Prior approaches
regularize the learned representation toward the pretrained model:
PromptSRC~\citep{khattak2023self} constrains prompt, feature, and
logit spaces simultaneously. \mergetune{}~\citep{wang2026mergetune}
combines an isotropic cosine score with a linear mode connectivity
path to merge zero-shot and fine-tuned solutions. Fisher-based
approaches~\citep{kirkpatrick2017overcoming} weight regularization
by parameter importance but introduce class-specific bias, which
we show violates class-permutation invariance; a Fisher-weighted
cosine brings no gain beyond retuning the cosine weight (Appendix~\ref{app:cpi}). A common
limitation is that all single-prompt methods enforce competing
objectives on the same parameters. Multi-task gradient analysis
methods such as PCGrad~\citep{yu2020gradient} manage such conflicts
by modifying gradient updates, but still optimize a single shared prompt.
DMC instead assigns the two objectives to separate prompts.

\paragraph{Model merging and mode connectivity.}
\citet{garipov2018loss} and \citet{draxler2018essentially}
established that independently trained solutions can be connected
by low-loss paths in weight space. Practical methods built on this
insight: WiSE-FT~\citep{wortsman2022robust} interpolates between
zero-shot and fine-tuned weights; model soups~\citep{wortsman2022model} average multiple
fine-tuned checkpoints; task
arithmetic~\citep{ilharco2023editing} composes task-specific weight
vectors through addition and subtraction;
TIES-Merging~\citep{yadav2023ties} and
DARE~\citep{yu2024language} resolve parameter interference during
merging. These methods operate in weight space and merge models
after training.
By contrast, DMC merges in text-feature space and trains the LMC
corridor as an objective, so both endpoints are optimized jointly
for corridor quality.

\definecolor{figzs}{HTML}{D55E00}
\definecolor{figcoop}{HTML}{117733}
\definecolor{figgen}{HTML}{0072B2}
\definecolor{figspec}{HTML}{CC79A7}
\definecolor{figpath}{HTML}{595959}
\definecolor{figbase}{HTML}{E69F00}
\definecolor{fignovel}{HTML}{CC79A7}

\begin{figure*}[t]
\centering
\includegraphics[width=\textwidth]{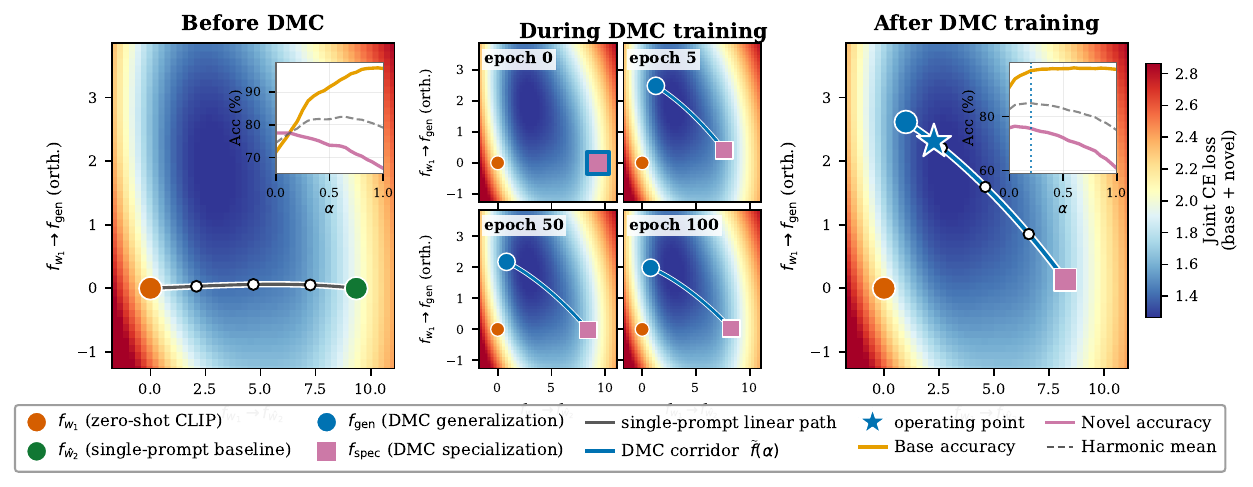}
\caption{\textbf{Decoupled Mode Connectivity on Flowers102}
(joint CE loss landscape). The 2D plane in CLIP
text-feature space is anchored at \textcolor{figzs}{\textbf{$f_{w_1}$}}
(zero-shot CLIP); the x-axis spans toward
\textcolor{figcoop}{\textbf{$f_{\hat{w}_2}$}} (fine-tuned CoOp checkpoint)
and the y-axis is the Gram--Schmidt orthogonal component of
\textcolor{figgen}{\textbf{$f_{\mathrm{gen}}$}} (text features of DMC's converged generalization
prompt) relative to $f_{w_1}$. Each grid point is re-normalized per class
before the joint base+novel CE loss is evaluated; construction details in
Appendix~\ref{app:hm_landscape}. \textbf{Before DMC} (left): the single-prompt
baseline traces the
\textcolor{figpath}{\textbf{linear path}}; the inset shows the base--novel
trade-off, with \textcolor{figbase}{\textbf{base}} and
\textcolor{fignovel}{\textbf{novel}} accuracies crossing over as $\alpha$
varies. \textbf{During DMC} (middle): the two prompts start collapsed at
$f_{\hat{w}_2}$ (epoch~0) and separate during training.
\textbf{After DMC} (right): the converged
\textcolor{figgen}{\textbf{DMC corridor}} lies inside the low-loss valley,
with the operating point at $\alpha = 0.20$ marked by the
\textcolor{figgen}{\textbf{star}}; the inset confirms base and novel
accuracies are both high and stable along the corridor.}
\label{fig:overview}
\end{figure*}

\section{Method}
\label{sec:method}

\subsection{Background and Problem Statement}
\label{sec:prelim}

\paragraph{CLIP prompt learning.} CLIP~\citep{radford2021learning} encodes images and text into a shared $\ell_2$-normalized embedding space via frozen visual and text encoders $V$ and $T$. An image $x$ is classified by computing cosine similarities between its visual feature $v = \ell_2(V(x))$ and the text features $\{f_k = \ell_2(T(p_k))\}_{k=1}^{C}$, where $p_k$ is a text prompt for class $k$.
Prompt learning~\citep{zhou2022coop} replaces hand-crafted templates with a learnable prompt $\mathbf{c} \in \mathbb{R}^{N \times D}$, shared across all classes, and optimizes it on base-class data while keeping both encoders frozen. Throughout, we write $\ell_2(\cdot)$ for row-wise $\ell_2$-normalization, $f_{w_1} = \ell_2(T(p_{\mathrm{hand}})) \in \mathbb{R}^{C \times D}$ for the zero-shot text features obtained from hand-crafted templates, and $\gamma$ for CLIP's learned logit scale.
A prompt $\mathbf{c}$ induces the text features $f(\mathbf{c}) = \ell_2(T(\mathbf{c}))$, whose row $f_k$ is the feature of class $k$; $f_{\mathrm{gen}}$ and $f_{\mathrm{spec}}$ denote the features of the two DMC prompts.

\paragraph{Single-prompt training and \mergetune{}.}
Let $\hat{w}_2$ denote the
prompt of a model (CoOp or KgCoOp) fine-tuned on base classes, with induced
features $f_{\hat{w}_2}=\ell_2(T(\hat{w}_2))$.
The generalization regularizer $\mathcal{R}$ anchors the learned
features to the zero-shot features $f_{w_1}$ by measuring their mean cosine
misalignment:
\begin{equation}
\textcolor{gencolor}{\mathcal{R}(f,\,f_{w_1})}
    = 1 - \frac{1}{C}\sum_{k=1}^{C}
      \cos\!\bigl(f_{k},\; f_{w_1,{k}}\bigr).
\label{eq:cosine_score}
\end{equation}
$\mathcal{R}$ is defined on any normalized text-feature matrix
$f \in \mathbb{R}^{C \times D}$; single-prompt training uses $f = f(\mathbf{c})$.
\mergetune{}~\citep{wang2026mergetune} continues training from $\hat{w}_2$. It learns a prompt $\mathbf{c}$ that combines $\mathcal{R}$ with cross-entropy and a linear mode connectivity~(LMC) term:
{%
\begin{equation}
\mathcal{L}_{\mergetune{}}(\mathbf{c})
    = \textcolor{speccolor}{\mathcal{L}_{\mathrm{CE}}(f(\mathbf{c}))}
    \;+\; \lambda\,\textcolor{gencolor}{\mathcal{R}(f(\mathbf{c}),\,f_{w_1})}
    \;+\; \beta\;\textcolor{corridorcolor}{\mathbb{E}_{\alpha}\,
      \mathcal{L}_{\mathrm{CE}}\!\left(
          \ell_2\!\bigl(f_{\hat{w}_2} + \alpha\,(f(\mathbf{c}) - f_{\hat{w}_2})\bigr)
      \right)},
\label{eq:mergetune}
\end{equation}}%
where \textcolor{speccolor}{$\mathcal{L}_{\mathrm{CE}}$} is cross-entropy on base-class training data, \textcolor{gencolor}{$\mathcal{R}$} is the cosine regularizer from Eq.~\eqref{eq:cosine_score}, and the \textcolor{corridorcolor}{third term} enforces a low-loss LMC path from $f_{\hat{w}_2}$ to $f(\mathbf{c})$. The weight $\lambda$ balances specialization against generalization and $\beta$ weights the LMC path; DMC reuses these terms, with the corridor redefined between its two prompts, and writes the regularizer weight as $\lambda_{\mathrm{gen}}$.

\paragraph{One prompt, two objectives.}
Specializing for base classes moves the text features away from their
zero-shot positions, and $\mathcal{R}$ pulls them back. With one shared prompt
$\mathbf{c}$, both gradients act on the same parameters, and they conflict: at
the converged \mergetune{} prompt, the cosine $\rho$ between
$\nabla_{\mathbf{c}} \mathcal{L}_{\mathrm{CE}}$ and
$\nabla_{\mathbf{c}} \mathcal{R}$ is negative on all eight dataset--baseline
pairs we measure (Figure~\ref{fig:gradient_conflict}). Every stationary point
of $\mathcal{L}_{\mathrm{CE}} + \lambda\,\mathcal{R}$ satisfies
$\nabla_{\mathbf{c}} \mathcal{L}_{\mathrm{CE}} = -\lambda\,\nabla_{\mathbf{c}} \mathcal{R}$,
which gives the following lemma (proof in Appendix~\ref{app:proofs}).
\begin{lemma}[Pareto stationarity of single-prompt solutions]
\label{lem:pareto}
Let $\mathcal{L}_{\mathrm{CE}}$ and $\mathcal{R}(\cdot,\, f_{w_1})$ be continuously
differentiable in $\mathbf{c}$ (we write $\mathcal{L}_{\mathrm{CE}}(\mathbf{c})$ and $\mathcal{R}(\mathbf{c}, f_{w_1})$ for $\mathcal{L}_{\mathrm{CE}}(f(\mathbf{c}))$ and $\mathcal{R}(f(\mathbf{c}), f_{w_1})$), and let $\mathbf{c}^*$ be a stationary point of
$\mathcal{L}(\mathbf{c}) = \mathcal{L}_{\mathrm{CE}}(\mathbf{c}) + \lambda\,\mathcal{R}(\mathbf{c},\, f_{w_1})$
for some $\lambda > 0$.
Then $\mathbf{c}^*$ is Pareto-stationary for the bi-objective problem
$(\mathcal{L}_{\mathrm{CE}},\,\mathcal{R})$: no direction in $\mathbb{R}^{N \times D}$
simultaneously decreases both objectives to first order.
If additionally $\nabla_{\mathbf{c}}\mathcal{L}_{\mathrm{CE}}(\mathbf{c}^*) \neq \mathbf{0}$
and $\nabla_{\mathbf{c}}\mathcal{R}(\mathbf{c}^*) \neq \mathbf{0}$,
then neither objective is individually minimized at $\mathbf{c}^*$.
\end{lemma}
Changing $\lambda$, or replacing $\mathcal{R}$ by another regularizer, therefore
only selects another compromise inside the same prompt. The lemma does not show
that two prompts do better; we show this empirically: none of seven
single-prompt variants improves on the baseline by clearly more than retuning
$\lambda$ (Appendix~\ref{app:loss_modifications}), while DMC does on CoOp and
KgCoOp (Section~\ref{sec:exp}).

A scalar model shows what decoupling changes
(Appendix~\ref{app:quadratic_decoupling}). With one parameter and two quadratic
objectives minimized at $a \neq b$, every weighting gives a point between $a$
and $b$ that minimizes neither objective. With one parameter per objective,
each reaches its own minimum, and interpolating between the two recovers every
single-parameter compromise, so the trade-off becomes a choice of an
interpolation coefficient at test time instead of a choice of $\lambda$ at
training time. In the high-dimensional prompt space, the interpolation path
between two prompts need not follow the single-prompt path, so it can also
reach points that no single prompt reaches; Figure~\ref{fig:pareto} measures
this.

\subsection{Decoupled Mode Connectivity}
\label{sec:dmc}

DMC replaces the continued training of one prompt in \mergetune{} with two
prompts, one per objective, both initialized from $\hat{w}_2$:
\begin{itemize}[leftmargin=*, itemsep=3pt]
    \item \textcolor{speccolor}{\textbf{Specialization prompt}
    $\mathbf{c}_{\mathrm{spec}}$}: trained by cross-entropy on base-class
    data, without the cosine pull toward $f_{w_1}$, and regularized only by
    the label-free Visual Anchor (below).
    \item \textcolor{gencolor}{\textbf{Generalization prompt}
    $\mathbf{c}_{\mathrm{gen}}$}: trained by $\mathcal{R}$ to stay close to
    $f_{w_1}$. It is learned rather than fixed to $f_{w_1}$, because the
    learned endpoint keeps discriminative structure that the zero-shot features
    lack (Section~\ref{sec:ablation}).
\end{itemize}
Neither prompt receives both endpoint losses, so each can move toward its
own optimum. For the adapter-based MMA~\citep{yang2024mma}, the two endpoints
are two copies of its adapters; only their text features are interpolated, and
images are encoded once with the specialization adapters.

\paragraph{Coupling the prompts: the LMC corridor.}
Two independently trained prompts give two unrelated classifiers. DMC
couples them with an LMC corridor that requires low classification loss at
every re-normalized convex combination of their text features. Because
classification scores are $v^{\top} f_k$ and $T$ is nonlinear, a linear path
between prompts is not a linear path between classifiers, so the corridor
operates in text-feature space:
{%
\begin{equation}
\widetilde{f}(\alpha)
    = \ell_2\!\Bigl(
        \textcolor{gencolor}{f_{\mathrm{gen}}}
        + \alpha\,\bigl(
            \textcolor{speccolor}{f_{\mathrm{spec}}}
            - \textcolor{gencolor}{f_{\mathrm{gen}}}
        \bigr)
      \Bigr), \qquad \alpha \in [0,1].
\label{eq:interp}
\end{equation}}%
The full DMC objective is
{%
\begin{equation}
\mathcal{L}_{\mathrm{DMC}}
    = \underbrace{
        \textcolor{speccolor}{
            \mathcal{L}_{\mathrm{CE}}(f_{\mathrm{spec}})
            + \lambda_{\mathrm{VA}}\,\mathcal{L}_{\mathrm{VA}}
        }
      }_{\textcolor{speccolor}{\mathbf{c}_{\mathrm{spec}}}\;\text{only}}
    \;+\;
      \underbrace{
        \textcolor{gencolor}{
            \lambda_{\mathrm{gen}}\,
            \mathcal{R}(f_{\mathrm{gen}},\,f_{w_1})
        }
      }_{\textcolor{gencolor}{\mathbf{c}_{\mathrm{gen}}}\;\text{only}}
    \;+\;
      \underbrace{
        \textcolor{corridorcolor}{
            \beta\;\mathbb{E}_{\alpha \sim \mathcal{U}[0,1]}\,
            \mathcal{L}_{\mathrm{CE}}\!\bigl(
                \widetilde{f}(\alpha)
            \bigr)
        }
      }_{\textcolor{corridorcolor}{\text{LMC corridor (both)}}},
\label{eq:dmc_loss}
\end{equation}}%
where $\mathcal{L}_{\mathrm{VA}}$ is defined in Eq.~\eqref{eq:va} below.
The corridor term ($\mathcal{L}_{\mathrm{LMC}}$ in Algorithm~\ref{alg:dmc}) is
the only coupling: it passes gradients, including cross-entropy gradients, to
both prompts, while the endpoint losses act on one prompt each. Sampling
$\alpha$ over the whole interval trains every point of the corridor to classify
base data well, so the classifier deployed at test time was itself optimized
during training. At inference, $\alpha$ is fixed to
$\alpha_{\mathrm{fix}} = 0.20$, selected on two development datasets
(Section~\ref{sec:alpha_select}); a small value keeps the classifier close to
$f_{\mathrm{gen}}$, which favors novel classes. The merged features
$\widetilde{f}(\alpha_{\mathrm{fix}})$ are computed once, so inference costs
the same as with a single prompt. Figure~\ref{fig:overview} shows this geometry
on Flowers102: $f_{\mathrm{gen}}$ and $f_{\mathrm{spec}}$ separate during
training, and the converged corridor lies inside a low-loss region of the
cross-entropy landscape that the path from zero-shot to the fine-tuned CoOp
checkpoint does not reach.

\paragraph{Regularizing $\mathbf{c}_{\mathrm{spec}}$: class-permutation invariance.}
\label{sec:cpi}
Without $\mathcal{R}$, cross-entropy alone tends to widen the angles
between classes to maximize the base-class margin, which discards the smaller
similarities that $f_{w_1}$ encodes between related classes. Which regularizer
should replace $\mathcal{R}$ on $\mathbf{c}_{\mathrm{spec}}$? It must transfer
to classes it never sees, so it should not depend on which classes are in the
base set. We formalize this as class-permutation invariance~(CPI).
\begin{definition}[Class-permutation invariance]
\label{def:cpi}
A regularizer $\mathcal{R}(f,\, f_{w_1})$ is \emph{class-permutation invariant} (CPI) if
\[
  \mathcal{R}(\Pi f,\;\Pi f_{w_1}) = \mathcal{R}(f,\; f_{w_1})
\]
for every $C \times C$ permutation matrix $\Pi$,
where $f \in \mathbb{R}^{C \times D}$ is the text-feature matrix with rows indexed by class.
\end{definition}
\begin{proposition}[CPI violation lower-bounds the novel-accuracy drop; informal]
\label{prop:cpi_fano}
Let $\mathbf{c}$ be the learned prompt of a training procedure that uses a
regularizer $\mathcal{R}$, and $\mathbf{c}^\star$ the learned prompt of the same
procedure with $\mathcal{R}$ replaced by a CPI regularizer. Under four idealized
assumptions, if $\mathcal{R}$ violates CPI, the novel-class accuracy of
$\mathbf{c}$ is lower than that of $\mathbf{c}^\star$ by at least
$I(\mathbf{c}; Y_{\mathrm{base}}) / \log|\mathcal{Y}_{\mathrm{novel}}| > 0$,
where $I(\mathbf{c}; Y_{\mathrm{base}})$ is the mutual information between the
prompt and a uniformly drawn base label.
\end{proposition}
Appendix~\ref{app:proof_cpi} gives the formal statement and proof (where
the prompt is written $\theta$), and Appendix~\ref{app:cpi_applicability}
discusses when the assumptions fail. Empirically, the non-CPI regularizers we
measure violate the symmetry by orders of magnitude more than the CPI ones, the
one we trained (Fisher-weighted cosine) gives no gain beyond retuning
$\lambda$, and CPI alone does not guarantee transfer (Appendix~\ref{app:cpi}).
The cosine regularizer weights all classes equally:
\begin{corollary}[Cosine regularizer satisfies CPI]
\label{cor:cpi_cosine}
The cosine regularizer $\mathcal{R}$ in Eq.~\eqref{eq:cosine_score} satisfies
Definition~\ref{def:cpi}. It uses no image labels, so $\mathcal{R}$ itself
adds no base-label information to the prompt it regularizes.
\end{corollary}

\paragraph{Visual Anchor.}
We regularize $\mathbf{c}_{\mathrm{spec}}$ with the \emph{Visual Anchor}~(VA),
a CPI regularizer that preserves \CLIP{}'s zero-shot class-similarity
distribution of each image without using its label. For each training image
$x$, let $v_a = \ell_2(V(\mathrm{aug}(x)))$ and
$v_b = \ell_2(V(\mathrm{aug}'(x)))$ denote frozen visual features of two
independently augmented views. VA penalizes the KL divergence between the
class-similarity distributions induced by \textcolor{gencolor}{$f_{w_1}$}
(zero-shot) and \textcolor{speccolor}{$f_{\mathrm{spec}}$}:
{%
\begin{equation}
\mathcal{L}_{\mathrm{VA}}
    = \mathbb{E}_{x}\;\mathrm{KL}\!\Bigl(
        \,\underbrace{
            \sigma\!\bigl(\gamma\,
            \textcolor{gencolor}{f_{w_1}}\, v_a \,/\, \tau\bigr)
        }_{\text{teacher (zero-shot, view $a$)}}
        \;\Big\|\;
        \underbrace{
            \sigma\!\bigl(\gamma\,
            \textcolor{speccolor}{f_{\mathrm{spec}}}\, v_b
            \,/\, \tau\bigr)
        }_{\text{student ($\mathbf{c}_{\mathrm{spec}}$, view $b$)}}
      \Bigr),
\label{eq:va}
\end{equation}}%
where $f_{w_1} v_a \in \mathbb{R}^C$ denotes the vector of inner products
$[f_{w_1,k}^\top v_a]_{k=1}^C$ (likewise for $f_{\mathrm{spec}} v_b$),
$\sigma(\cdot)$ is the softmax function, $\gamma$ puts both logit vectors
on CLIP's zero-shot scale, and $\tau$ is a distillation temperature that
softens both distributions.
Permuting the classes permutes both distributions in the same way and
leaves the KL unchanged, so VA satisfies CPI.
Proposition~\ref{prop:va_kl} (Appendix~\ref{app:proof_va}) relates this two-view
loss to the single-view case $v_a = v_b$. Algorithm~\ref{alg:dmc} summarizes the
full procedure.

\begin{algorithm}[t]
\caption{Decoupled Mode Connectivity (DMC)}%
\label{alg:dmc}
\begin{algorithmic}[1]
\Require Base-class data $\mathcal{D}_{\mathrm{base}}$,
    frozen CLIP encoders $V$, $T$,
    zero-shot features $f_{w_1}$,
    fine-tuned checkpoint $\hat{w}_2$,
    hyperparameters $\lambda_{\mathrm{gen}}, \lambda_{\mathrm{VA}}, \beta, \tau$,
    inference coefficient $\alpha_{\mathrm{fix}}$
    (we use $\lambda_{\mathrm{gen}}{=}8$, $\lambda_{\mathrm{VA}}{=}0.5$, $\beta{=}4$, $\tau{=}2$, $\alpha_{\mathrm{fix}}{=}0.20$)
\Ensure Test-time classifier

\Statex
\Statex \textbf{\textsc{Initialization}}
\State $\textcolor{speccolor}{\mathbf{c}_{\mathrm{spec}}} \gets \hat{w}_2$,
    \quad $\textcolor{gencolor}{\mathbf{c}_{\mathrm{gen}}} \gets \hat{w}_2$
    \Comment{Both initialized from the fine-tuned checkpoint}
\State Freeze $V$, $T$, $\hat{w}_2$;
    \quad mark only
    $\textcolor{speccolor}{\mathbf{c}_{\mathrm{spec}}}$,
    $\textcolor{gencolor}{\mathbf{c}_{\mathrm{gen}}}$ as learnable

\Statex
\Statex \textbf{\textsc{Training}}
\For{each mini-batch $(x, y) \sim \mathcal{D}_{\mathrm{base}}$}
    \State $f_{\mathrm{spec}} \gets \ell_2(T(\textcolor{speccolor}{\mathbf{c}_{\mathrm{spec}}}))$,
        \quad $f_{\mathrm{gen}} \gets \ell_2(T(\textcolor{gencolor}{\mathbf{c}_{\mathrm{gen}}}))$,
        \quad $v_a \gets \ell_2(V(\mathrm{aug}(x)))$, $v_b \gets \ell_2(V(\mathrm{aug}'(x)))$
    \State Compute $\textcolor{speccolor}{\mathcal{L}_{\mathrm{CE}}}$
        and $\textcolor{speccolor}{\mathcal{L}_{\mathrm{VA}}}$
        from $f_{\mathrm{spec}}$ (VA also uses $f_{w_1}$, $v_a$, $v_b$)
        \Comment{Eqs.~\eqref{eq:dmc_loss},~\eqref{eq:va}}
    \State Compute $\textcolor{gencolor}{\mathcal{R}}$
        from $f_{\mathrm{gen}}$
        \Comment{Eq.~\eqref{eq:cosine_score}}
    \State Sample $\alpha \sim \mathcal{U}[0,1]$; compute $\textcolor{corridorcolor}{\mathcal{L}_{\mathrm{LMC}}}$
        from $\widetilde{f}(\alpha)$
        via Eq.~\eqref{eq:interp}
        \Comment{Eq.~\eqref{eq:dmc_loss}, couples both}
    \State $\mathcal{L}_{\mathrm{DMC}} \gets
        \textcolor{speccolor}{\mathcal{L}_{\mathrm{CE}}
        + \lambda_{\mathrm{VA}}\, \mathcal{L}_{\mathrm{VA}}}
        + \textcolor{gencolor}{\lambda_{\mathrm{gen}}\, \mathcal{R}}
        + \textcolor{corridorcolor}{\beta\, \mathcal{L}_{\mathrm{LMC}}}$
        \Comment{Eq.~\eqref{eq:dmc_loss}}
    \State Update
        $\textcolor{speccolor}{\mathbf{c}_{\mathrm{spec}}}$,
        $\textcolor{gencolor}{\mathbf{c}_{\mathrm{gen}}}$
        via $\nabla \mathcal{L}_{\mathrm{DMC}}$
\EndFor

\Statex
\Statex \textbf{\textsc{Inference}} (test image $x$)
\State $f_{\mathrm{spec}} \gets \ell_2(T(\textcolor{speccolor}{\mathbf{c}_{\mathrm{spec}}}))$,
    \quad $f_{\mathrm{gen}} \gets \ell_2(T(\textcolor{gencolor}{\mathbf{c}_{\mathrm{gen}}}))$
    \Comment{Recompute from final trained prompts}
\State $\widetilde{f} \gets \ell_2\!\bigl(f_{\mathrm{gen}} + \alpha_{\mathrm{fix}}(f_{\mathrm{spec}} - f_{\mathrm{gen}})\bigr)$
    \Comment{Merge endpoints once; $\alpha_{\mathrm{fix}} = 0.20$}
\State \textbf{return} $\hat{y} = \arg\max_{k}\;
    \gamma \, \widetilde{f}_k^{\top}\, \ell_2(V(x))$
\end{algorithmic}
\end{algorithm}

\section{Experiments}
\label{sec:exp}

\subsection{Setup}

\paragraph{Datasets and protocol.}
We follow the base-to-novel generalization protocol
of~\citet{zhou2022coop}. Each dataset's classes are split into base
(seen) and novel (unseen) halves. Models are trained on 16-shot
base-class data and evaluated on both base and novel test sets. We
report Base accuracy, Novel accuracy, and their Harmonic Mean (HM), $\mathrm{HM} = 2\,\mathrm{Base}\cdot\mathrm{Novel}/(\mathrm{Base}+\mathrm{Novel})$, which is high only when both accuracies are high.
\mergetune{} and DMC rows are means over 3 seeds (DMC at $\alpha = 0.20$ and its best base-validation checkpoint; \mergetune{} evaluates its final prompt) (std in Table~\ref{tab:hm_meanstd_full}); prior-work rows are the published numbers.
We evaluate on 11 benchmarks: ImageNet, Caltech101, OxfordPets, StanfordCars, Flowers102, Food101, FGVCAircraft, SUN397, DTD, EuroSAT, and UCF101.

\paragraph{Baselines.}
We compare against prompt learning methods:
CoOp~\citep{zhou2022coop}, CoCoOp~\citep{zhou2022cocoop},
KgCoOp~\citep{yao2023visual}, MaPLe~\citep{khattak2023maple},
PromptSRC~\citep{khattak2023self}, MMA~\citep{yang2024mma};
methods using external knowledge:
CoPrompt~\citep{roy2024coprompt},
ATPrompt~\citep{li2025atprompt}; and model merging baselines:
TIES~\citep{yadav2023ties}, DARE~\citep{yu2024language}.
\mergetune{}~\citep{wang2026mergetune} is a continued fine-tuning stage applied post hoc to existing parameter-efficient fine-tuning (PEFT) methods. DMC replaces this stage: it starts from the same fine-tuned checkpoint and trains two prompts instead of one.

\paragraph{Implementation details.}
\label{sec:alpha_select}
All methods use a CLIP ViT-B/16 backbone. DMC trains two prompts initialized from a fine-tuned CoOp, KgCoOp, or MMA checkpoint, with $4 \times 512$ parameters per prompt on the prompt-tuning baselines (${\sim}2$K parameters per prompt). Training hyperparameters: $\lambda_{\text{gen}} = 8.0$, $\beta = 4.0$, $\lambda_{\mathrm{VA}} = 0.5$, $\tau = 2.0$; learning rate and epochs follow \mergetune{}'s defaults; batch size in Appendix~\ref{app:implementation}. The interpolation coefficient $\alpha$ is sampled uniformly from $[0, 1]$ at every training step.
At inference we use a single fixed $\alpha = 0.20$, selected on Flowers102 and Caltech101, which are also included in all reported averages. On the eight held-out datasets (ImageNet excluded), $\alpha = 0.20$ is on average $0.18$~HM below the per-dataset oracle (Appendices~\ref{app:implementation} and~\ref{app:alpha_selection}). The choice is not specific to this pair: over all 55 two-dataset development pairs, 47 select an $\alpha$ in $[0.15, 0.25]$, and 54 stay within 0.50~HM of the oracle on their held-out datasets (Table~\ref{tab:alpha_devset_reshuffle}). Only the interpolated text feature $\widetilde{f}(\alpha)$ is used, so DMC adds no per-image inference cost over a single-prompt method. Training adds only the second prompt (4{,}096 trainable parameters in total): one run on Flowers102 (100 epochs, batch size 128, as in the ablations; the main runs use 32) takes 15~min~34~s and 5.9~GB of peak memory on a single A100 (Appendix~\ref{app:implementation}). The $\beta$ sweep is in Appendix~\ref{app:beta_sweep}.

\subsection{Base-to-novel generalization}
\label{sec:base_novel}

We apply DMC to three fine-tuning baselines (CoOp, KgCoOp, MMA) and compare it with the \mergetune{} continued-training stage and with two training-free merging methods (TIES, DARE). Table~\ref{table:hm_main} reports HM on 11 datasets; Base and Novel accuracies are in Appendix~\ref{app:full_table}. In Table~\ref{table:base_to_novel} (best-of-3-seed rows, Appendix~\ref{app:full_table}), DMC raises Base and Novel together over \mergetune{} on 9 of 11 datasets for CoOp, so the CoOp HM gains mostly come from a Pareto shift. For KgCoOp this holds on only 4 of 11 datasets, and the gain comes mainly from Base. MMA also raises both on 8 of 11, but its two DMC endpoints collapse (see below). For MMA, both \mergetune{} and DMC remain below the published MMA; for KgCoOp, DMC roughly matches the published KgCoOp ($+0.05$~HM when both are averaged over per-dataset HMs), while \mergetune{} stays below it. Figure~\ref{fig:pareto} shows DMC's interpolation curves on four datasets.

\definecolor{figpDMC}{HTML}{0072B2}
\definecolor{figpNoVA}{HTML}{D55E00}
\definecolor{figpMT}{HTML}{595959}
\definecolor{figpArrow}{HTML}{117733}

\begin{figure*}[t]
\centering
\includegraphics[width=\textwidth]{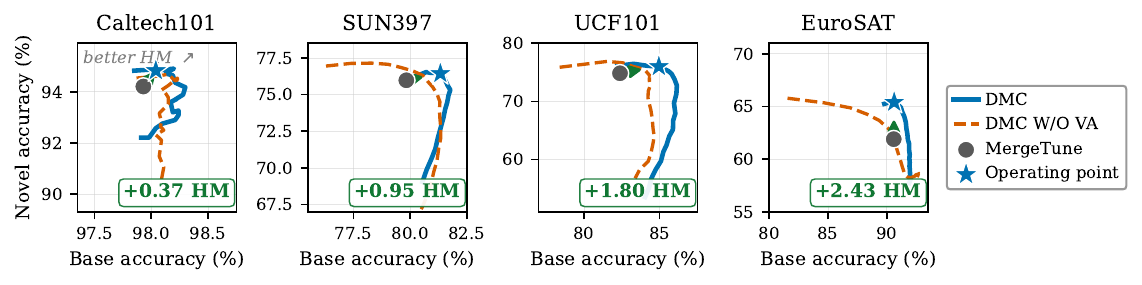}
\caption{\textbf{Empirical interpolation curves} on Caltech101, SUN397,
UCF101, and EuroSAT (CoOp base, 3-seed mean). Each curve traces the family of
classifiers from $f_{\mathrm{gen}}$~($\alpha{=}0$, near
zero-shot) to $f_{\mathrm{spec}}$~($\alpha{=}1$, fully
specialized). Solid: \textcolor{figpDMC}{\textbf{DMC}} (full
method); dashed: \textcolor{figpNoVA}{\textbf{DMC w/o VA}}.
\textcolor{figpMT}{\textbf{Gray dots}} mark the
\textcolor{figpMT}{\textbf{\ourmeth{} baseline}};
\textcolor{figpDMC}{\textbf{blue stars}} mark the operating point;
the \textcolor{figpArrow}{\textbf{green arrow}} shows the shift
from baseline to operating point. The boxed $\Delta\mathrm{HM}$
value reports the gap (operating point vs.\ baseline), read off the seed-averaged curves; it can differ from Table~\ref{table:hm_main} by up to $0.23$~HM.}%
\label{fig:pareto}
\end{figure*}

\textbf{Training-free merging.}
On average, TIES and DARE reduce HM relative to the fine-tuned baseline: CoOp loses $5.34$ and $1.07$~HM, KgCoOp $4.45$ and $1.84$, and MMA $10.46$ and $8.06$. Training-free merging therefore cannot replace the trained corridor.

\textbf{Comparison with \mergetune{}.}
In the same pipeline, DMC improves HM over \mergetune{} on 9 of 11 datasets for CoOp and 8 of 11 for KgCoOp. The largest gains are on EuroSAT ($+2.30$), FGVCAircraft ($+1.90$) and StanfordCars ($+1.83$) for CoOp. Every CoOp and KgCoOp regression is within seed variance. Averaged over the 11 datasets, the gain is $+0.78$~HM (CoOp), $+0.33$ (KgCoOp) and $+0.58$ (MMA).

\textbf{Dependence on the base method.}
The average gain is smallest on KgCoOp, whose built-in cosine regularization already anchors the prompt to zero-shot features, and largest on CoOp, which has no such anchor. This is a method-level pattern only: per dataset, the magnitude of the gradient conflict does not predict the gain (Pearson $r = -0.05$, Appendix~\ref{app:gamma_table}). StanfordCars, where zero-shot CLIP is weak, improves for all three baselines.

\textbf{When does DMC help?} DMC helps only if the two endpoints stay apart. If $f_{\mathrm{gen}} \approx f_{\mathrm{spec}}$, every $\widetilde{f}(\alpha)$ is the same classifier. In the runs of Table~\ref{table:hm_main}, $\cos(f_{\mathrm{gen}}, f_{\mathrm{spec}})$ averages $0.628$ (CoOp) and $0.704$ (KgCoOp) over 10 datasets (ImageNet not measured). On MMA, an adapter-based method, it averages $0.996$ and exceeds $0.99$ on 9 of 10 datasets (Table~\ref{tab:endpoint_collapse}). The two endpoints collapse, so the corridor has almost no effect on MMA. MMA+DMC still gains $0.58$~HM over MMA+\mergetune{}, but these gains do not track the remaining separation. Both stay below published MMA on every dataset, and the component ablation on MMA removes or reverses every gain of $\mathcal{R}$ and VA seen on CoOp (Appendix~\ref{app:endpoint_collapse}).

\textbf{Interpolation curves.}
DMC's interpolation curve passes northeast of the single-prompt operating point on every panel of Figure~\ref{fig:pareto}. The single-prompt \mergetune{} point does not reach these (Base, Novel) points, consistent with Lemma~\ref{lem:pareto}: rescaling $\lambda$ only moves a shared prompt along its Pareto-stationary set.

\begin{table*}[t]
\centering
\caption{Base-to-novel HM on 11 datasets. \mergetune{} and DMC: mean over 3 seeds ($\alpha=0.20$). Bold: better of \mergetune{} and DMC within each base method. Avg: mean of the 11 per-dataset HMs for our runs; for published, TIES and DARE rows, HM of the average Base and Novel, as in Table~\ref{table:base_to_novel}. $^\dag$: uses LLM or teacher-model knowledge.}
\label{table:hm_main}
\setlength{\tabcolsep}{4pt}
\renewcommand{\arraystretch}{1.05}
\resizebox{\linewidth}{!}{%
\begin{tabular}{l>{\columncolor{avgcol}}c|ccccccccccc}
\toprule
\multirow{2}[2]{*}{\textbf{Method}}
  & \multirow{2}[2]{*}{\textbf{Avg}}
  & \multicolumn{11}{c}{\textbf{HM per Dataset}} \\
\cmidrule(lr){3-13}
& & IN & Cal & Pets & Cars & Flo & Food & Air & SUN & DTD & Euro & UCF \\
\midrule
CLIP~\scriptsize{\textcolor{gray}{(ICML 21)}}
  & 71.70 & 70.22 & 95.40 & 94.12 & 68.65 & 74.83 & 90.66 & 31.09 & 72.23 & 56.37 & 60.03 & 73.85 \\
CoOp~\scriptsize{\textcolor{gray}{(IJCV 22)}}
  & 71.66 & 71.92 & 93.73 & 94.47 & 68.13 & 74.06 & 85.19 & 28.75 & 72.51 & 54.24 & 68.69 & 67.46 \\
KgCoOp~\scriptsize{\textcolor{gray}{(CVPR 23})}
  & 77.01 & 72.78 & 96.03 & 96.18 & 73.36 & 83.65 & 91.10 & 34.83 & 78.36 & 64.35 & 73.48 & 79.66 \\
MaPLe~\scriptsize{\textcolor{gray}{(CVPR 23)}}
  & 78.55 & 73.47 & 96.02 & 96.58 & 73.47 & 82.56 & 91.38 & 36.50 & 79.75 & 68.16 & 82.35 & 80.77 \\
PromptSRC~\scriptsize{\textcolor{gray}{(ICCV 23)}}
  & 79.97 & 74.01 & 96.02 & 96.30 & 76.58 & 85.95 & 91.10 & 40.15 & 80.52 & 71.75 & 82.32 & 82.74 \\
MMA~\scriptsize{\textcolor{gray}{(CVPR 24)}}
  & 79.87 & 74.02 & 96.15 & 96.72 & 75.70 & 85.48 & 90.71 & 38.33 & 80.38 & 73.38 & 83.87 & 82.20 \\
\midrule
CoPrompt$^\dag$~\scriptsize{\textcolor{gray}{(ICLR 24)}}
  & 80.48 & 74.33 & 96.55 & 96.87 & 75.66 & 85.71 & 91.40 & 39.76 & 81.30 & 72.79 & 85.84 & 83.07 \\
CoOp + ATPrompt$^\dag$~\scriptsize{\textcolor{gray}{(ICCV 25)}}
  & 74.65 & 73.33 & 95.74 & 95.67 & 71.58 & 79.77 & 88.09 & 32.52 & 74.24 & 58.22 & 71.96 & 73.45 \\
\midrule
CoOp + TIES~\scriptsize{\textcolor{gray}{(NeurIPS 23)}}
  & 66.32 & 65.60 & 92.50 & 92.56 & 60.56 & 69.27 & 88.02 & 29.54 & 62.70 & 48.40 & 58.40 & 58.30 \\
CoOp + DARE~\scriptsize{\textcolor{gray}{(ICML 24)}}
  & 70.59 & 68.93 & 93.67 & 94.84 & 64.57 & 74.92 & 87.50 & 31.03 & 66.47 & 57.04 & 67.39 & 66.28 \\
CoOp + \ourmeth{}~\scriptsize{\textcolor{gray}{(ICLR 26)}}
  & 76.45 & 72.89 & 96.27 & \textbf{96.13} & 73.38 & 83.92 & 91.05 & 33.33 & 77.85 & \textbf{64.26} & 73.45 & 78.42 \\
\rowcolor{dmcrow}
\textbf{CoOp + DMC (Ours)}
  & \textbf{77.23} & \textbf{72.90} & \textbf{96.41} & 96.04 & \textbf{75.21} & \textbf{84.47} & \textbf{91.13} & \textbf{35.23} & \textbf{78.75} & 63.53 & \textbf{75.75} & \textbf{80.07} \\
\midrule
KgCoOp + TIES~\scriptsize{\textcolor{gray}{(NeurIPS 23)}}
  & 72.56 & 70.93 & 95.83 & 95.23 & 69.73 & 73.39 & 90.91 & 29.86 & 73.46 & 56.93 & 67.02 & 74.18 \\
KgCoOp + DARE~\scriptsize{\textcolor{gray}{(ICML 24)}}
  & 75.17 & 71.78 & 95.97 & 94.73 & 70.33 & 79.33 & 91.02 & 31.85 & 76.12 & 62.01 & 72.19 & 78.71 \\
KgCoOp + \ourmeth{}~\scriptsize{\textcolor{gray}{(ICLR 26)}}
  & 76.43 & 72.82 & 95.94 & \textbf{96.28} & 74.90 & 83.39 & 91.08 & \textbf{34.10} & 78.03 & 62.66 & \textbf{73.44} & 78.08 \\
\rowcolor{dmcrow}
\textbf{KgCoOp + DMC (Ours)}
  & \textbf{76.76} & \textbf{73.09} & \textbf{96.11} & 96.23 & \textbf{75.38} & \textbf{83.67} & \textbf{91.19} & 33.17 & \textbf{78.85} & \textbf{64.34} & 73.38 & \textbf{78.90} \\
\midrule
MMA + TIES~\scriptsize{\textcolor{gray}{(NeurIPS 23)}}
  & 69.41 & 69.66 & 93.77 & 92.17 & 65.08 & 72.07 & 89.31 & 28.65 & 67.56 & 51.63 & 66.80 & 65.60 \\
MMA + DARE~\scriptsize{\textcolor{gray}{(ICML 24)}}
  & 71.81 & 70.27 & 93.70 & 92.96 & 68.01 & 73.92 & 88.72 & 29.25 & 72.02 & 57.74 & 70.17 & 70.78 \\
MMA + \ourmeth{}~\scriptsize{\textcolor{gray}{(ICLR 26)}}
  & 72.58 & \textbf{73.15} & \textbf{95.75} & 93.22 & 70.16 & 75.22 & 90.18 & 32.72 & 76.70 & 58.07 & 59.57 & 73.59 \\
\rowcolor{dmcrow}
\textbf{MMA + DMC (Ours)}
  & \textbf{73.16} & 72.90 & 95.74 & \textbf{93.38} & \textbf{70.51} & \textbf{76.37} & \textbf{90.33} & \textbf{33.10} & \textbf{77.12} & \textbf{59.62} & \textbf{60.57} & \textbf{75.11} \\
\bottomrule
\end{tabular}}
\end{table*}

\subsection{Ablation study}
\label{sec:ablation}

\paragraph{DMC component analysis.}
\begin{table*}[t]
\centering
\caption{Component ablation (HM, 3 seeds, $\beta = 4$ for all rows; batch size 128, whereas Table~\ref{table:hm_main} uses 32, so values differ from it). $\Delta$: change from the row above; bold: best per column. Rows without $\mathcal{R}$ omit VA.}
\label{tab:ablation_R}
\setlength{\tabcolsep}{3pt}
\renewcommand{\arraystretch}{1.08}
\resizebox{\linewidth}{!}{%
\begin{tabular}{l cccc cc cc cc cc cc}
\toprule
 & & & & & \multicolumn{2}{c}{\textbf{Flowers102}} & \multicolumn{2}{c}{\textbf{Caltech101}} & \multicolumn{2}{c}{\textbf{StanfordCars}} & \multicolumn{2}{c}{\textbf{EuroSAT}} & \multicolumn{2}{c}{\textbf{Average}} \\
\cmidrule(lr){6-7}\cmidrule(lr){8-9}\cmidrule(lr){10-11}\cmidrule(lr){12-13}\cmidrule(lr){14-15}
\textbf{Method} & \textbf{CE} & $\boldsymbol{\mathcal{R}}$ & \textbf{LMC} & \textbf{VA} & HM & $\Delta$ & HM & $\Delta$ & HM & $\Delta$ & HM & $\Delta$ & HM & $\Delta$ \\
\midrule
CoOp & \checkmark &  &  &  & 80.51{\scriptsize$\pm$1.54} & --- & 94.42{\scriptsize$\pm$0.60} & --- & 71.98{\scriptsize$\pm$0.70} & --- & 71.18{\scriptsize$\pm$5.75} & --- & 79.52 & --- \\
\mergetune{} w/o $\mathcal{R}$ & \checkmark &  & \checkmark &  & 78.19{\scriptsize$\pm$2.40} & \decli{$-2.32$} & 94.43{\scriptsize$\pm$1.89} & \incre{$+0.01$} & 71.04{\scriptsize$\pm$1.21} & \decli{$-0.94$} & 73.39{\scriptsize$\pm$4.93} & \incre{$+2.21$} & 79.26 & \decli{$-0.26$} \\
\mergetune{} & \checkmark & \checkmark & \checkmark &  & 82.86{\scriptsize$\pm$0.17} & \incre{$+4.67$} & 96.03{\scriptsize$\pm$0.21} & \incre{$+1.60$} & 73.60{\scriptsize$\pm$0.02} & \incre{$+2.56$} & \textbf{73.45}{\scriptsize$\pm$3.48} & \incre{$+0.06$} & 81.48 & \incre{$+2.22$} \\
DMC w/o $\mathcal{R}$ & \checkmark &  & \checkmark &  & 78.66{\scriptsize$\pm$1.43} & \decli{$-4.20$} & 94.60{\scriptsize$\pm$0.41} & \decli{$-1.43$} & 70.23{\scriptsize$\pm$1.89} & \decli{$-3.37$} & 68.79{\scriptsize$\pm$3.72} & \decli{$-4.66$} & 78.07 & \decli{$-3.42$} \\
DMC w/o VA & \checkmark & \checkmark & \checkmark &  & 83.91{\scriptsize$\pm$0.46} & \incre{$+5.25$} & \textbf{96.25}{\scriptsize$\pm$0.04} & \incre{$+1.65$} & 73.97{\scriptsize$\pm$0.44} & \incre{$+3.74$} & 73.24{\scriptsize$\pm$0.96} & \incre{$+4.45$} & 81.84 & \incre{$+3.77$} \\
\rowcolor{dmcrow}\textbf{DMC (Ours)} & \checkmark & \checkmark & \checkmark & \checkmark & \textbf{84.33}{\scriptsize$\pm$0.84} & \incre{$+0.42$} & 96.15{\scriptsize$\pm$0.38} & \decli{$-0.10$} & \textbf{74.28}{\scriptsize$\pm$0.52} & \incre{$+0.31$} & 73.05{\scriptsize$\pm$4.02} & \decli{$-0.19$} & \textbf{81.95} & \incre{$+0.11$} \\
\bottomrule
\end{tabular}}
\end{table*}
Table~\ref{tab:ablation_R} toggles one component per row on four datasets. The anchor $\mathcal{R}$ is what makes decoupling work: without it, both prompts collapse to the same base-specialized solution (Appendix~\ref{app:lambda_sweep}) and DMC falls below \mergetune{} on every dataset, so a second prompt alone does not explain the gain. With $\mathcal{R}$, DMC beats single-prompt \mergetune{} on three of four datasets ($+1.47$ on Flowers102); the EuroSAT gap ($-0.40$) is within its seed spread ($\pm 4$). VA is a small refinement, positive on two of four datasets. HM plateaus for $\lambda_{\text{gen}} \geq 4$ (Appendix~\ref{app:lambda_sweep}).

\paragraph{Capacity and view controls.}
DMC uses two 2K-parameter prompts; a single CoOp prompt with $N_{\mathrm{ctx}} = 16$ (8K parameters) still trails CoOp+DMC on all four datasets, by $1.07$ to $5.07$~HM (Table~\ref{tab:larger_single_prompt}). For VA, neither the second augmented view nor the augmentation policy matters much: all variants lie within about $1$~HM of each other, mostly within one seed standard deviation (Tables~\ref{tab:va_single_vs_two_view} and~\ref{tab:va_aug_ablation}).

\paragraph{Is the learnable generalization prompt necessary?}
As discussed in Section~\ref{sec:dmc}, $f_{\mathrm{gen}}$ retains
task-discriminative structure from the CoOp initialization while
recovering zero-shot geometry, making it a stronger interpolation
endpoint than the fixed zero-shot features $f_{w_1}$. We verify this empirically by
removing $\mathbf{c}_{\mathrm{gen}}$ and replacing the generalization endpoint with
the fixed $f_{w_1}$, while retraining $\mathbf{c}_{\mathrm{spec}}$ with the same loss
(CE + VA + LMC). The corridor is now anchored to $f_{w_1}$:
$\widetilde{f}(\alpha) = \ell_2\!\bigl(f_{w_1} + \alpha
(f_{\mathrm{spec}} - f_{w_1})\bigr)$.
Note that this configuration makes the LMC corridor structurally
identical to \mergetune{}'s: a path from a fixed reference to a
single learned prompt. Figure~\ref{fig:ablation_gen} compares the HM curves.
At $\alpha = 0.20$, the learned $f_{\mathrm{gen}}$ gives a higher HM on all four datasets. The gap is $+2.5$~HM on Flowers102, $+1.8$ on StanfordCars, $+7.4$ on EuroSAT, and smaller on Caltech101. It is largest on EuroSAT, where zero-shot CLIP is weakest among the four datasets (Table~\ref{table:hm_main}).

\definecolor{figaLearned}{HTML}{2196F3}
\definecolor{figaFixed}{HTML}{FF9800}

\begin{figure}[t]
\centering
\includegraphics[width=\linewidth]{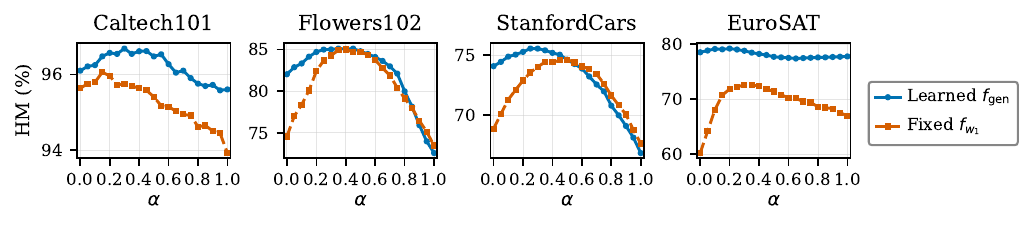}
\caption{\textbf{Ablation: learned $f_{\mathrm{gen}}$ vs.\ fixed $f_{w_1}$.}
Both variants retrain $\mathbf{c}_{\mathrm{spec}}$ with the same loss
(CE + VA + LMC), differing only in the generalization endpoint.
At small $\alpha$, including $\alpha = 0.20$, the \textcolor{figaLearned}{\textbf{learned $f_{\mathrm{gen}}$}} (solid)
gives a higher HM than the \textcolor{figaFixed}{\textbf{fixed $f_{w_1}$}} (dashed).}
\label{fig:ablation_gen}
\end{figure}

\paragraph{Gradient conflict at convergence.}
At the converged \mergetune{} checkpoint $\mathbf{c}^*$, we accumulate $\nabla_{\mathbf{c}}\mathcal{L}_{\mathrm{CE}}$ and $\nabla_{\mathbf{c}}\mathcal{R}$ over the full training set and report $\rho = \cos(\nabla_{\mathbf{c}}\mathcal{L}_{\mathrm{CE}},\, \nabla_{\mathbf{c}}\mathcal{R})$, following~\citet{yu2020gradient}; $\rho < 0$ means that the two steepest-descent directions are locally opposed. All eight values in Figure~\ref{fig:gradient_conflict} are negative (per-dataset values in Appendix~\ref{app:gamma_table}), with the strongest conflict on Flowers102 ($\rho \approx -0.95$). KgCoOp's conflict is weaker than CoOp's on three of four datasets, consistent with its built-in anchor, and the magnitude of $\rho$ does not predict DMC's gain. DMC removes the direct conflict: $\mathcal{R}$ acts only on $\mathbf{c}_{\mathrm{gen}}$, the endpoint cross-entropy only on $\mathbf{c}_{\mathrm{spec}}$, and the prompts interact only through the corridor.

\begin{figure}[t]
\centering
\includegraphics[width=\linewidth]{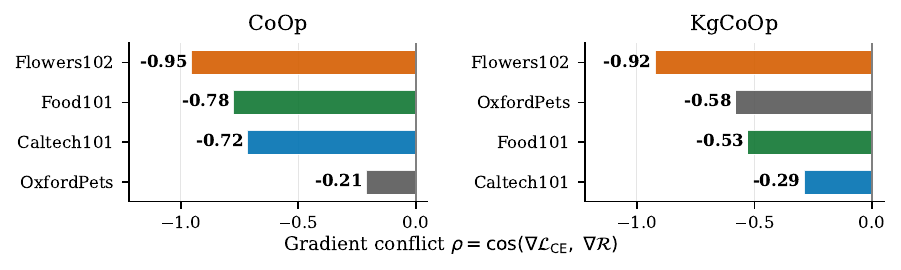}
\caption{\textbf{Gradient conflict at convergence.} $\rho = \cos(\nabla_{\mathbf{c}}\mathcal{L}_{\mathrm{CE}},\, \nabla_{\mathbf{c}}\mathcal{R})$ at the converged single-prompt checkpoint, averaged over the full training set. $\rho < 0$ on every dataset-baseline pair; Flowers102 shows the strongest conflict ($\rho \approx {-}0.95$ for CoOp).}
\label{fig:gradient_conflict}
\end{figure}

\begin{figure*}[t]
\centering
\includegraphics[width=0.95\textwidth]{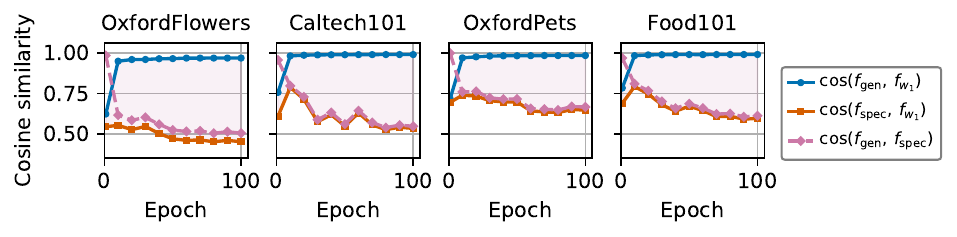}
\caption{\textbf{DMC decoupling dynamics during training}. Both prompts are initialized from $\hat{w}_2$.
\textcolor{gencolor}{\textbf{Blue}}: $\cos(f_{\mathrm{gen}},
f_{w_1})$ rises sharply within 10 epochs and saturates near
$0.97$--$0.99$, confirming rapid convergence toward zero-shot
geometry.
\textcolor{speccolor}{\textbf{Orange}}:
$\cos(f_{\mathrm{spec}}, f_{w_1})$ decreases overall under CE
supervision, with oscillations from the LMC coupling. Final values
range from ${\sim}0.45$ (Flowers102, strongest separation) to
${\sim}0.65$ (OxfordPets, weakest).
\textcolor{sepcolor}{\textbf{Purple dashed}}:
$\cos(f_{\mathrm{gen}},\,f_{\mathrm{spec}})$
reflects the feature separation, decreasing from ${\sim}1.0$ to
$0.50$--$0.67$. The shaded region visualizes the decoupling
gap. Single seed, with VA; Table~\ref{tab:endpoint_collapse} reports the 3-seed endpoint cosines (without VA; with-VA averages in its caption).}
\label{fig:dynamics}
\end{figure*}

\paragraph{Decoupling dynamics.}
Figure~\ref{fig:dynamics} shows the training dynamics of DMC on four
datasets (one run, seed~1). Both prompts start at $\hat{w}_2$. $f_{\mathrm{gen}}$
returns quickly to the zero-shot geometry ($\cos(f_{\mathrm{gen}}, f_{w_1})$
reaches $0.95$--$0.98$ within 10 epochs), so $\mathcal{R}$ dominates early
training, while $f_{\mathrm{spec}}$ moves away from $f_{w_1}$, with oscillations
from the LMC coupling. The final separation is largest on Flowers102 and
smallest on the near-ceiling OxfordPets.

\paragraph{Single-prompt loss modifications.}
Seven single-prompt loss modifications, evaluated on Caltech101 and Flowers102, produce no meaningful improvement over the cosine baseline: gains are at most $+0.40$~HM, similar to retuning $\lambda$, and never clear on both datasets.
For the same Flowers102 setup (seed~1, $\beta = 1$, best $\alpha$, matching the best-over-sweep protocol), DMC reaches $+0.86$~HM over the cosine baseline (Table~\ref{tab:wgen_sweep}), more than twice the best single-prompt gain on this dataset ($+0.36$). Full results are in Appendix~\ref{app:loss_modifications}.

\section{Conclusion}
\label{sec:conclusion}

Single-prompt methods for base-to-novel generalization settle at a Pareto-stationary compromise between the two objectives: the two objectives send opposing gradients to one shared prompt, and none of the single-prompt loss modifications we test escapes this constraint. DMC resolves this by assigning each objective to a dedicated prompt connected via an LMC corridor in text-feature space. A Visual Anchor KL term keeps CLIP's zero-shot class-similarity distribution of each image, and in an idealized information-theoretic model, violating class-permutation invariance lower-bounds the drop in novel accuracy. On CoOp and KgCoOp, DMC improves HM over \mergetune{} on most datasets. On MMA, the two endpoints converge to nearly the same features and DMC's regularizers no longer help.

\paragraph{Limitations.}
DMC adds one prompt (${\sim}$2K parameters) over the single-prompt baseline. The coefficient $\alpha$ is chosen offline and fixed at $0.20$ across datasets, since the HM-optimal $\alpha$ cannot be identified from base classes alone. On the eight datasets not used to select it (ImageNet excluded), the fixed value is on average $0.18$~HM below the per-dataset oracle (Appendix~\ref{app:alpha_selection}). DMC also needs the two prompts to stay distinct. On near-ceiling datasets they converge to similar features and the HM gain is small, and on MMA they collapse. Proposition~\ref{prop:cpi_fano} holds only under idealized assumptions (Appendix~\ref{app:cpi_applicability}).

\begin{ack}
This work benefited from Hi! PARIS and State funding managed by the French National Research Agency (ANR) under the France 2030 program, reference ANR-23-IACL-0005.
\end{ack}

\bibliographystyle{plainnat}
\bibliography{references}

\clearpage
\newpage
\appendix
\section*{Appendix Overview}
The appendix follows the order of the main argument: extended results first, then the evidence for the single-prompt ceiling, the ablations of DMC, its scope, the regularizer theory, the design choices, and finally the proofs. Unless stated otherwise, numbers for the rows we trained are means over seeds 1--3 (DMC at the fixed $\alpha = 0.20$, \mergetune{} at its final prompt), and prior-work numbers are as published.

\noindent\textbf{Extended results}
\begin{enumerate}[label=\Alph*.,leftmargin=2em,itemsep=1pt,topsep=2pt]
  \item \textbf{Full Base-to-Novel Results} (Appendix~\ref{app:full_table}). Base, Novel and HM for every method on all 11 datasets, per-cell seed variability, and an analysis of the weak cases.
  \item \textbf{Extended Interpolation Curves} (Appendix~\ref{app:pareto}). The $\alpha$-sweep curves on the datasets not shown in Figure~\ref{fig:pareto}, including ImageNet.
\end{enumerate}

\smallskip\noindent\textbf{Why a single prompt hits a ceiling}
\begin{enumerate}[label=\Alph*.,leftmargin=2em,itemsep=1pt,topsep=2pt,resume]
  \item \textbf{Quadratic Model of the Single-Prompt Ceiling} (Appendix~\ref{app:quadratic_decoupling}). A scalar model of the ceiling and Lemma~\ref{lem:pareto}.
  \item \textbf{Gradient Conflict per Dataset} (Appendix~\ref{app:gamma_table}). Measured gradient cosines at convergence for CoOp and KgCoOp.
  \item \textbf{Single-Prompt Loss Modifications} (Appendix~\ref{app:loss_modifications}). Seven alternative losses on Caltech101 and Flowers102.
\end{enumerate}

\smallskip\noindent\textbf{Ablations of DMC}
\begin{enumerate}[label=\Alph*.,leftmargin=2em,itemsep=1pt,topsep=2pt,resume]
  \item \textbf{Supporting Tables for the Ablations} (Appendix~\ref{app:ablation_tables}). The larger single-prompt control and the two Visual Anchor studies.
  \item \textbf{HM Landscape} (Appendix~\ref{app:hm_landscape}). Harmonic-mean landscapes on Flowers102 and Caltech101, complementing Figure~\ref{fig:overview}.
\end{enumerate}

\smallskip\noindent\textbf{Scope}
\begin{enumerate}[label=\Alph*.,leftmargin=2em,itemsep=1pt,topsep=2pt,resume]
  \item \textbf{When Does DMC Compose?} (Appendix~\ref{app:endpoint_collapse}). The endpoint-separation condition, the MMA analysis, and a falsification test.
\end{enumerate}

\smallskip\noindent\textbf{Regularizer theory}
\begin{enumerate}[label=\Alph*.,leftmargin=2em,itemsep=1pt,topsep=2pt,resume]
  \item \textbf{Class-Permutation Invariance} (Appendix~\ref{app:cpi}). Numerical CPI test, informal justification, the ZSDD case, and the applicability of the bound.
\end{enumerate}

\smallskip\noindent\textbf{Design choices and reproducibility}
\begin{enumerate}[label=\Alph*.,leftmargin=2em,itemsep=1pt,topsep=2pt,resume]
  \item \textbf{Training and Inference Protocol} (Appendix~\ref{app:implementation}).
  \item \textbf{Interpolation Coefficient Selection} (Appendix~\ref{app:alpha_selection}). Eight base-only selection criteria, the oracle $\alpha$ per dataset, and robustness of the fixed $\alpha=0.20$.
  \item \textbf{Hyperparameter Sensitivity} (Appendix~\ref{app:hyperparam}). Sweeps over $\beta$ and $\lambda_{\text{gen}}$.
\end{enumerate}

\smallskip\noindent\textbf{Proofs}
\begin{enumerate}[label=\Alph*.,leftmargin=2em,itemsep=1pt,topsep=2pt,resume]
  \item \textbf{Proofs} (Appendix~\ref{app:proofs}). Lemma~\ref{lem:pareto}, Proposition~\ref{prop:cpi_fano}, Corollary~\ref{cor:cpi_cosine}, Proposition~\ref{prop:va_kl}, and a summary of all assumptions.
\end{enumerate}

\section{Full Base-to-Novel Results}
\label{app:full_table}

This section asks whether DMC's HM gains come from Base, Novel or both, how stable they are across seeds, and where they fail.
Table~\ref{table:base_to_novel} reports Base, Novel, and HM for every method across all 11 datasets.
The main paper (Table~\ref{table:hm_main}) shows HM only; this table provides the full breakdown for readers who wish to verify the Base and Novel trends independently.
For the \mergetune{} and DMC rows it reports the \emph{best of three seeds}, so that each Base/Novel pair comes from a single run, as in the prior-work rows. These numbers are therefore higher than the 3-seed means of Table~\ref{table:hm_main} (e.g., CoOp+DMC average HM $79.04$ here versus $77.23$ there) and should not be compared with them; all headline numbers in the paper are the seed means.

\begin{table*}
    \caption{Base, Novel and HM on 11 datasets. For the \mergetune{} and DMC rows we report the \emph{best of 3 seeds}, so these numbers are higher than, and not directly comparable with, the 3-seed means of Table~\ref{table:hm_main}, which are the headline results; prior-work rows are published numbers. Parentheses: average-HM change relative to the base method.
    $^\dag$: Using large language model or teacher model's knowledge.}
    \begin{subtable}[t]{\textwidth}
    \centering
    \resizebox{0.95\linewidth}{!}
    {
    \centering
    \begin{tabular}{c>{\columncolor{avgcol}}c>{\columncolor{avgcol}}c>{\columncolor{avgcol}}c|ccc|ccc|ccc}
    \toprule
    \multirow{2}[3]{*}{Method} & \multicolumn{3}{>{\columncolor{avgcol}}c}{Average} & \multicolumn{3}{c}{ImageNet} & \multicolumn{3}{c}{ Caltech101} & \multicolumn{3}{c}{OxfordPets} \\
    \cmidrule(lr){2-4}\cmidrule(lr){5-7}\cmidrule(lr){8-10}\cmidrule(lr){11-13}
    & Base & Novel & HM & Base & Novel & HM & Base & Novel & HM & Base & Novel & HM \\
    \midrule
    CLIP~\scriptsize{\textcolor{gray}{(ICML 21)}} & 69.34 & 74.22 & 71.70 & 72.43 & 68.14 & 70.22 & 96.84 & 94.00 & 95.40 & 91.17 & 97.26 & 94.12 \\
    CoOp~\scriptsize{\textcolor{gray}{(IJCV 22)}} & 82.69 & 63.22 & 71.66 & 76.47 & 67.88 & 71.92 & 98.00 & 89.81 & 93.73 & 93.67 & 95.29 & 94.47 \\
    KgCoOp~\scriptsize{\textcolor{gray}{(CVPR 23)}} & 80.73 & 73.61 & 77.01 & 75.83 & 69.96 & 72.78 & 97.72 & 94.39 & 96.03 & 94.65 & 97.76 & 96.18 \\
    MaPLe~\scriptsize{\textcolor{gray}{(CVPR 23)}} & 82.28 & 75.14 & 78.55 & 76.66 & 70.54 & 73.47 & 97.74 & 94.36 & 96.02 & 95.43 & 97.76 & 96.58 \\
    PromptSRC~\scriptsize{\textcolor{gray}{(ICCV 23)}} & 84.26 & 76.10 & 79.97 & 77.60 & 70.73 & 74.01 & 98.10 & 94.03 & 96.02 & 95.33 & 97.30 & 96.30 \\

    MMA~\scriptsize{\textcolor{gray}{(CVPR 24)}} & 83.20 & 76.80 & 79.87 & 77.31 & 71.00 & 74.02 & 98.40 & 94.00 & 96.15 & 95.40 & 98.07 & 96.72 \\
    \midrule
    CoPrompt$^\dag$ ~\scriptsize{\textcolor{gray}{(ICLR 24)}} & 84.00 & 77.23 & 80.48 & 77.67 & 71.27 & 74.33 & 98.27 & 94.90 & 96.55 & 95.67 & 98.10 & 96.87 \\
    CoOp + ATPrompt$^\dag$ ~\scriptsize{\textcolor{gray}{(ICCV 25)}}& 82.68 & 68.04 & 74.65~\scriptsize{\incre{(+2.99)}}  & 76.27 & 70.60 & 73.33 & 97.95 & 93.63 & 95.74 & 94.77 & 96.59 & 95.67 \\

    \midrule
    CoOp + TIES~\scriptsize{\textcolor{gray}{(NeurIPS 23)}} & 66.95 & 65.71 & 66.32~\scriptsize{\decli{(-5.34)}} & 70.09 & 61.65& 65.60 & 93.63 & 91.34 & 92.50 & 90.12 & 95.13 & 92.56 \\
    CoOp + DARE~\scriptsize{\textcolor{gray}{(ICML 24)}} & 75.91 & 65.97 & 70.59~\scriptsize{\decli{(-1.07)}} & 74.25 &64.32&68.93&96.36&91.12&93.67&93.41&96.33&94.84 \\
    CoOp + \ourmeth{}~\scriptsize{\textcolor{gray}{(ICLR 26)}} & 80.82 & 73.97 & 77.24~\scriptsize{\incre{(+5.58)}} & 75.96 & 69.91 & 72.81 & 97.91 & 94.65 & 96.25 & 95.09 & 97.75 & 96.40 \\
    \rowcolor{dmcrow} \textbf{CoOp + DMC (Ours)} & 83.67 & 74.89 & 79.04~\scriptsize{\incre{(+7.38)}} & 76.13 & 69.93 & 72.90 & 98.32 & 95.41 & 96.84 & 95.37 & 98.32 & 96.82 \\
    \midrule
    KgCoOp + TIES~\scriptsize{\textcolor{gray}{(NeurIPS 23)}} &73.03 &72.09&72.56~\scriptsize{\decli{(-4.45)}}&74.06&68.06&70.93&97.27&94.43&95.83& 93.99 & 96.51 & 95.23 \\
    KgCoOp + DARE~\scriptsize{\textcolor{gray}{(ICML 24)}} &78.15&72.41&75.17~\scriptsize{\decli{(-1.84)}}&75.10&68.73&71.78&97.53&94.47&95.97&92.22&97.37&94.73 \\
    KgCoOp + \ourmeth{}~\scriptsize{\textcolor{gray}{(ICLR 26)}} &  81.85 & 74.46  &  77.98~\scriptsize{\incre{(+0.97)}} & 76.49 & 69.35 & 72.75 & 97.87 & 94.91 & 96.37 & 95.32 & 97.76 & 96.53 \\
    \rowcolor{dmcrow} \textbf{KgCoOp + DMC (Ours)} & 83.58 & 74.19 & 78.61~\scriptsize{\incre{(+1.60)}} & 76.32 & 70.12 & 73.09 & 98.19 & 94.98 & 96.56 & 95.43 & 97.82 & 96.61 \\
    \midrule
    MMA + TIES~\scriptsize{\textcolor{gray}{(NeurIPS 23)}} & 70.39 & 68.46 & 69.41 \scriptsize{\decli{(-10.46)}}& 73.16 & 66.48 & 69.66 &95.36 & 92.23&93.77 &  91.24 & 93.11  & 92.17  \\
    MMA + DARE~\scriptsize{\textcolor{gray}{(ICML 24)}} &74.25 & 69.52 & 71.81~\scriptsize{\decli{(-8.06)}}&74.17 &66.76 &70.27 & 94.95& 92.49 &93.70 & 90.52 &95.53 & 92.96 \\
    MMA + \ourmeth{}~\scriptsize{\textcolor{gray}{(ICLR 26)}} & 70.59 & 75.07 & 72.76~\scriptsize{\decli{(-7.11)}} & 73.68 & 69.02 & 71.27 & 97.48 & 94.43 & 95.93 & 90.01 & 96.92 & 93.34 \\
    \rowcolor{dmcrow} \textbf{MMA + DMC (Ours)} & 72.14 & 75.52 & 73.79~\scriptsize{\decli{(-6.08)}} & 76.41 & 71.06 & 73.64 & 97.48 & 94.43 & 95.93 & 90.06 & 97.20 & 93.49 \\
    \bottomrule
    \end{tabular}
    }
    \end{subtable}

    \vspace{5pt}
    \begin{subtable}[t]{\textwidth}
    \centering
    \resizebox{0.95\linewidth}{!}
    {
    \begin{tabular}{cccc|ccc|ccc|ccc}
    \toprule \multirow{2}[3]{*}{ Method } & \multicolumn{3}{c}{StanfordCars} & \multicolumn{3}{c}{ Flowers102} & \multicolumn{3}{c}{ Food101} & \multicolumn{3}{c}{ FGVCAircraft} \\
    \cmidrule(lr){2-4}\cmidrule(lr){5-7}\cmidrule(lr){8-10}\cmidrule(lr){11-13}
     & Base & Novel & HM & Base & Novel & HM & Base & Novel & HM & Base & Novel & HM \\
     \midrule
    CLIP~\scriptsize{\textcolor{gray}{(ICML 21)}} & 63.37 & 74.89 & 68.65 & 72.08 & 77.80 & 74.83 & 90.10 & 91.22 & 90.66 & 27.19 & 36.29 & 31.09 \\
    CoOp~\scriptsize{\textcolor{gray}{(IJCV 22)}} & 78.12 & 60.40 & 68.13 & 97.60 & 59.67 & 74.06 & 88.33 & 82.26 & 85.19 & 40.44 & 22.30 & 28.75 \\
    KgCoOp~\scriptsize{\textcolor{gray}{(CVPR 23)}} & 71.76 & 75.04 & 73.36 & 95.00 & 74.73 & 83.65 & 90.50 & 91.70 & 91.10 & 36.21 & 33.55 & 34.83 \\
    MaPLe~\scriptsize{\textcolor{gray}{(CVPR 23)}} & 72.94 & 74.00 & 73.47 & 95.92 & 72.46 & 82.56 & 90.71 & 92.05 & 91.38 & 37.44 & 35.61 & 36.50 \\
    PromptSRC~\scriptsize{\textcolor{gray}{(ICCV 23)}} & 78.27 & 74.97 & 76.58 & 98.07 & 76.50 & 85.95 & 90.67 & 91.53 & 91.10 & 42.73 & 37.87 & 40.15 \\
    MMA~\scriptsize{\textcolor{gray}{(CVPR 24)}} & 78.50 & 73.10 & 75.70 & 97.77 & 75.93 & 85.48 & 90.13 & 91.30 & 90.71 & 40.57 & 36.33 & 38.33 \\
    \midrule

    CoPrompt$^\dag$ ~\scriptsize{\textcolor{gray}{(ICLR 24)}} & 76.97 & 74.40 & 75.66 & 97.27 & 76.60 & 85.71 & 90.73 & 92.07 & 91.40 & 40.20 & 39.33 & 39.76 \\
    CoOp + ATPrompt$^\dag$ ~\scriptsize{\textcolor{gray}{(ICCV 25)}} & 77.43 & 66.55 & 71.58 & 97.44 & 67.52 & 79.77 & 88.74 & 87.44 & 88.09 & 40.38 & 27.22 & 32.52 \\

    \midrule
    CoOp + TIES~\scriptsize{\textcolor{gray}{(NeurIPS 23)}} & 57.94 & 63.42 & 60.56 & 73.60 & 65.41 & 69.27 &88.42&87.64&88.02&30.29&28.83&29.54 \\
    CoOp + DARE~\scriptsize{\textcolor{gray}{(ICML 24)}}  &67.82&61.62&64.57&90.88&63.73&74.92&87.78&87.22&87.50&34.41&28.25&31.03 \\
    CoOp + \ourmeth{}~\scriptsize{\textcolor{gray}{(ICLR 26)}} & 71.94 & 75.12 & 73.49& 95.32 & 74.30  & 83.51 & 90.56  &  91.75 &  91.15 & 36.17  &  34.35 & 35.24 \\
    \rowcolor{dmcrow} \textbf{CoOp + DMC (Ours)} & 79.19 & 74.70 & 76.88 & 97.91 & 76.38 & 85.82 & 90.84 & 92.01 & 91.42 & 40.94 & 35.03 & 37.76 \\
    \midrule
    KgCoOp + TIES~\scriptsize{\textcolor{gray}{(NeurIPS 23)}} &65.79&74.17&69.73&74.26&72.53&73.39&90.25&91.58&90.91&30.81&28.97&29.86\\
    KgCoOp + DARE~\scriptsize{\textcolor{gray}{(ICML 24)}} &67.84&73.02&70.33&87.02&72.88&79.33&90.44&91.60&91.02&33.49&30.37&31.85\\
    KgCoOp + \ourmeth{}~\scriptsize{\textcolor{gray}{(ICLR 26)}} & 73.21 & 75.14 & 74.17 & 96.49 & 74.68 & 84.20 & 90.51 & 91.83 & 91.17 & 37.09 & 35.11 & 36.07 \\
    \rowcolor{dmcrow} \textbf{KgCoOp + DMC (Ours)} & 78.66 & 74.75 & 76.66 & 98.01 & 74.61 & 84.72 & 90.86 & 91.89 & 91.37 & 39.44 & 32.21 & 35.46 \\
    \midrule
    MMA + TIES~\scriptsize{\textcolor{gray}{(NeurIPS 23)}} &60.72 & 70.12& 65.08 & 73.46 &70.74 & 72.07 &88.79 &89.83& 89.31& 29.27 & 28.06 & 28.65 \\
    MMA + DARE~\scriptsize{\textcolor{gray}{(ICML 24)}} & 64.82&71.52 & 68.01&77.15 & 70.95 &73.92  & 89.85 &87.61 & 88.72&30.35 &28.23 &29.25 \\
    MMA + \ourmeth{}~\scriptsize{\textcolor{gray}{(ICLR 26)}} & 65.74 & 75.61 & 70.33 & 73.03 & 78.23 & 75.55 & 89.67 & 90.77 & 90.22 & 31.57 & 34.13 & 32.80 \\
    \rowcolor{dmcrow} \textbf{MMA + DMC (Ours)} & 66.62 & 75.46 & 70.76 & 76.54 & 77.94 & 77.23 & 89.91 & 90.95 & 90.43 & 32.11 & 35.03 & 33.51 \\
    \bottomrule
    \end{tabular}
    }
    \end{subtable}

    \vspace{5pt}
    \begin{subtable}[t]{\textwidth}
    \centering
    \resizebox{0.95\linewidth}{!}
    {
    \begin{tabular}{cccc|ccc|ccc|ccc}
    \toprule
    \multirow{2}[3]{*}{Method} & \multicolumn{3}{c}{SUN397} & \multicolumn{3}{c}{DTD} & \multicolumn{3}{c}{EuroSAT} & \multicolumn{3}{c}{UCF101} \\
    \cmidrule(lr){2-4}\cmidrule(lr){5-7}\cmidrule(lr){8-10}\cmidrule(lr){11-13}
    & Base & Novel & HM & Base & Novel & HM & Base & Novel & HM & Base & Novel & HM \\
    \midrule
    CLIP~\scriptsize{\textcolor{gray}{(ICML 21)}} & 69.36 & 75.35 & 72.23 & 53.24 & 59.90 & 56.37 & 56.48 & 64.05 & 60.03 & 70.53 & 77.50 & 73.85 \\
    CoOp~\scriptsize{\textcolor{gray}{(IJCV 22)}} & 80.60 & 65.89 & 72.51 & 79.44 & 41.18 & 54.24 & 92.19 & 54.74 & 68.69 & 84.69 & 56.05 & 67.46 \\
    KgCoOp~\scriptsize{\textcolor{gray}{(CVPR 23)}} & 80.29 & 76.53 & 78.36 & 77.55 & 54.99 & 64.35 & 85.64 & 64.34 & 73.48 & 82.89 & 76.67 & 79.66 \\
    MaPLe~\scriptsize{\textcolor{gray}{(CVPR 23)}} & 80.82 & 78.70 & 79.75 & 80.36 & 59.18 & 68.16 & 94.07 & 73.23 & 82.35 & 83.00 & 78.66 & 80.77 \\
    PromptSRC~\scriptsize{\textcolor{gray}{(ICCV 23)}} & 82.67 & 78.47 & 80.52 & 83.37 & 62.97 & 71.75 & 92.90 & 73.90 & 82.32 & 87.10 & 78.80 & 82.74 \\
    MMA~\scriptsize{\textcolor{gray}{(CVPR 24)}} & 82.27 & 78.57 & 80.38 & 83.20 & 65.63 & 73.38 & 85.46 & 82.34 & 83.87 & 86.23 & 80.03 & 82.20 \\
    \midrule

    CoPrompt$^\dag$ ~\scriptsize{\textcolor{gray}{(ICLR 24)}} & 82.63 & 80.03 & 81.30 & 83.13 & 64.73 & 72.79 & 94.60 & 78.57 & 85.84 & 86.90 & 79.57 & 83.07 \\
    CoOp + ATPrompt$^\dag$ ~\scriptsize{\textcolor{gray}{(ICCV 25)}} & 80.84 & 68.64 & 74.24 & 80.83 & 45.49 & 58.22 & 90.34 & 59.79 & 71.96 & 84.49 & 64.96 & 73.45 \\

    \midrule
    CoOp + TIES~\scriptsize{\textcolor{gray}{(NeurIPS 23)}} & 62.41 & 62.99 & 62.70 & 57.33 & 41.87 & 48.40 &50.17 & 69.85 &58.40&62.45&54.68&58.30\\
    CoOp + DARE~\scriptsize{\textcolor{gray}{(ICML 24)}} &70.33&63.01&66.47&72.49&47.02&57.04&69.71&65.22&67.39&77.61&57.84& 66.28 \\
    CoOp + \ourmeth{}~\scriptsize{\textcolor{gray}{(ICLR 26)}} & 80.37 & 76.19 & 78.22 & 78.51 & 56.84  &  65.93 &  85.15 & 64.51  &  73.40 &  82.04 & 78.33  & 80.14  \\
    \rowcolor{dmcrow} \textbf{CoOp + DMC (Ours)} & 82.14 & 77.25 & 79.62 & 82.29 & 55.07 & 65.98 & 90.81 & 70.31 & 79.26 & 86.45 & 79.34 & 82.74 \\
    \midrule
    KgCoOp + TIES~\scriptsize{\textcolor{gray}{(NeurIPS 23)}} &73.05&73.88&73.46&60.77&53.54&56.93&68.66&65.45&67.02&74.46&73.91&74.18\\
    KgCoOp + DARE~\scriptsize{\textcolor{gray}{(ICML 24)}} &77.47&74.81&76.12&72.46&54.19&62.01&84.48&63.03&72.19&81.63&75.99&78.71 \\
    KgCoOp + \ourmeth{}~\scriptsize{\textcolor{gray}{(ICLR 26)}} & 80.53 & 77.28 & 78.87 & 79.90 & 57.33 & 66.75 & 89.56 & 67.89 & 77.22 & 83.35 & 77.76 & 80.45 \\
    \rowcolor{dmcrow} \textbf{KgCoOp + DMC (Ours)} & 82.06 & 76.47 & 79.17 & 81.25 & 56.52 & 66.67 & 93.05 & 71.03 & 80.56 & 86.14 & 75.66 & 80.56 \\
    \midrule
    MMA + TIES~\scriptsize{\textcolor{gray}{(NeurIPS 23)}} & 69.46&65.77&67.56&55.38&48.36&51.63&65.69&67.95&66.80&71.74&60.42&65.60\\
    MMA + DARE~\scriptsize{\textcolor{gray}{(ICML 24)}} &71.43&72.61&72.02&67.65&50.37&57.74&77.42&64.16&70.17&78.46&64.47&70.78 \\
    MMA + \ourmeth{}~\scriptsize{\textcolor{gray}{(ICLR 26)}} & 75.20 & 78.64 & 76.88 & 57.06 & 60.75 & 58.85 & 52.10 & 70.59 & 59.95 & 70.99 & 76.69 & 73.73 \\
    \rowcolor{dmcrow} \textbf{MMA + DMC (Ours)} & 76.95 & 78.87 & 77.90 & 59.03 & 61.23 & 60.11 & 53.69 & 71.21 & 61.22 & 74.77 & 77.29 & 76.01 \\
    \bottomrule
    \end{tabular}
    }
    \end{subtable}
    \vspace{-5pt}
    \label{table:base_to_novel}
\end{table*}

Table~\ref{tab:hm_meanstd_full} gives the seed-resolved view: HM mean
$\pm$ sample standard deviation for every row we trained. Of the 33 DMC cells,
23 have $\sigma_{\mathrm{HM}} \leq 0.5$ and 27 have
$\sigma_{\mathrm{HM}} \leq 1.0$. The seed-fragile cells are concentrated on
EuroSAT, DTD, FGVCAircraft and UCF101, where the corresponding \mergetune{} runs
are equally noisy (e.g., $\sigma_{\mathrm{HM}} = 3.81$ and $4.03$ on EuroSAT), so
DMC inherits this variance rather than introducing it. MMA+DMC is markedly
seed-stable (maximum $\sigma_{\mathrm{HM}} = 0.45$), consistent with the
endpoint collapse of Appendix~\ref{app:endpoint_collapse}: a degenerate corridor
leaves little to vary.

\begin{table*}[t]
\centering
\small
\setlength{\tabcolsep}{5pt}
\renewcommand{\arraystretch}{1.10}
\caption{Seed variability for every row we trained ourselves: HM mean $\pm$ sample
standard deviation over 3 seeds (DMC at $\alpha = 0.20$; \mergetune{} at its final prompt). Per-cell companion to
Table~\ref{table:hm_main}, which reports the same runs as means. Cells with
$\sigma_{\mathrm{HM}} > 1$ are \textbf{bold} and flagged as seed-fragile.}
\label{tab:hm_meanstd_full}
\resizebox{\linewidth}{!}{%
\begin{tabular}{lcccccc}
\toprule
& \multicolumn{2}{c}{\textbf{CoOp}} & \multicolumn{2}{c}{\textbf{KgCoOp}} & \multicolumn{2}{c}{\textbf{MMA}} \\
\cmidrule(lr){2-3}\cmidrule(lr){4-5}\cmidrule(lr){6-7}
Dataset & +\mergetune{} & +DMC & +\mergetune{} & +DMC & +\mergetune{} & +DMC \\
\midrule
ImageNet       & 72.89 $\pm$ 0.18 & 72.90 $\pm$ 0.18 & 72.82 $\pm$ 0.10 & 73.09 $\pm$ 0.26 & 73.15 $\pm$ 0.10 & 72.90 $\pm$ 0.01 \\
Caltech101     & 96.27 $\pm$ 0.26 & 96.41 $\pm$ 0.15 & 95.94 $\pm$ 0.03 & 96.11 $\pm$ 0.08 & 95.75 $\pm$ 0.16 & 95.74 $\pm$ 0.16 \\
OxfordPets     & 96.13 $\pm$ 0.66 & 96.04 $\pm$ 0.29 & 96.28 $\pm$ 0.10 & 96.23 $\pm$ 0.35 & 93.22 $\pm$ 0.11 & 93.38 $\pm$ 0.01 \\
StanfordCars   & 73.38 $\pm$ 0.37 & 75.21 $\pm$ 0.10 & 74.90 $\pm$ 0.29 & 75.38 $\pm$ 0.24 & 70.16 $\pm$ 0.24 & 70.51 $\pm$ 0.04 \\
Flowers102     & 83.92 $\pm$ 0.57 & 84.47 $\pm$ 0.58 & 83.39 $\pm$ 0.27 & 83.67 $\pm$ 0.33 & 75.22 $\pm$ 0.40 & 76.37 $\pm$ 0.23 \\
Food101        & 91.05 $\pm$ 0.11 & 91.13 $\pm$ 0.15 & 91.08 $\pm$ 0.11 & 91.19 $\pm$ 0.10 & 90.18 $\pm$ 0.06 & 90.33 $\pm$ 0.09 \\
FGVCAircraft   & 33.33 $\pm$ 0.42 & \textbf{35.23 $\pm$ 1.37} & \textbf{34.10 $\pm$ 1.35} & 33.17 $\pm$ 0.90 & 32.72 $\pm$ 0.10 & 33.10 $\pm$ 0.04 \\
SUN397         & 77.85 $\pm$ 0.35 & 78.75 $\pm$ 0.54 & 78.03 $\pm$ 0.53 & 78.85 $\pm$ 0.03 & 76.70 $\pm$ 0.17 & 77.12 $\pm$ 0.12 \\
DTD            & \textbf{64.26 $\pm$ 1.93} & \textbf{63.53 $\pm$ 1.37} & \textbf{62.66 $\pm$ 2.83} & \textbf{64.34 $\pm$ 1.13} & 58.07 $\pm$ 0.68 & 59.62 $\pm$ 0.13 \\
EuroSAT        & \textbf{73.45 $\pm$ 3.81} & \textbf{75.75 $\pm$ 3.65} & \textbf{73.44 $\pm$ 4.03} & \textbf{73.38 $\pm$ 3.16} & 59.57 $\pm$ 0.48 & 60.57 $\pm$ 0.45 \\
UCF101         & \textbf{78.42 $\pm$ 1.13} & \textbf{80.07 $\pm$ 1.47} & 78.08 $\pm$ 0.90 & 78.90 $\pm$ 0.94 & 73.59 $\pm$ 0.21 & 75.11 $\pm$ 0.45 \\
\midrule
\textbf{Average} & 76.45 & 77.23 & 76.43 & 76.76 & 72.58 & 73.16 \\
\bottomrule
\end{tabular}}
\end{table*}

\subsection{Failure and weak-case analysis}
\label{app:failure_modes}

DMC does not gain uniformly across datasets or baselines. We list the cases
where the gain over \mergetune{} shrinks, disappears, or reverses
(Table~\ref{tab:hm_meanstd_full}), and explain each.

\paragraph{Small regressions.}
Seven of the 33 cells are below \mergetune{}: CoOp on OxfordPets ($-0.09$) and
DTD ($-0.73$); KgCoOp on OxfordPets ($-0.05$), FGVCAircraft ($-0.93$) and
EuroSAT ($-0.06$); MMA on ImageNet ($-0.25$) and Caltech101 ($-0.01$). All but
the MMA ImageNet cell lie within one combined seed standard deviation of the two cells. The two largest prompt-tuning regressions (CoOp DTD, KgCoOp
FGVCAircraft) occur on datasets whose \mergetune{} runs are themselves
seed-fragile ($\sigma_{\mathrm{HM}} = 1.93$ and $1.35$).

\paragraph{Near-ceiling regime: Caltech101 and OxfordPets.}
Both datasets have zero-shot HM already close to $95$. The interpolation
frontier is flat there: base and novel accuracy move together across $\alpha$,
so the gap between the single-prompt operating point and the DMC corridor is
small and DMC's margin shrinks to at most $0.17$~HM in either direction. We
attribute this to the ceiling rather than to a failure of the method.

\paragraph{MMA endpoint collapse.}
On MMA the two learned endpoints converge to nearly identical text features
($\cos(f_{\mathrm{gen}}, f_{\mathrm{spec}}) > 0.99$ on 9 of 10 datasets), and the
interpolation curve carries little of DMC's effect. It is the clearest failure case in our study and is analyzed in full in
Appendix~\ref{app:endpoint_collapse}.

\paragraph{Seed-fragile cells.}
Six DMC cells have $\sigma_{\mathrm{HM}} > 1$: EuroSAT (CoOp $3.65$, KgCoOp
$3.16$), DTD (CoOp $1.37$, KgCoOp $1.13$), FGVCAircraft (CoOp $1.37$) and UCF101
(CoOp $1.47$). Four have positive mean gains over \mergetune{} (CoOp on
EuroSAT, FGVCAircraft and UCF101; KgCoOp on DTD) and two are slightly negative
(CoOp on DTD, KgCoOp on EuroSAT), but in all but one case the seed spread overlaps
zero (CoOp on FGVCAircraft, $+1.90$, is just outside one combined standard deviation), so we do not claim a reliable effect on these cells individually. EuroSAT
is the extreme case: with five classes per split, a single class changing its
predictions moves HM by several points.

\FloatBarrier

\FloatBarrier

\section{Extended Interpolation Curves}
\label{app:pareto}

This section shows whether the operating-point shift of Figure~\ref{fig:pareto} holds on the remaining datasets.
Each figure shows the interpolation curves traced by \textbf{DMC}
(solid blue) and \textbf{DMC W/O VA} (dashed vermillion)
across $\alpha \in [0, 1]$, with the \ourmeth{} baseline (gray
dot), the shift to the operating point (green arrow) and the operating point (blue star), on the CoOp base. The boxed
$\Delta\mathrm{HM}$ value reports the gap between the operating
point and the \ourmeth{} baseline, read off the plotted seed-averaged curves; it differs from the corresponding entry of Table~\ref{tab:hm_meanstd_full} by at most $0.18$~HM (OxfordPets: $+0.09$ here, $-0.09$ in the table).

\begin{figure*}[!htbp]
\centering
\includegraphics[width=\textwidth]{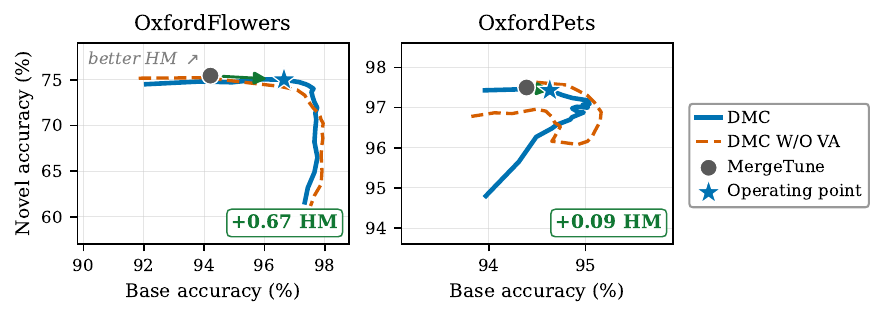}
\caption{\textbf{Flowers102 and OxfordPets.}
Flowers102 shows a shift toward higher base accuracy at nearly unchanged novel accuracy ($+0.67$~HM).
OxfordPets has a near-ceiling baseline; DMC is within $0.1$~HM of it ($+0.09$ on the plotted curves, $-0.09$ in Table~\ref{tab:hm_meanstd_full}).}
\label{fig:pareto_flowers_pets}
\end{figure*}

\begin{figure*}[!htbp]
\centering
\includegraphics[width=\textwidth]{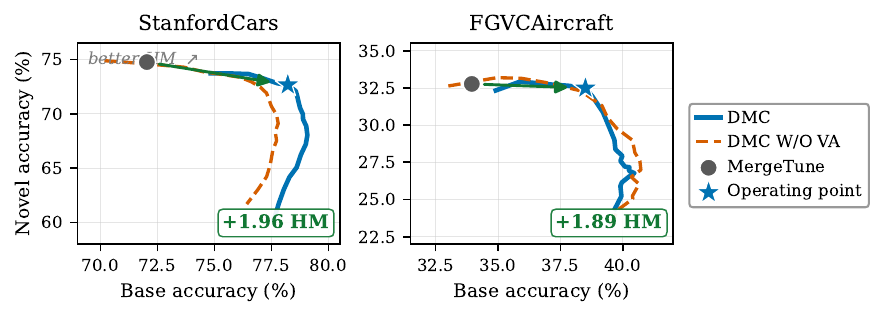}
\caption{\textbf{StanfordCars and FGVCAircraft.}
Both fine-grained datasets show clear operating-point shifts:
$+1.96$~HM on StanfordCars and $+1.89$ on FGVCAircraft.}
\label{fig:pareto_cars_air}
\end{figure*}

\begin{figure*}[!htbp]
\centering
\includegraphics[width=\textwidth]{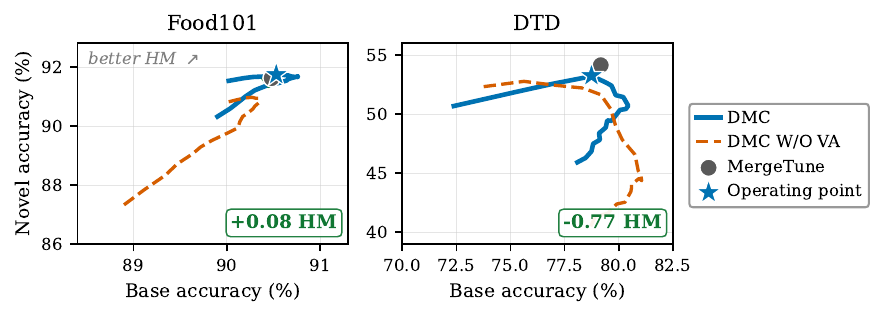}
\caption{\textbf{Food101 and DTD.}
Food101 has a flat frontier; DMC shifts the operating point
slightly ($+0.08$~HM). DTD is the only dataset in these figures where DMC
is slightly below the baseline ($-0.77$~HM).}
\label{fig:pareto_food_dtd}
\end{figure*}

\begin{figure*}[!htbp]
\centering
\includegraphics[width=\textwidth]{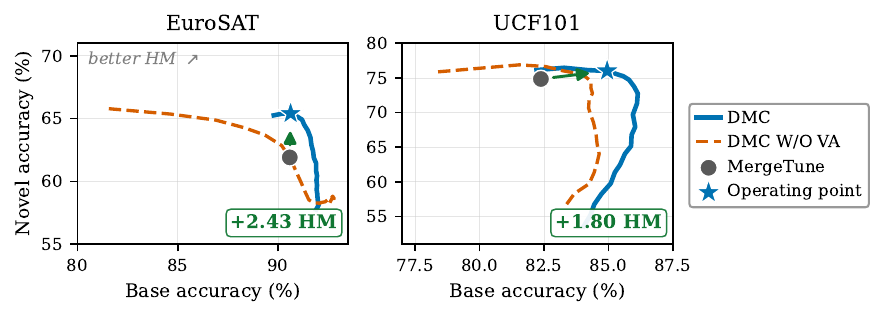}
\caption{\textbf{EuroSAT and UCF101.}
EuroSAT shows the largest improvement among these datasets ($+2.43$~HM).
UCF101 achieves $+1.80$~HM. (Both datasets are also shown in
the main paper, Figure~\ref{fig:pareto}.)}
\label{fig:pareto_euro_ucf}
\end{figure*}

\begin{figure*}[!htbp]
\centering
\includegraphics[width=\textwidth]{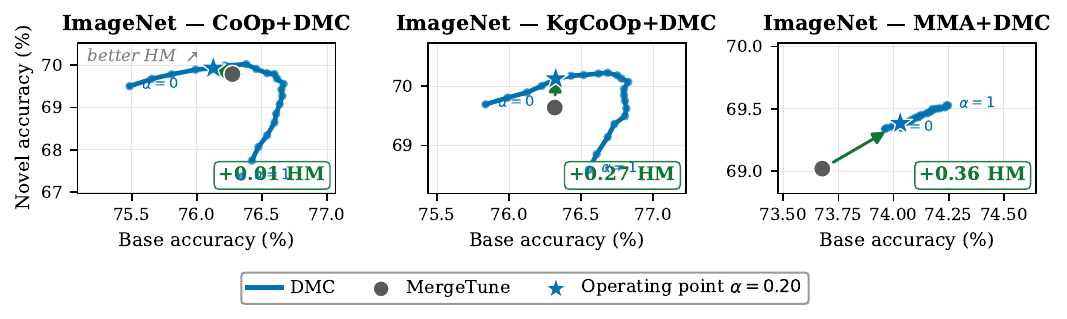}
\caption{\textbf{ImageNet — CoOp+DMC, KgCoOp+DMC, MMA+DMC.}
Interpolation curves on ImageNet (seed mean of the 21-point $\alpha$
sweep). The blue star marks the operating point at the fixed
$\alpha = 0.20$ (the same value used for all held-out datasets,
selected on Flowers102+Caltech101). At the operating point, DMC yields
+0.01~HM on CoOp+DMC ($72.89 \to 72.90$), +0.27~HM on KgCoOp+DMC
($72.82 \to 73.09$), and $-0.25$~HM on MMA+DMC ($73.15 \to 72.90$; the MMA panel shows an earlier single run, whose boxed value differs)
over MergeTune. The operating-point shift on ImageNet is small relative to
the specialized datasets (Figure~\ref{fig:pareto}); this is
consistent with ImageNet's high-diversity 1000-class set, where
the single-prompt envelope already covers most of the achievable
operating points and decoupling has less headroom than on
specialized tasks like EuroSAT or StanfordCars. Unlike the other interpolation-curve figures
in this appendix, only the DMC (with VA) curve is shown: a
DMC-without-VA ablation was not run on ImageNet.}
\label{fig:pareto_imagenet}
\end{figure*}

\FloatBarrier

\FloatBarrier

\section{Quadratic Model of the Single-Prompt Ceiling}
\label{app:quadratic_decoupling}

This section asks why a single prompt trained on a weighted sum of two objectives must settle for a compromise.
To build intuition for the structural bottleneck in Section~\ref{sec:method},
consider a minimal scalar model with two competing objectives:
\begin{gather*}
  L_1(\theta) = (\theta - a)^2 \quad \text{(specialization, minimized at } a\text{)},
  \\
  L_2(\theta) = (\theta - b)^2 \quad \text{(generalization, minimized at } b \neq a\text{)}.
\end{gather*}

\paragraph{Single-prompt scalarization.}
The weighted objective $(1-\lambda)L_1(\theta) + \lambda L_2(\theta)$ has the unique
stationary point $\theta^*(\lambda) = (1-\lambda)a + \lambda b$ for $\lambda \in (0,1)$.
By Lemma~\ref{lem:pareto} below, $\theta^*(\lambda)$ is Pareto-stationary, and neither
objective is individually minimized:
\[
  L_1\!\bigl(\theta^*(\lambda)\bigr) = \lambda^2(b - a)^2 > 0, \qquad
  L_2\!\bigl(\theta^*(\lambda)\bigr) = (1-\lambda)^2(b - a)^2 > 0.
\]
As $\lambda$ varies over $(0,1)$, the pair $(L_1(\theta^*(\lambda)),\, L_2(\theta^*(\lambda)))$
traces a quadratic arc that never reaches either axis of the $(L_1,L_2)$ plane.
The individual optima $(0,\,(b-a)^2)$ and $((b-a)^2,\,0)$
are unreachable by any single-prompt scalarization.

\paragraph{Decoupled two-parameter model.}
With two independent parameters $(\theta_1, \theta_2)$, decoupled optimization of
$L_1(\theta_1) + L_2(\theta_2)$ gives $\theta_1^* = a$ and $\theta_2^* = b$.
Each objective is then exactly minimized by its own parameter, so the joint
objective reaches $0$, whereas any single $\theta$ incurs
$L_1(\theta) + L_2(\theta) \geq (b-a)^2/2$.
The interpolated predictor $\tilde\theta(\alpha) = (1-\alpha)\theta_2^* + \alpha\theta_1^*$ (as in Eq.~\eqref{eq:interp}, $\alpha = 0$ is the generalization end)
recovers every single-parameter compromise ($\tilde\theta(1-\lambda) = \theta^*(\lambda)$),
so decoupling loses nothing and turns the trade-off into a post-hoc choice of
$\alpha$ instead of a training-time choice of $\lambda$.
In one dimension the interpolation path and the single-parameter path coincide,
so this model explains why a single prompt must compromise, not how far a
decoupled corridor can move beyond the single-prompt operating point; the latter
is an empirical property of the high-dimensional prompt space, which
Figure~\ref{fig:pareto} and Appendix~\ref{app:pareto} measure.

The same picture holds for the prompt (Lemma~\ref{lem:pareto}, Section~\ref{sec:prelim}).

The proof is in Appendix~\ref{app:proofs}. The lemma is the high-dimensional analogue of the scalar picture: whenever both gradients are nonzero at convergence, the single prompt is a compromise that minimizes neither objective. Appendix~\ref{app:gamma_table} checks the nonzero-and-opposed gradient condition empirically.

\FloatBarrier

\section{Gradient Conflict: Per-dataset Measurements}
\label{app:gamma_table}

This section asks whether the two single-prompt gradients actually conflict at convergence, and whether the size of the conflict predicts DMC's gain.
Figure~\ref{fig:gradient_conflict} in the main paper reports the
gradient conflict statistic $\rho = \cos(\nabla_{\mathbf{c}}
\mathcal{L}_{\mathrm{CE}},\, \nabla_{\mathbf{c}} \mathcal{R})$
at the converged \mergetune{} checkpoint. Table~\ref{tab:gamma}
provides the exact per-dataset values alongside the corresponding
DMC HM gain.

\begin{table}[!htbp]
\centering
\caption{Gradient conflict $\rho = \cos(\nabla_{\mathbf{c}}
\mathcal{L}_{\mathrm{CE}},\, \nabla_{\mathbf{c}} \mathcal{R})$
at convergence, grouped by baseline. $\rho < 0$ on all eight
dataset--baseline pairs. The DMC gain is the HM gain over \mergetune{} in the
runs where $\rho$ was measured; it can differ from the seed means of
Table~\ref{table:hm_main}.}
\label{tab:gamma}
\small
\setlength{\tabcolsep}{5pt}
\begin{tabular}{lcc}
\toprule
\textbf{Dataset} &
$\boldsymbol{\rho}$ & \textbf{DMC gain} \\
\midrule
\multicolumn{3}{l}{\textit{CoOp baseline}} \\
\quad Flowers102 & $-0.955$ & $+0.23$ \\
\quad Caltech101    & $-0.716$ & $+0.36$ \\
\quad Food101       & $-0.776$ & $+0.08$ \\
\quad OxfordPets    & $-0.211$ & $+0.22$ \\
\midrule
\multicolumn{3}{l}{\textit{KgCoOp baseline}} \\
\quad Flowers102 & $-0.923$ & $+0.33$ \\
\quad Caltech101    & $-0.289$ & $+0.32$ \\
\quad Food101       & $-0.528$ & $+0.11$ \\
\quad OxfordPets    & $-0.580$ & $+0.04$ \\
\midrule
\multicolumn{2}{l}{Pearson $r$ ($\rho$ vs gain)}
    & $-0.05$ \\
\bottomrule
\end{tabular}
\end{table}

The conflict is present everywhere, but its magnitude does not predict
DMC's gain: Caltech101 gains $+0.36$~HM at $\rho = -0.72$, while Food101 gains
only $+0.08$ at $\rho = -0.78$, and across the eight pairs the correlation is
negligible (Pearson $r = -0.05$). A negative $\rho$ indicates that the single
prompt sits at a first-order trade-off (Lemma~\ref{lem:pareto}; at an exact
stationary point of $\mathcal{L}_{\mathrm{CE}} + \lambda\mathcal{R}$ alone,
$\rho = -1$, and values above $-1$ reflect approximate convergence and the
additional terms of the \mergetune{} objective); how much DMC
gains depends on the shape of the interpolation curve beyond that operating
point, which $\rho$ does not measure. We therefore use $\rho$ as evidence that
the ceiling exists, not as a predictor of where decoupling helps most.

\FloatBarrier

\FloatBarrier

\section{Single-Prompt Loss Modifications}
\label{app:loss_modifications}

This section asks whether a better single-prompt loss, rather than a second prompt, could deliver DMC's gain.
We test whether targeted single-prompt loss modifications can match the DMC frontier.
Seven variants are evaluated on Caltech101 and Flowers102 as drop-in replacements or
additions to the cosine score regularizer in \mergetune{}'s objective:

\emph{1. Fisher-weighted cosine.} Replace isotropic cosine distance with Fisher-diagonal-weighted cosine distance, with two normalizations (\textit{max}, \textit{per-class}) and $\lambda \in \{4, 8, 10, 16\}$.
\emph{2. Prompt-space EWC.} Add a Fisher-weighted $\ell_2$ penalty on prompt parameters toward $\hat{w}_2$, swept over $\lambda_{\mathrm{pf}} \in \{0.01, 0.05, 0.1, 0.5, 1.0\}$.
\emph{3. KL path.} Enforce KL-divergence preservation along the $f_{w_1} \to f(\mathbf{c})$ path, swept over $\lambda_{\mathrm{kl}} \in \{0.1, 0.3, 0.5, 1.0\}$.
\emph{4. ZSDD.} Distill the zero-shot prediction distribution at temperature $\tau$, swept over $\lambda_{\mathrm{zsdd}} \in \{0.5, 1, 2, 4\}$ and $\tau \in \{2, 4\}$.
\emph{5. Weight scheduling.} Anneal the cosine weight $\lambda$ over training, with linear or cosine schedules and final $\lambda_{\min} \in \{0, 2, 5\}$.
\emph{6. Visual Anchor (single-prompt).} Apply VA on the same prompt that receives CE, instead of the decoupled placement used in DMC.
\emph{7. TTAI.} Test-time adaptive interpolation with per-image $\alpha$ predicted from visual features.

Table~\ref{tab:loss_mods} reports the best HM achieved by each modification on each dataset (best across the swept hyperparameters, seed~1, $\beta = 1.0$, the \mergetune{} trainer with a CoOp initialization, ViT-B/16). The cosine-score baseline at $\lambda = 8$ achieves $96.03$~HM on Caltech101 and $83.80$~HM on Flowers102; as single-seed numbers, they differ from the seed means of Table~\ref{table:hm_main}.

\begin{table}[!htbp]
\centering
\caption{Single-prompt loss modifications on Caltech101 and Flowers102. For each modification we report the best HM across the hyperparameter sweep; $\Delta$~HM is the gap to the cosine-score baseline at $\lambda = 8$. No modification improves on the baseline by clearly more than retuning $\lambda$ does, and modifications that gain on one dataset typically regress on the other.}
\label{tab:loss_mods}
\small
\setlength{\tabcolsep}{4pt}
\renewcommand{\arraystretch}{1.05}
\resizebox{\linewidth}{!}{%
\begin{tabular}{lcccc|cccc}
\toprule
& \multicolumn{4}{c}{\textbf{Caltech101}} & \multicolumn{4}{c}{\textbf{Flowers102}} \\
\cmidrule(lr){2-5}\cmidrule(lr){6-9}
\textbf{Modification} & Base & Novel & HM & $\Delta$ & Base & Novel & HM & $\Delta$ \\
\midrule
Cosine score, $\lambda = 8$ (\mergetune{}) & 97.93 & 94.21 & 96.03 & --- & 94.21 & 75.46 & 83.80 & --- \\
\midrule
Fisher-weighted cosine & 97.42 & 95.09 & 96.24 & \incre{$+0.21$} & 94.11 & 75.89 & 84.02 & \incre{$+0.22$} \\
Prompt-space EWC & 97.87 & 94.32 & 96.06 & \incre{$+0.03$} & 94.59 & 75.25 & 83.82 & \incre{$+0.02$} \\
KL path & 97.87 & 94.10 & 95.95 & \decli{$-0.08$} & 94.02 & 75.60 & 83.81 & \incre{$+0.01$} \\
ZSDD & 97.48 & 95.41 & 96.43 & \incre{$+0.40$} & 87.46 & 76.17 & 81.43 & \decli{$-2.37$} \\
Weight scheduling & 98.13 & 92.79 & 95.39 & \decli{$-0.64$} & 94.68 & 72.84 & 82.34 & \decli{$-1.46$} \\
Visual Anchor (single-prompt) & 97.61 & 94.43 & 95.99 & \decli{$-0.04$} & 95.25 & 75.39 & 84.16 & \incre{$+0.36$} \\
TTAI & 97.93 & 94.43 & 96.15 & \incre{$+0.12$} & 90.98 & 76.45 & 83.08 & \decli{$-0.72$} \\
\bottomrule
\end{tabular}}
\end{table}

The largest single-dataset gain is $+0.40$~HM (ZSDD on Caltech101), and the largest gain on Flowers102 is $+0.36$~HM (single-prompt Visual Anchor). Both are within seed-variance scale and of the same order as what cosine-weight tuning alone achieves: sweeping $\lambda \in \{4, 8, 10, 16\}$ moves the cosine baseline by $+0.22$~HM on Caltech101 and $+0.26$~HM on Flowers102; they exceed it by only $0.18$ and $0.10$~HM, respectively. Modifications that gain on one dataset typically regress on the other: ZSDD ($+0.40$ Caltech101, $-2.37$ Flowers102), single-prompt VA ($-0.04$ Caltech101, $+0.36$ Flowers102), TTAI ($+0.12$ Caltech101, $-0.72$ Flowers102). Fisher-weighted cosine is the only variant that gains noticeably on both datasets, by $+0.21$ and $+0.22$~HM, i.e., by about as much as retuning $\lambda$. Fisher-weighted cosine, which violates CPI, also fails to break the ceiling, consistent with Appendix~\ref{app:cpi}. The CPI test there measures a feature-space EWC variant, which is distinct from the prompt-space EWC evaluated here.

For comparison, on the same Flowers102 setup (seed~1, $\beta = 1.0$, best $\alpha$, matching the best-over-sweep protocol above), DMC reaches HM~$84.66$ (Table~\ref{tab:wgen_sweep}), a $+0.86$~HM gain over the cosine baseline that is two to three times larger than the best single-prompt modification on this dataset. The pattern supports the structural-bottleneck claim: with the parameterization fixed to a single prompt, none of the seven regularizers we tested produces gains comparable to architectural decoupling.

\FloatBarrier

\FloatBarrier

\section{Supporting Tables for the Ablations}
\label{app:ablation_tables}

This section collects the tables behind two ablation questions of
Section~\ref{sec:ablation}: does extra single-prompt capacity explain DMC's
gain, and how much does the Visual Anchor contribute? (The four-dataset
component ablation is Table~\ref{tab:ablation_R} in the main text.)

\paragraph{Larger single prompt.} Table~\ref{tab:larger_single_prompt} trains
CoOp with $n_{\text{ctx}} \in \{8, 16\}$. At $n_{\text{ctx}} = 16$ the single
prompt has twice DMC's parameter count, yet CoOp+DMC leads by $5.07$, $1.07$ and
$5.00$~HM on Flowers102, Caltech101 and StanfordCars; on EuroSAT the $3.46$~HM
lead is within the seed spread. Parameter count alone does not account for the
gain.

\paragraph{Visual Anchor.} Tables~\ref{tab:va_single_vs_two_view}
and~\ref{tab:va_aug_ablation} vary the VA views and augmentation. The
configurations span about $1.0$~HM on Flowers102 and $0.32$~HM on Caltech101, and
most differences between them are within one seed standard deviation, consistent with the main-text
finding that VA is a small refinement rather than the source of DMC's gain. The
exact single-view case of Proposition~\ref{prop:va_kl} ($v_a = v_b$, preset
\textit{none}) is slightly below the two-view default on both datasets. These
runs retrain the configuration of Table~\ref{table:hm_main} (batch size 32): Caltech101 reproduces exactly ($96.41$), while Flowers102 gives $84.17$ versus $84.47$, within the seed spread. The \textit{No VA} row comes from the component-ablation runs (batch size 128; Table~\ref{tab:ablation_R}).

\begin{table}[t]
\centering
\small
\setlength{\tabcolsep}{6pt}
\renewcommand{\arraystretch}{1.10}
\caption{Extra single-prompt capacity does not match CoOp+DMC.
Base-to-novel harmonic mean (HM) on four datasets. We train CoOp with
enlarged context length $n_\text{ctx} \in \{8, 16\}$ using the same protocol as
the standard $n_\text{ctx}=4$ CoOp; $n_\text{ctx}=16$ gives a single prompt
with $16 \times 512 = 8192$ tunable parameters, strictly larger than the
$2 \times 4 \times 512 = 4096$ parameters of CoOp+DMC.
Values are mean $\pm$ sample standard deviation over 3 seeds; the
$n_\text{ctx}=4$ row is the single number reported in the main paper.
CoOp+DMC still leads on every dataset; on EuroSAT the margin is within the seed spread.}
\label{tab:larger_single_prompt}
\begin{tabular}{lcccc}
\toprule
Method & Flowers102 & Caltech101 & StanfordCars & EuroSAT \\
\midrule
CoOp ($n_\text{ctx}=4$, from Table~\ref{table:hm_main})   & 74.06            & 93.73            & 68.13            & 68.69 \\
CoOp ($n_\text{ctx}=8$)                & 77.84 $\pm$ 0.36 & 92.19 $\pm$ 3.20 & 68.35 $\pm$ 1.23 & 69.95 $\pm$ 4.91 \\
CoOp ($n_\text{ctx}=16$)               & 79.40 $\pm$ 1.50 & 95.34 $\pm$ 0.64 & 70.21 $\pm$ 1.05 & 72.29 $\pm$ 9.52 \\
\midrule
\textbf{CoOp + DMC (Ours)}             & \textbf{84.47 $\pm$ 0.58} & \textbf{96.41 $\pm$ 0.15} & \textbf{75.21 $\pm$ 0.10} & \textbf{75.75 $\pm$ 3.65} \\
\bottomrule
\end{tabular}
\end{table}

\begin{table}[t]
\centering

\small
\setlength{\tabcolsep}{6pt}
\renewcommand{\arraystretch}{1.10}
\caption{Decomposing the Visual Anchor (VA) benefit into KL anchoring vs.
augmentation consistency. \textit{No VA} disables the term entirely.
\textit{Single-view VA}: the student branch sees the raw image and the
teacher branch one train-transform view; this is the closest variant to the
single-view case of Proposition~\ref{prop:va_kl}, which it matches exactly only
without augmentation (preset \textit{none} in Table~\ref{tab:va_aug_ablation}).
\textit{Two-view VA} is the default used in all reported experiments: both
branches see independent train-transform views. $\Delta$ is the HM change relative to
the row above. All cells at $\alpha = 0.20$, $\beta = 4.0$, 3 seeds.
Trainer: CoOp + DMC. \textit{No VA} is from the component-ablation runs (batch size 128); the VA rows use batch size 32, so the first $\Delta$ also includes the batch-size change.}
\label{tab:va_single_vs_two_view}
\begin{tabular}{lcc cc}
\toprule
 & \multicolumn{2}{c}{Flowers102} & \multicolumn{2}{c}{Caltech101} \\
\cmidrule(lr){2-3} \cmidrule(lr){4-5}
Configuration       & HM              & $\Delta$        & HM              & $\Delta$ \\
\midrule
No VA               & 83.91 $\pm$ 0.46 & ---             & 96.25 $\pm$ 0.04 & --- \\
Single-view VA            & 84.23 $\pm$ 0.16 & $+0.32$    & 96.26 $\pm$ 0.25 & $+0.01$ \\
Two-view VA (default)    & 84.17 $\pm$ 0.84 & $-0.06$    & 96.41 $\pm$ 0.15 & $+0.15$ \\
\bottomrule
\end{tabular}
\end{table}

\begin{table}[t]
\centering
\small
\setlength{\tabcolsep}{5pt}
\renewcommand{\arraystretch}{1.10}
\caption{Sensitivity of CoOp + DMC to the visual-anchor augmentation preset.
Presets are: \textit{none} (raw image, no augmentation),
\textit{weak} (mild crop and jitter, no grayscale),
\textit{default} (the paper setting: RandomResizedCrop $0.7$--$1.0$, HorizontalFlip $0.5$, ColorJitter $0.4/0.4/0.2/0.1$, RandomGrayscale $0.1$), and
\textit{strong} (aggressive crop $0.2$--$1.0$, strong jitter, RandomGrayscale $0.2$).
All cells are evaluated at $\alpha = 0.20$ and reported as mean $\pm$ sample standard deviation over 3 seeds.
Trainer: CoOp + DMC.}
\label{tab:va_aug_ablation}
\begin{tabular}{lccc@{\hskip 12pt}ccc}
\toprule
 & \multicolumn{3}{c}{Flowers102} & \multicolumn{3}{c}{Caltech101} \\
\cmidrule(lr){2-4} \cmidrule(lr){5-7}
Preset  & Base & Novel & HM & Base & Novel & HM \\
\midrule
none    & 96.65 $\pm$ 0.64 & 73.88 $\pm$ 0.74 & 83.74 $\pm$ 0.70 & 97.89 $\pm$ 0.32 & 94.36 $\pm$ 0.33 & 96.09 $\pm$ 0.30 \\
weak    & 96.68 $\pm$ 0.17 & 75.44 $\pm$ 0.27 & 84.75 $\pm$ 0.14 & 97.96 $\pm$ 0.38 & 94.58 $\pm$ 0.71 & 96.24 $\pm$ 0.24 \\
default & 97.02 $\pm$ 0.79 & 74.32 $\pm$ 1.10 & 84.17 $\pm$ 0.84 & 98.13 $\pm$ 0.30 & 94.76 $\pm$ 0.55 & 96.41 $\pm$ 0.15 \\
strong  & 97.12 $\pm$ 0.24 & 74.73 $\pm$ 0.84 & 84.46 $\pm$ 0.53 & 98.23 $\pm$ 0.20 & 94.58 $\pm$ 0.28 & 96.37 $\pm$ 0.13 \\
\bottomrule
\end{tabular}
\end{table}

\FloatBarrier

\FloatBarrier

\section{HM Landscape: Complementary View}
\label{app:hm_landscape}

This section asks whether the loss-landscape picture of Figure~\ref{fig:overview} carries over to the metric we report, the harmonic mean.

\begin{figure}[!htbp]
\centering
\includegraphics[width=\linewidth]{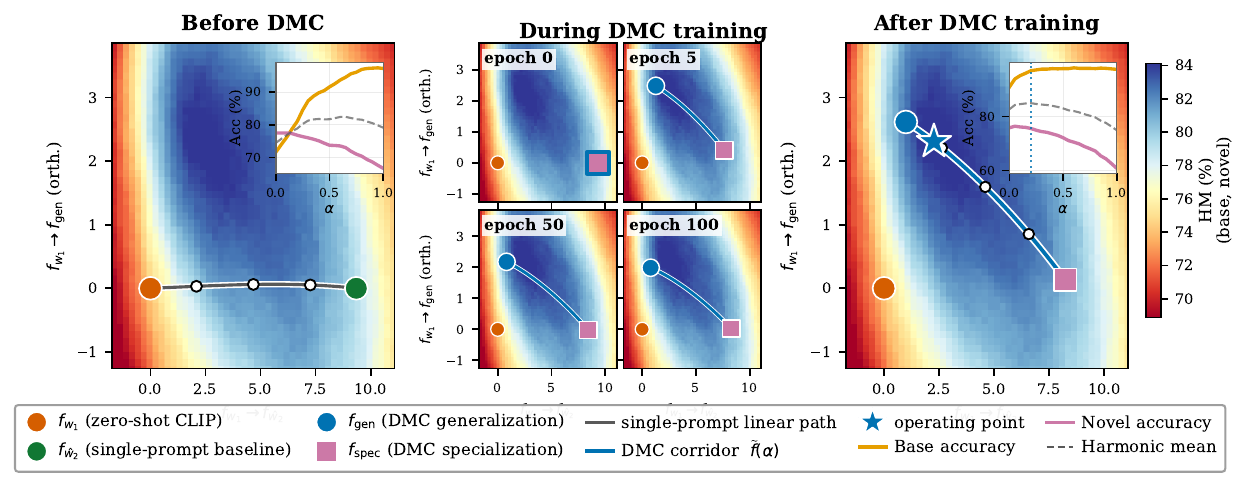}
\caption{\textbf{HM landscape on Flowers102.} Same 2D plane and overlays as the main-paper Figure~\ref{fig:overview}, with the heatmap recolored by harmonic mean of base and novel accuracy (high HM in blue). The DMC corridor lies on the high-HM peak; the operating point at $\alpha = 0.20$ sits near the corridor's HM maximum $\approx 84$. The single-prompt linear path along the bottom of the plane traverses an HM ridge that peaks below $\approx 80$ at every mixing weight, consistent with the single-prompt ceiling of Lemma~\ref{lem:pareto} along this path.}
\label{fig:hm_landscape_flowers}
\end{figure}

\begin{figure}[!htbp]
\centering
\includegraphics[width=\linewidth]{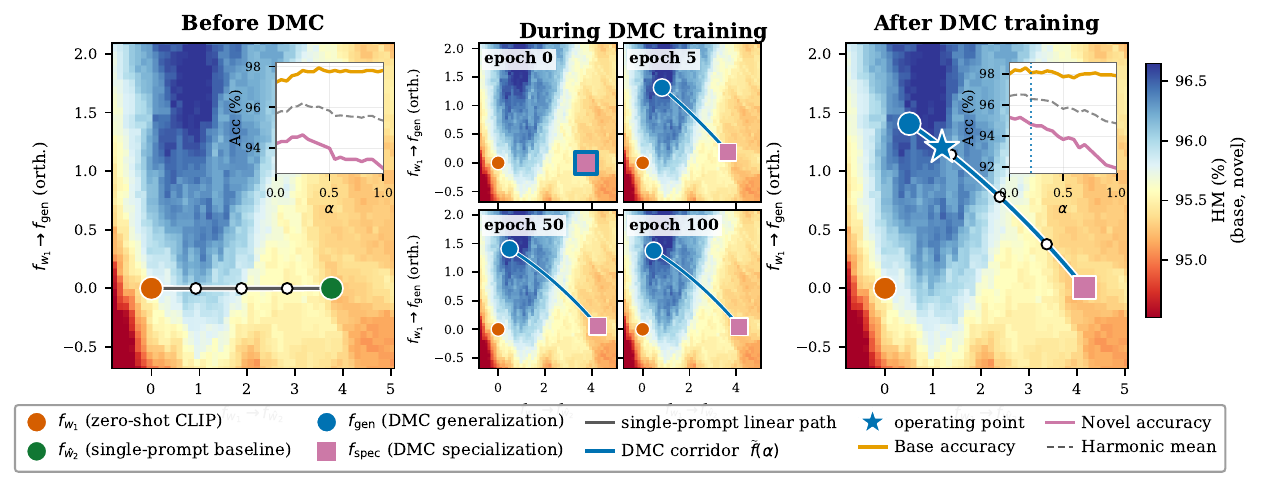}
\caption{\textbf{HM landscape on Caltech101.} Caltech101 is a near-ceiling task (zero-shot CLIP attains HM~$\approx 95$), so the landscape varies over a much narrower range, $[92.8, 96.9]$, than Flowers102 ($[64.9, 84.6]$). Despite the compressed dynamic range, the same geometry is visible: the DMC corridor lies on a high-HM ridge offset from the linear single-prompt path, and the operating point at $\alpha = 0.20$ sits near the ridge maximum. Flat-but-shifted performance is the expected near-ceiling behavior of decoupling: when single-prompt HM is already close to the achievable upper bound, DMC's gain shrinks (CoOp+DMC gains $+0.14$~HM over \mergetune{} on Caltech101, Table~\ref{table:hm_main}).}
\label{fig:hm_landscape_caltech}
\end{figure}

\paragraph{Plane construction.}
The 2D plane shared by Figure~\ref{fig:overview} (joint CE loss) and the HM landscapes below is anchored at the zero-shot CLIP features $f_{w_1}$ in concatenated base$\cup$novel text-feature space ($\mathbb{R}^{C \times 512}$, flattened). Two basis directions are derived from a trained DMC checkpoint: $d_1$ is the unit vector toward $f_{\hat{w}_2}$ (single-prompt CoOp baseline), and $d_2$ is the Gram--Schmidt orthogonalization of $f_{\mathrm{gen}} - f_{w_1}$ against $d_1$. Each grid point $(x, y)$ corresponds to a per-class feature matrix $f(x, y) = \ell_2\bigl(f_{w_1} + x\, d_1 + y\, d_2\bigr)$, $\ell_2$-normalized per class, where $x$ is in units of $\lVert f_{\hat{w}_2} - f_{w_1} \rVert$ and $y$ is in units of $\lVert (f_{\mathrm{gen}} - f_{w_1})_{\perp} \rVert$. At each grid point we recompute the classification logits with the frozen CLIP image encoder over the held-out base and novel splits and report (i)~the joint base+novel cross-entropy (Figure~\ref{fig:overview}) or (ii)~the harmonic mean of base and novel accuracies (Figures~\ref{fig:hm_landscape_flowers},~\ref{fig:hm_landscape_caltech}). By construction, $f_{w_1}$ sits at the origin, $f_{\hat{w}_2}$ on the x-axis at $(1, 0)$, and $f_{\mathrm{gen}}$ in the upper half-plane; $f_{\mathrm{spec}}$, the operating point at $\alpha = 0.20$, and the DMC corridor are projected onto the same plane and overlaid on top of the heatmap. The single-prompt linear path corresponds to varying $x \in [0, 1]$ at $y = 0$.

\paragraph{HM and CE views agree.}
The CE landscape (main-paper Figure~\ref{fig:overview}) and the HM landscape share the same plane and overlays; only the heatmap value differs (CE loss vs.\ HM accuracy). Where the CE landscape shows a low-loss \emph{valley}, the HM landscape shows a high-HM \emph{peak}. Two structural observations transfer cleanly between the two views:
(i)~the high-HM region is offset from the single-prompt linear path, so moving the single-prompt solution toward (or away from) $f_{w_1}$ does not reach it;
(ii)~the DMC corridor sits inside the high-HM region, as the LMC term encourages by keeping classification loss low along the interpolation. Figure~\ref{fig:hm_landscape_flowers} (Flowers102) and Figure~\ref{fig:hm_landscape_caltech} (Caltech101) display these effects on two datasets with very different dynamic ranges.

\paragraph{Flowers102: large dynamic range.}
The HM on Flowers102 spans roughly $20$ points over the plotted plane ($[64.9, 84.6]$), with the DMC operating point at $\approx 84$, so the offset between the high-HM peak and the linear single-prompt path is visually dramatic. The DMC corridor traces the rim of this peak, and the fixed operating point at $\alpha = 0.20$ sits near its HM maximum.

\paragraph{Caltech101: near-ceiling regime, same geometry.}
Caltech101 covers an HM range of $\sim 4$ points because zero-shot accuracy is already high. The compressed dynamic range obscures the visual offset, yet the same geometry holds: the DMC corridor sits on a high-HM ridge that the single-prompt linear path never reaches. The corridor's offset thus scales with the dataset's HM range, which mirrors the dataset-level pattern in Table~\ref{table:hm_main}: large gains on fine-grained tasks, small gains on near-ceiling tasks. Both figures support the same conclusion as the loss landscape: DMC's benefit comes from reaching regions of this plane that the single-prompt linear path does not visit, rather than from a different choice along the same curve.

\FloatBarrier

\FloatBarrier

\section{When Does DMC Compose? The Endpoint-Separation Condition}
\label{app:endpoint_collapse}

This section asks when DMC can help a given base method. DMC interpolates
between two learned endpoints; if they coincide, every $\widetilde{f}(\alpha)$ is
the same classifier and the corridor carries no information. We make this
condition measurable, use it to explain the MMA result, test its predictions,
and check it on a third parameterization.

\subsection{The diagnostic}
We measure $\cos(f_{\mathrm{gen}}, f_{\mathrm{spec}})$ at the evaluated (best-validation) checkpoint on
10 datasets (Table~\ref{tab:endpoint_collapse}). The three base methods separate
cleanly. CoOp averages $0.600$ and KgCoOp $0.719$, with no dataset above $0.99$.
MMA averages $0.993$ and exceeds $0.99$ on 9 of 10 datasets (all but SUN397, at
$0.940$); on Caltech101 its $\alpha$-sweep is flat to within $0.02$~HM
(Table~\ref{tab:alpha_oracle_per_dataset}). The ordering CoOp $<$ KgCoOp $<$ MMA
matches the ordering of DMC's average HM change relative to the published base
method ($+6.40$, $+0.05$ and $-6.56$, averaging the per-dataset HMs for every row; Table~\ref{table:hm_main}), and the
diagnostic is measurable at training time without any novel-class data.

\begin{table}[t]
\centering

\small
\setlength{\tabcolsep}{6pt}
\renewcommand{\arraystretch}{1.10}
\caption{Endpoint-collapse diagnostic for DMC. We report
$\cos(f_{\text{gen}}, f_{\text{spec}})$, the cosine similarity between the two
DMC endpoints (class text features, averaged over classes) at the evaluated best-validation checkpoint, as 3-seed mean $\pm$ sample std. Runs: $\lambda_{\text{gen}} = 8$, $\beta = 4$, without VA; with VA (the runs of Table~\ref{table:hm_main}) the averages are $0.628$, $0.704$ and $0.996$, with SUN397 again the only MMA dataset below $0.99$ ($0.964$). Values close
to $1.0$ mean the two endpoints have collapsed to the same solution, so the
linear connector degenerates and the $\alpha$-sweep is flat. CoOp and KgCoOp
endpoints remain distinct on every dataset (maximum $0.987$). MMA endpoints
exceed $0.99$ on 9 of 10 datasets (all but SUN397, at $0.940$); this collapse
delimits DMC's scope: on MMA it cannot close the gap to published MMA
(Section~\ref{sec:exp}).}
\label{tab:endpoint_collapse}
\begin{tabular}{lccc}
\toprule
Dataset & CoOp + DMC & KgCoOp + DMC & MMA + DMC \\
\midrule
Caltech101   & 0.608 $\pm$ 0.065 & 0.919 $\pm$ 0.034 & \textbf{1.000 $\pm$ 0.000} \\
OxfordPets   & 0.785 $\pm$ 0.054 & 0.971 $\pm$ 0.022 & \textbf{1.000 $\pm$ 0.000} \\
StanfordCars & 0.673 $\pm$ 0.038 & 0.701 $\pm$ 0.009 & \textbf{0.999 $\pm$ 0.000} \\
Flowers102   & 0.542 $\pm$ 0.030 & 0.606 $\pm$ 0.027 & \textbf{0.997 $\pm$ 0.002} \\
Food101      & 0.741 $\pm$ 0.007 & 0.987 $\pm$ 0.014 & \textbf{0.999 $\pm$ 0.000} \\
FGVCAircraft & 0.514 $\pm$ 0.058 & 0.521 $\pm$ 0.005 & \textbf{1.000 $\pm$ 0.000} \\
SUN397       & 0.720 $\pm$ 0.030 & 0.888 $\pm$ 0.040 & \textbf{0.940 $\pm$ 0.009} \\
DTD          & 0.377 $\pm$ 0.142 & 0.348 $\pm$ 0.043 & \textbf{1.000 $\pm$ 0.000} \\
EuroSAT      & 0.490 $\pm$ 0.147 & 0.684 $\pm$ 0.106 & \textbf{1.000 $\pm$ 0.000} \\
UCF101       & 0.550 $\pm$ 0.070 & 0.561 $\pm$ 0.036 & \textbf{1.000 $\pm$ 0.000} \\
\midrule
Average      & 0.600             & 0.719             & \textbf{0.993} \\
\bottomrule
\end{tabular}
\end{table}

\subsection{Gains within the collapsed regime}
MMA+DMC still improves on MMA+\mergetune{} (by $0.58$~HM on average over 11
datasets). If these gains came from the corridor, the datasets that retain some
separation should gain most. Table~\ref{tab:mma_gap_decomposition} does not bear
this out: SUN397, the only dataset below $0.99$, gains $+0.42$~HM, while the three
largest gains (DTD $+1.55$, UCF101 $+1.52$, Flowers102 $+1.15$) occur at
$\cos \geq 0.997$, and $\cos$ and gain are essentially uncorrelated
(Pearson $r = 0.14$). Because interpolation cannot change the classifier when the
endpoints coincide, we attribute these gains to the different training objective
of the (shared) endpoint rather than to the corridor. The collapse accounts for
the scope limit, not for these residual gains: on every dataset MMA+DMC stays
below published MMA (Table~\ref{table:hm_main}).

\begin{table}[t]
\centering

\small
\setlength{\tabcolsep}{6pt}
\renewcommand{\arraystretch}{1.10}
\caption{Per-dataset gap decomposition for MMA + DMC: endpoint cosine
$\cos(f_{\text{gen}}, f_{\text{spec}})$ (Table~\ref{tab:endpoint_collapse}) and
HM of MMA + \mergetune{} and MMA + DMC (mean over 3 seeds, Table~\ref{table:hm_main}).
ImageNet has no entry in Table~\ref{tab:endpoint_collapse} and is omitted; the
average is over the 10 datasets shown (the 11-dataset averages are $72.58$ and
$73.16$). The gain does not track the residual separation (Pearson $r = 0.14$
between $\cos$ and gain): the largest gains occur on DTD, UCF101 and Flowers102,
where $\cos \geq 0.997$.}
\label{tab:mma_gap_decomposition}
\begin{tabular}{lcccc}
\toprule
Dataset & $\cos(f_{\text{gen}}, f_{\text{spec}})$ & MMA + \mergetune{} HM & MMA + DMC HM & Gain \\
\midrule
Caltech101   & 1.000 & 95.75 & 95.74 & $-0.01$ \\
OxfordPets   & 1.000 & 93.22 & 93.38 & $+0.16$ \\
StanfordCars & 0.999 & 70.16 & 70.51 & $+0.35$ \\
Flowers102   & 0.997 & 75.22 & 76.37 & $+1.15$ \\
Food101      & 0.999 & 90.18 & 90.33 & $+0.15$ \\
FGVCAircraft & 1.000 & 32.72 & 33.10 & $+0.38$ \\
SUN397       & 0.940 & 76.70 & 77.12 & $+0.42$ \\
DTD          & 1.000 & 58.07 & 59.62 & $+1.55$ \\
EuroSAT      & 1.000 & 59.57 & 60.57 & $+1.00$ \\
UCF101       & 1.000 & 73.59 & 75.11 & $+1.52$ \\
\midrule
Average (10 datasets) & 0.993 & 72.52 & 73.19 & $+0.67$ \\
\bottomrule
\end{tabular}
\end{table}

\subsection{Intervening on capacity}
The condition predicts that giving the generalization endpoint more capacity
should break the collapse. We re-ran MMA+DMC on Flowers102 and DTD with a larger
generalization-adapter scale ($10^{-2}$ and $10^{-1}$, up from $10^{-3}$) and
with a fresh Kaiming initialization of that adapter
(Table~\ref{tab:mma_adapter_capacity}). Capacity does govern whether the
endpoints \emph{can} diverge: $\cos$ falls to $0.95$ at scale $10^{-2}$ and to
$0.78$ at $10^{-1}$. Moderate separation helps where it is achieved without
overfitting (Flowers102 HM $+2.81$ at scale $10^{-2}$; DTD unchanged within
noise), but more separation is not better: at scale $10^{-1}$ the enlarged
endpoint overfits the base classes (Flowers102 Base $93.16$, Novel $61.09$) and
DTD loses $11.3\%$ relative HM. Fresh initialization does not break the collapse
at all, and no configuration we tested recovers MMA-alone. Endpoint separation is
therefore necessary but not sufficient: the endpoints must be distinct \emph{and}
each must remain a good solution to its own objective.

\begin{table}[t]
\centering

\small
\setlength{\tabcolsep}{5pt}
\renewcommand{\arraystretch}{1.10}
\caption{MMA adapter-capacity ablation for DMC. We test two interventions
against the canonical MMA+DMC endpoint collapse
($\cos(f_{\text{gen}}, f_{\text{spec}}) \approx 1.00$): Config A enlarges the
generalization-adapter scale from $10^{-3}$ to $10^{-2}$ or $10^{-1}$;
Config B re-initializes the generalization adapter with a fresh Kaiming draw
instead of copying the base adapter. HM is base-to-novel harmonic mean at
$\alpha = 0.20$; $\cos$ is $\cos(f_{\text{gen}}, f_{\text{spec}})$ at end of
training; both reported as mean $\pm$ sample std over 3 seeds. $\cos \approx 1$
marks endpoint collapse. MMA-alone is the published number; MMA + \mergetune{}
is the mean over 3 seeds from Table~\ref{table:hm_main}. A scale of $10^{-2}$
partly separates the endpoints ($\cos \approx 0.95$) and raises Flowers102 HM
by $2.81$; a scale of $10^{-1}$ separates them further ($\cos \approx 0.78$)
but the enlarged endpoint overfits the base classes and HM drops. Fresh
initialization does not break the collapse. No intervention closes the gap to
MMA-alone.}
\label{tab:mma_adapter_capacity}
\begin{tabular}{lcccc}
\toprule
 & \multicolumn{2}{c}{Flowers102} & \multicolumn{2}{c}{DTD} \\
\cmidrule(lr){2-3}\cmidrule(lr){4-5}
Method & HM & $\cos$ & HM & $\cos$ \\
\midrule
MMA-alone (published)          & 85.48             & \textemdash        & 73.38             & \textemdash        \\
MMA + \mergetune{}             & 75.22             & \textemdash        & 58.07             & \textemdash        \\
MMA + DMC (canonical)          & 76.37 $\pm$ 0.23  & 0.997 $\pm$ 0.002  & 59.62 $\pm$ 0.13  & 1.000 $\pm$ 0.000  \\
\midrule
\multicolumn{5}{l}{\emph{Config A: enlarge generalization-adapter scale}} \\
\quad scale $=10^{-2}$         & 79.18 $\pm$ 1.45  & 0.951 $\pm$ 0.004  & 59.92 $\pm$ 1.50  & 0.980 $\pm$ 0.002  \\
\quad scale $=10^{-1}$         & 73.78 $\pm$ 1.75  & 0.780 $\pm$ 0.017  & 52.90 $\pm$ 1.31  & 0.795 $\pm$ 0.017  \\
\midrule
\multicolumn{5}{l}{\emph{Config B: fresh Kaiming init for generalization adapter}} \\
\quad fresh init               & 73.71 $\pm$ 0.16  & 0.999 $\pm$ 0.000  & 57.49 $\pm$ 0.12  & 1.000 $\pm$ 0.000  \\
\bottomrule
\end{tabular}
\end{table}

\subsection{A falsification test}
A stronger prediction is that on a collapsed base the DMC components should be
inert or harmful, reversing the pattern of Table~\ref{tab:ablation_R}. We
repeated that component ablation with MMA as the base method, with the same
seeds and code paths (Table~\ref{tab:mma_component_ablation}; the two \mergetune{} rows use $\beta = 1$). The outcome matches the
prediction: every $\Delta$ of $\mathcal{R}$ or VA that is positive on CoOp is zero or negative on MMA. Adding $\mathcal{R}$ lowers HM in seven of eight cases (from $-0.04$ to $-3.25$~HM; $0.00$ for \mergetune{} on Caltech101), and VA changes HM by at most $0.06$. The best MMA row on all four datasets is DMC
without $\mathcal{R}$, and that row still sits $6.75$~HM below published MMA on
Flowers102. Because the deltas reverse sign rather than merely shrink, the
result favors the endpoint-separation explanation over a generic ``MMA is harder
to improve'' explanation.

\subsection{Beyond prompts and adapters}
To check that the condition concerns separation rather than prompts
specifically, we implemented LoRA on the text encoder ($Q$, $K$, $V$
projections, all 12 layers, ranks 8 and 16) as a third parameterization, on two
datasets (Table~\ref{tab:lora_dmc}). On Flowers102, $\cos = 0.45$ at both ranks and LoRA+DMC gains $+10.3$
to $+11.3$~HM over LoRA alone; on Caltech101, $\cos = 0.86$--$0.91$ and DMC adds
only $+0.3$ to $+0.4$. Ordering all measurements by $\cos$: LoRA Flowers102
$0.45$ ($+10$--$11$), CoOp $0.60$ ($+6.40$), KgCoOp $0.72$ ($+0.05$), LoRA
Caltech101 $0.86$--$0.91$ ($+0.3$--$0.4$) and MMA $0.99$ ($-6.56$), where the
prompt and adapter numbers are 11-dataset average changes relative to the
published base method. Clear separation goes with large gains and near-total
collapse with the one large loss; in between (KgCoOp, LoRA Caltech101) the gains
are small and not strictly ordered. We therefore use $\cos < 0.9$ at convergence
as a coarse screening heuristic, not as a predictor of the gain. Three caveats
apply: the LoRA evidence covers two datasets; our LoRA baseline is under-tuned,
so the LoRA $\Delta$ values overstate what a well-tuned LoRA would leave on the
table (the $\cos$ measurement is unaffected); and the condition is symmetric
between the two encoders, so vision-side LoRA or adapters of comparable capacity
should also satisfy it, which we have not verified.

\begin{table}[!htbp]
\centering

\small
\setlength{\tabcolsep}{4pt}
\renewcommand{\arraystretch}{1.05}
\caption{Component ablation with MMA as the base method (HM, mean $\pm$ sample std over 3 seeds; DMC rows at $\alpha = 0.20$). Same components as Table~\ref{tab:ablation_R}; the two \mergetune{} rows use $\beta = 1$, the DMC rows $\beta = 4$. Bottom row: published MMA (Table~\ref{table:hm_main}).}
\label{tab:mma_component_ablation}
\resizebox{\linewidth}{!}{%
\begin{tabular}{lcccc}
\toprule
Method & Flowers102 & Caltech101 & StanfordCars & EuroSAT \\
\midrule
MMA+\mergetune{} w/o $\mathcal{R}$ & 76.63 $\pm$ 0.64 & 95.75 $\pm$ 0.16 & 70.71 $\pm$ 0.16 & 61.32 $\pm$ 0.28 \\
MMA+\mergetune{}                   & 75.22 $\pm$ 0.40 & 95.75 $\pm$ 0.16 & 70.16 $\pm$ 0.24 & 59.57 $\pm$ 0.48 \\
MMA+DMC w/o $\mathcal{R}$          & \textbf{78.73 $\pm$ 0.46} & \textbf{95.77 $\pm$ 0.15} & \textbf{71.62 $\pm$ 0.25} & \textbf{63.78 $\pm$ 0.34} \\
MMA+DMC w/o VA                     & 76.38 $\pm$ 0.13 & 95.73 $\pm$ 0.17 & 70.57 $\pm$ 0.13 & 60.53 $\pm$ 0.49 \\
MMA+DMC (full)                     & 76.37 $\pm$ 0.23 & 95.74 $\pm$ 0.16 & 70.51 $\pm$ 0.04 & 60.57 $\pm$ 0.45 \\
\midrule
\textit{MMA (published)}           & \textit{85.48} & \textit{96.15} & \textit{75.70} & \textit{83.87} \\
\bottomrule
\end{tabular}}
\end{table}

\begin{table}[!htbp]
\centering

\small
\setlength{\tabcolsep}{5pt}
\renewcommand{\arraystretch}{1.05}
\caption{LoRA on the text encoder ($Q$, $K$, $V$, all 12 layers) with and without DMC ($\lambda_{\text{gen}} = 8$, $\beta = 4$, no VA, 50 epochs). HM mean $\pm$ sample std over 3 seeds; DMC at $\alpha = 0.20$; $\cos = \cos(f_{\mathrm{gen}}, f_{\mathrm{spec}})$. The LoRA baseline is under-tuned (Flowers102 base accuracy $69.5$).}
\label{tab:lora_dmc}
\begin{tabular}{llcccc}
\toprule
Dataset & Rank & LoRA & LoRA+DMC & $\Delta$ & $\cos$ \\
\midrule
Flowers102 & 8  & 73.03 $\pm$ 0.03 & 84.37 $\pm$ 0.33 & $+11.34$ & 0.454 $\pm$ 0.010 \\
Flowers102 & 16 & 73.18 $\pm$ 0.11 & 83.51 $\pm$ 0.41 & $+10.33$ & 0.454 $\pm$ 0.015 \\
Caltech101 & 8  & 95.69 $\pm$ 0.00 & 96.01 $\pm$ 0.48 & $+0.32$  & 0.861 $\pm$ 0.146 \\
Caltech101 & 16 & 95.68 $\pm$ 0.05 & 96.06 $\pm$ 0.30 & $+0.38$  & 0.908 $\pm$ 0.092 \\
\bottomrule
\end{tabular}
\end{table}

\FloatBarrier

\FloatBarrier

\section{Class-Permutation Invariance: Extended Analysis}
\label{app:cpi}
\label{sec:cpi_ablation}

This section tests which candidate regularizers satisfy class-permutation invariance (CPI) and whether the size of a violation relates to how the regularizer behaves.
Table~\ref{tab:cpi} reports $\delta_{\max}$ computed over $10{,}000$ random class permutations on Caltech101~($C = 50$).
CPI terms (cosine score, Visual Anchor, ZSDD, and the Drift-Penalty criterion) return $\delta_{\max} \leq 2.4 \times 10^{-7}$ (float32 round-off), consistent with exact floating-point invariance.
Non-CPI regularizers return values orders of magnitude larger: Fisher-weighted cosine yields $\delta_{\max} = 0.014$ and EWC yields $\delta_{\max} = 1.07$.
Fisher-weighted cosine does not break the single-prompt ceiling (Appendix~\ref{app:loss_modifications}). We did not train feature-space EWC as a regularizer, so we make no claim about its accuracy; the $\delta$ values only show that both non-CPI regularizers violate the symmetry by orders of magnitude more than the CPI ones.
ZSDD is CPI ($\delta_{\max} = 2.4 \times 10^{-7}$) but works only when the zero-shot teacher is accurate; CPI is a desirable symmetry property, not a sufficient one.

\begin{table}[!htbp]
\centering

\caption{Class-permutation invariance~(CPI) over $10{,}000$ random
permutations on Caltech101~($C = 50$). $\Delta$HM: change of the best single-prompt configuration over the cosine baseline on Caltech101 / Flowers102 (Table~\ref{tab:loss_mods}); ---, symmetry test only, not trained. $^\ddagger$Drift-Penalty is an $\alpha$-selection criterion (Appendix~\ref{app:alpha_selection}), not a training regularizer.}
\label{tab:cpi}
\small
\setlength{\tabcolsep}{5pt}
\begin{tabular}{lcccl}
\toprule
\textbf{Regularizer}
    & \textbf{CPI}
    & $\boldsymbol{\delta_{\max}}$
    & $\boldsymbol{\delta_{\mathrm{mean}}}$
    & \textbf{$\Delta$HM (Cal / Flo)} \\
\midrule
Cosine score~\citep{wang2026mergetune}
    & \checkmark
    & $1.2 \times 10^{-7}$
    & $4 \times 10^{-8}$
    & baseline \\
Visual Anchor~(ours)
    & \checkmark
    & $6 \times 10^{-8}$
    & $2 \times 10^{-8}$
    & $-0.04$ / $+0.36$ \\
Drift-Penalty~(ours)$^\ddagger$
    & \checkmark
    & $1.2 \times 10^{-7}$
    & $4 \times 10^{-8}$
    & --- \\
ZSDD
    & \checkmark
    & $2.4 \times 10^{-7}$
    & $7 \times 10^{-8}$
    & $+0.40$ / $-2.37$ \\
\midrule
Fisher-weighted cosine
    & $\times$
    & $1.4 \times 10^{-2}$
    & $3.3 \times 10^{-3}$
    & $+0.21$ / $+0.22$ \\
EWC~(feature space)
    & $\times$
    & $1.07$
    & $0.297$
    & --- \\
\bottomrule
\end{tabular}
\end{table}

\subsection{Informal justification}
The formal statement is Proposition~\ref{prop:cpi_fano}; here we give the intuition.

A regularizer $\mathcal{R}(f, f_{w_1})$ is estimated on base
classes $\mathcal{Y}_{\mathrm{base}}$ and must benefit
generalization to novel classes
$\mathcal{Y}_{\mathrm{novel}}$, with
$\mathcal{Y}_{\mathrm{base}} \cap \mathcal{Y}_{\mathrm{novel}} = \emptyset$.

If $\mathcal{R}$ is not CPI, it assigns asymmetric importance to
specific base-class directions. This asymmetry depends on base-class
statistics (curvature, frequency, confidence) that have no defined
value for novel-class directions. The regularizer therefore preserves
a property specific to $\mathcal{Y}_{\mathrm{base}}$ that may
conflict with novel-class preservation.

If $\mathcal{R}$ is CPI, it preserves a property that holds for any
permutation of the class set, and therefore any subset of the same
size. Since $|\mathcal{Y}_{\mathrm{novel}}| \approx
|\mathcal{Y}_{\mathrm{base}}|$ (equal, or differing by one class) in standard protocols, a CPI
regularizer preserves a property equally valid for
$\mathcal{Y}_{\mathrm{novel}}$.

\subsection{Interpreting the magnitude of \texorpdfstring{$\delta$}{delta}}

The $\delta_{\max}$ value measures how much the regularizer changes
under class relabeling. A large $\delta_{\max}$ indicates strong
class-specific encoding. When applied to novel classes, this
class-specific information acts as noise: the regularizer pulls the
prompt toward geometry optimal for specific base classes, away from
the geometry needed for novel classes. This is a heuristic account; we have not tested whether larger $\delta$ predicts a larger accuracy loss.

EWC~($\delta_{\max} = 1.07$) computes its penalty from the full
gradient of the classification loss, encoding the specific direction
each base class contributes. Fisher-weighted cosine
($\delta_{\max} = 0.014$) uses only the Fisher diagonal as a
per-class weight, less specific but still asymmetric. The cosine
score ($\delta_{\max} = 1.2 \times 10^{-7}$) uses a uniform weight
of $1/C$ for every class, achieving exact CPI.

\subsection{The ZSDD special case}

ZSDD is CPI ($\delta_{\max} = 2.4 \times 10^{-7}$) because its
softmax treats all class features symmetrically. However, ZSDD fails
on fine-grained datasets where zero-shot CLIP has low accuracy.
On Flowers102 (zero-shot base $\approx 72\%$), the teacher
distribution has high entropy, teaching the student to be similarly
uncertain and capping base-class confidence. On Caltech101
(zero-shot base $\approx 97\%$), the teacher is sharp and ZSDD
produces a small gain.

CPI is therefore not sufficient on its own. The cosine score avoids the teacher-quality issue by anchoring to
geometric structure rather than output predictions. Visual Anchor is also a distillation-style term and shares this caveat in principle; in DMC it acts only on $\mathbf{c}_{\mathrm{spec}}$, alongside CE, and its measured effect is small (Appendix~\ref{app:ablation_tables}).

\subsection{Reproducibility}

The CPI test requires no training. Given a text encoder $T$ and
zero-shot features $f_{w_1} \in \mathbb{R}^{C \times D}$, compute:
\begin{equation}
\delta_{\max}(\mathcal{R}) = \max_{\Pi \in S}
    \left|\mathcal{R}(\Pi f, \Pi f_{w_1})
    - \mathcal{R}(f, f_{w_1})\right|,
\end{equation}
where $S$ is a random sample of $10{,}000$ permutation matrices from the symmetric group of degree $C$,
and $f = \ell_2(T(\mathbf{c}))$ is the current text-feature matrix.
For $C = 50$ classes, this takes under one second.

\FloatBarrier

\subsection{Applicability and limitations of the CPI bound}
\label{app:cpi_applicability}

Proposition~\ref{prop:cpi_fano} is a capacity-allocation argument that holds
under four assumptions (Appendix~\ref{app:proof_cpi}): distinct minimizers~(0),
equal prompt entropy~(i), conditional label independence~(ii) and baseline
efficiency~(iii). Each is a modeling choice that can fail in practice. We state
what each failure does to the bound and what the evidence still supports.

\paragraph{(0) Distinct minimizers.}
A CPI-violating penalty changes the objective under relabeling, but two
different objectives can share a minimizer; if they do, $I(\theta;
Y_{\mathrm{base}})$ can be zero and the bound is vacuous rather than wrong.

\paragraph{(i) Equal prompt entropy $H(\theta) = H(\theta^*) = \kappa$.}
This normalizes the comparison so that an accuracy gap reflects \emph{how}
capacity is spent rather than \emph{how much}. It fails when the two algorithms
use different regularization strengths. When entropies differ, the per-prompt
inequality $I(\theta; Y_{\mathrm{novel}}) \leq H(\theta) - I(\theta;
Y_{\mathrm{base}})$ still holds, but Eq.~\eqref{eq:info_gap} acquires
the correction term $H(\theta) - H(\theta^*)$ on its right-hand side.

\paragraph{(ii) Conditional independence $Y_{\mathrm{base}} \perp Y_{\mathrm{novel}} \mid \theta$.}
Natural when base and novel labels share no fine-grained attributes. It fails on
splits where both halves draw from the same higher-level taxa, since capacity
that helps novel classes then also helps base classes and the additive
decomposition underlying the capacity bound becomes approximate.

\paragraph{(iii) Baseline efficiency.}
We assume the CPI-satisfying baseline converts its capacity into novel-class
information at the Fano ceiling. Real algorithms fall short of it; then
Eq.~\eqref{eq:fano_main} is no longer guaranteed and can overstate the gap, and
the proposition indicates the direction of the effect rather than bounding its
size.

\paragraph{Uniform label priors.}
Uniform priors are part of the setup rather than an assumption on $A$; they give
the cleanest Fano expression. Long-tailed splits reduce the effective label
entropy and loosen the bound by the entropy gap $\log|\mathcal{Y}| - H(Y)$.

\paragraph{What survives.}
Empirically, the non-CPI regularizers violate the symmetry by orders of
magnitude more than the CPI ones (Appendix~\ref{app:cpi}), but the one we trained
(Fisher-weighted cosine) is not clearly harmful in its best configuration
(Table~\ref{tab:loss_mods}), so the evidence supports the direction of the
proposition, not a measured gap. We therefore present
Proposition~\ref{prop:cpi_fano} as an idealized bound that motivates the
choice of CPI regularizers and predicts the direction of failure, and we do not
rely on its constant.

\FloatBarrier

\FloatBarrier

\section{Training and Inference Protocol}
\label{app:implementation}

This section specifies how DMC is trained and evaluated and what it costs. The LMC corridor is a degree-1 (linear) path in $\ell_2$-normalized
text-feature space; no polynomial or B\'ezier basis is used.

\paragraph{Training-time $\alpha$ sampling.}
DMC is evaluated at the checkpoint with the best base-class validation accuracy; \mergetune{} is evaluated at its final prompt. At each gradient step, the LMC path term is evaluated at a single $\alpha$ sampled uniformly from $[0, 1]$, with the endpoints $\alpha \in \{0, 1\}$ included in the sampling support. The expectation over $\alpha$ enforces low classification loss across the full interpolation range, rather than at any fixed mixing weight.

\paragraph{Inference cost.}
The two prompts are needed only to compute the text features $f_{\mathrm{gen}}$ and $f_{\mathrm{spec}}$, which is performed once per dataset. At per-image inference, only the precomputed interpolated feature $\widetilde{f}(\alpha) = \ell_2\bigl(f_{\mathrm{gen}} + \alpha\,(f_{\mathrm{spec}} - f_{\mathrm{gen}})\bigr)$ is used. Per-image cost is therefore identical to that of a single-prompt method.

\paragraph{Compute cost.}
We profiled one full DMC training cell (CoOp+DMC, Flowers102, seed 1, 51 base
classes, 100 epochs, batch 128) with \texttt{/usr/bin/time~-v} and a 2\,s
\texttt{nvidia-smi} sampler on a single NVIDIA A100-PCIE-40GB. Wall-clock is
15\,min 34\,s (9.34\,s per epoch, 0.75\,s per iteration), peak GPU memory is
5949\,MiB ($14.5\%$ of the card), and the trainable parameter count is 4096 (two
prompts of $N = 4$ tokens at $D = 512$). Extrapolating this cell to all
11 datasets, 3 baselines and 3 seeds gives roughly 26 A100-hours, an
underestimate for larger datasets such as ImageNet. At test time the two
prompts are collapsed into one interpolated text feature per class, so
per-image inference matches single-prompt CoOp, KgCoOp and MMA exactly. DMC also
inherits \mergetune{}'s optimizer, schedule and epoch count unchanged, so the only added
cost is the second prompt. The profiled cell uses batch size 128, as in the component ablation; the DMC runs of Table~\ref{table:hm_main} use batch size 32, except CoOp and KgCoOp on ImageNet (128).

\paragraph{$\alpha$ selection on the development set.}
The fixed inference coefficient $\alpha = 0.20$ was selected on a two-dataset development set (Flowers102 and Caltech101) and evaluated on the remaining eight held-out datasets (ImageNet excluded). On the held-out set, $\alpha = 0.20$ achieves a mean HM gap to the per-dataset oracle of $-0.18$ (worst case $-0.28$). The method is robust to this choice: every $\alpha \in [0.15, 0.25]$ yields a mean oracle gap of at most $0.26$~HM. A per-criterion comparison against eight alternative selection rules is in Appendix~\ref{app:alpha_selection}.

\FloatBarrier

\FloatBarrier

\section{Interpolation Coefficient Selection}
\label{app:alpha_selection}

DMC produces a family of classifiers parameterized by
$\alpha \in [0,1]$. This section asks how to pick $\alpha$ without novel-class data, and how much the fixed $\alpha = 0.20$, chosen on Flowers102 and Caltech101, costs elsewhere. Here we analyze the $\alpha$ selection problem
and evaluate eight automatic criteria.

\paragraph{Problem statement.} We seek $\alpha^*$ maximizing
$\mathrm{HM} = 2 \cdot \mathrm{Base} \cdot \mathrm{Novel} /
(\mathrm{Base} + \mathrm{Novel})$,
but novel-class data is unavailable at selection time. Any
criterion evaluated on base classes tends to favor high $\alpha$
(more specialization), even when the HM-optimal $\alpha$ is low.

\paragraph{Criteria evaluated.}

We evaluate eight criteria on up to four datasets (Flowers102, Caltech101,
StanfordCars, EuroSAT), using CoOp + DMC + VA
checkpoints; the five criteria with large gaps on the two development datasets were evaluated only there.
\begin{itemize}
\item  \textbf{Val-Acc. }Selects $\alpha$ with highest base-class validation accuracy.
Fails because val accuracy increases monotonically with $\alpha$,
selecting $\alpha \approx 1.0$.

\item  \textbf{Proxy-HM. }Splits base classes into two random halves (10 splits, averaged),
treating one as proxy novel classes. Selects $\alpha$ maximizing
the proxy HM. Fails identically to Val-Acc: both halves were seen
during training, so the proxy does not simulate the base-novel
boundary.

\item  \textbf{Drift-Penalty. }Balances validation accuracy against cosine drift from $f_{w_1}$:
$\alpha^* = \arg\max_\alpha [
\overline{\mathrm{Acc}}_{\mathrm{val}}(\alpha) -
\mu \cdot \overline{d}_{\cos}(\alpha)]$,
with both terms min-max normalized to $[0,1]$.
Requires tuning $\mu$ per dataset.

\item  \textbf{Gram-Align.} Selects $\alpha$ maximizing Pearson correlation between the
class-prototype Gram matrices at $\alpha$ and at $f_{w_1}$.

\item  \textbf{Effective Rank.} Computes the effective rank of the prototype matrix
$\mathbf{F}(\alpha) \in \mathbb{R}^{C \times d}$ via
$\mathrm{erank} = \exp(-\sum_i p_i \log p_i)$.
Fails: effective rank \emph{increases} with $\alpha$ because CE
spreads prototypes apart, always selecting $\alpha = 1.0$.

\item  \textbf{Neighborhood Preservation.} For each class, measures the fraction of $K$-nearest neighbors
(in $f_{w_1}$ space) preserved at $\alpha$. Selects highest $\alpha$
with NP above a threshold. Fails: NP starts below 1.0 at
$\alpha=0$, so threshold criteria select $\alpha = 0$.

\item  \textbf{KL Divergence.} $\alpha^* = \arg\max_\alpha [
-\mathrm{CE}_{\mathrm{val}}(\alpha) - \nu \cdot \mathrm{KL}(q_\alpha \| q_{w_1})]$, with temperature $\tau$ for the softmax. Both terms in nats.
KL grows much faster than CE, making the balance sensitive to $\nu$ and $\tau$.

\item  \textbf{Pareto Knee.}
Plots the Pareto front $(-\mathrm{CE}_{\mathrm{val}}(\alpha),\; \cos(\mathbf{f}(\alpha), f_{w_1}))$ and selects the point of maximum curvature.
Hyperparameter-free, but overshoots on gentle fronts.

\end{itemize}

\paragraph{Results. }Table~\ref{tab:alpha_selection_full} compares all criteria. Among adaptive criteria, Drift-Penalty ($\mu{=}4$) performs best
($-0.22$ avg) but requires dataset-dependent $\mu$. KL-based selection is principled but sensitive to $\nu$/$\tau$. No adaptive criterion beats the fixed $\alpha = 0.20$ ($-0.19$ avg). The Pareto knee overshoots on gentle fronts (StanfordCars $-1.16$, EuroSAT $-2.07$).

\begin{table}[!htbp]
\centering

\caption{$\alpha$ selection comparison (mean $\Delta$ to oracle HM,
3 seeds). Fixed $\alpha = 0.20$ achieves the best average gap. ---: not evaluated (large gap already on the development datasets).}
\label{tab:alpha_selection_full}
\small
\setlength{\tabcolsep}{4pt}
\begin{tabular}{lcccc|c}
\toprule
\textbf{Criterion} & \textbf{Flowers102} & \textbf{Caltech101} & \textbf{StanfordCars} & \textbf{EuroSAT} & \textbf{Avg} \\
\midrule
Val-Acc & $-3.50$ & $-0.95$ & --- & --- & --- \\
Proxy-HM & $-5.29$ & $-1.05$ & --- & --- & --- \\
Gram-Align & $-2.29$ & $-0.24$ & --- & --- & --- \\
EffRank-Max & $-9.26$ & $-1.54$ & --- & --- & --- \\
NP$\geq$0.90 & $-2.29$ & $-0.28$ & --- & --- & --- \\
\midrule
Drift-Pen ($\mu{=}4$) & $-0.25$ & $-0.30$ & $-0.18$ & $-0.14$ & $-0.22$ \\
KL ($\tau{=}2$, $\nu{=}0.5$) & $-0.33$ & $-0.09$ & $-0.87$ & $-0.39$ & $-0.42$ \\
Pareto Knee & $-0.27$ & $-0.38$ & $-1.16$ & $-2.07$ & $-0.97$ \\
\midrule
Fixed $\alpha{=}0.10$ & $-1.01$ & $-0.19$ & $-0.30$ & $-0.14$ & $-0.41$ \\
Fixed $\alpha{=}0.15$ & $-0.44$ & $-0.08$ & $-0.20$ & $-0.18$ & $-0.22$ \\
\textbf{Fixed} $\boldsymbol{\alpha{=}0.20}$ & $\boldsymbol{-0.17}$ & $-0.19$ & $\boldsymbol{-0.14}$ & $-0.28$ & $\boldsymbol{-0.19}$ \\
\bottomrule
\end{tabular}
\end{table}

\subsection{Robustness of the fixed \texorpdfstring{$\alpha$}{alpha} across datasets and dev-set choices}
\label{app:alpha_robustness}

The analysis above evaluates $\alpha$ on four datasets. Two further experiments
extend it to all 33 (baseline $\times$ dataset) cells and, for CoOp+DMC, to every
alternative two-dataset development set.

\paragraph{Per-dataset oracle $\alpha^\star$.}
We swept $\alpha$ on a 21-point grid and computed the oracle of the
seed-averaged curve per cell (Table~\ref{tab:alpha_oracle_per_dataset}). The
median oracle is $\alpha^\star = 0.25$ on both prompt-tuning bases, one grid step
from the fixed $0.20$, and the mean HM sacrificed by using $0.20$ instead of the
per-dataset oracle is $0.08$ (CoOp) and $0.17$ (KgCoOp). MMA's median
$\alpha^\star = 0.95$ is a symptom of the endpoint collapse of
Appendix~\ref{app:endpoint_collapse} --- on a degenerate corridor the choice of
$\alpha^\star$ is poorly determined (mean gap $0.26$~HM, up to $0.66$ on ImageNet) --- not of a failure of the fixed choice.

\begin{table}[t]
\centering

\small
\setlength{\tabcolsep}{3pt}
\caption{Per-dataset oracle-best $\alpha^\star$ for DMC across three baselines. HM is averaged over 3 seeds. $\Delta$ is $\mathrm{HM}(\alpha^\star)-\mathrm{HM}(0.20)$. Bottom rows: median $\alpha^\star$ across the 11 datasets and range $[\min,\max]$. For CoOp and KgCoOp the median oracle $\alpha^\star$ lies one grid step from the fixed $\alpha=0.20$; for MMA the corridor is degenerate (Appendix~\ref{app:endpoint_collapse}), so $\alpha^\star$ is poorly determined.}
\label{tab:alpha_oracle_per_dataset}
\resizebox{\linewidth}{!}{%
\begin{tabular}{lcccccccccccc}
\toprule
Dataset & \multicolumn{4}{c}{CoOp+DMC} & \multicolumn{4}{c}{KgCoOp+DMC} & \multicolumn{4}{c}{MMA+DMC} \\
\cmidrule(lr){2-5} \cmidrule(lr){6-9} \cmidrule(lr){10-13}
 & $\alpha^\star$ & HM$^\star$ & HM$_{.20}$ & $\Delta$ & $\alpha^\star$ & HM$^\star$ & HM$_{.20}$ & $\Delta$ & $\alpha^\star$ & HM$^\star$ & HM$_{.20}$ & $\Delta$ \\
\midrule
Caltech101 & 0.15 & 96.52 & 96.41 & +0.11 & 0.00 & 96.26 & 96.11 & +0.15 & 0.00 & 95.76 & 95.74 & +0.02 \\
DTD & 0.20 & 63.54 & 63.54 & +0.00 & 0.25 & 64.51 & 64.37 & +0.14 & 0.65 & 59.68 & 59.62 & +0.06 \\
EuroSAT & 0.10 & 75.98 & 75.84 & +0.14 & 0.00 & 74.12 & 73.39 & +0.73 & 0.95 & 60.77 & 60.57 & +0.20 \\
FGVCAircraft & 0.20 & 35.24 & 35.24 & +0.00 & 0.05 & 33.33 & 33.22 & +0.11 & 1.00 & 33.21 & 33.10 & +0.10 \\
Food101 & 0.45 & 91.22 & 91.13 & +0.09 & 0.45 & 91.20 & 91.19 & +0.01 & 0.15 & 90.33 & 90.33 & +0.00 \\
ImageNet & 0.30 & 73.06 & 72.90 & +0.17 & 0.45 & 73.32 & 73.09 & +0.23 & 1.00 & 73.56 & 72.90 & +0.66 \\
Flowers102 & 0.25 & 84.49 & 84.47 & +0.02 & 0.10 & 83.72 & 83.67 & +0.05 & 0.95 & 77.01 & 76.37 & +0.64 \\
OxfordPets & 0.25 & 96.09 & 96.04 & +0.05 & 0.30 & 96.33 & 96.23 & +0.09 & 1.00 & 93.44 & 93.38 & +0.06 \\
StanfordCars & 0.25 & 75.34 & 75.21 & +0.13 & 0.25 & 75.48 & 75.39 & +0.10 & 0.90 & 70.70 & 70.51 & +0.19 \\
SUN397 & 0.25 & 78.81 & 78.75 & +0.06 & 0.25 & 78.94 & 78.85 & +0.08 & 0.80 & 77.70 & 77.12 & +0.58 \\
UCF101 & 0.15 & 80.23 & 80.08 & +0.15 & 0.15 & 79.06 & 78.91 & +0.15 & 0.95 & 75.48 & 75.11 & +0.36 \\
\midrule
Median $\alpha^\star$ & 0.25 & -- & -- & -- & 0.25 & -- & -- & -- & 0.95 & -- & -- & -- \\
Range $\alpha^\star$ & [0.10,0.45] & -- & -- & -- & [0.00,0.45] & -- & -- & -- & [0.00,1.00] & -- & -- & -- \\
Mean $\Delta$ & -- & -- & -- & +0.08 & -- & -- & -- & +0.17 & -- & -- & -- & +0.26 \\
\bottomrule
\end{tabular}}
\end{table}

\paragraph{Dev-set reshuffle.}
Our fixed $\alpha$ was chosen on Flowers102 and Caltech101, which raises the
question of how much that particular pair matters. We repeated the selection for
all $\binom{11}{2} = 55$ two-dataset development pairs and measured the held-out
loss on the remaining 9 datasets (Table~\ref{tab:alpha_devset_reshuffle}). The
selected $\alpha^\star$ lands in $[0.15, 0.25]$ for 47 of 55 pairs ($85\%$),
47 pairs keep the held-out mean gap below $0.30$~HM, and 54 of 55 stay below
$0.50$~HM. The median held-out gap is $0.15$~HM. The paper's own pair sits inside
this dominant mode with a gap of $0.09$~HM (measured against the seed-averaged oracle
and on 9 held-out datasets, so not directly comparable with the $0.18$ of
Table~\ref{tab:alpha_crossval}), so
the choice of development set does not drive the results. The worst pair, Food101 with ImageNet, selects
$\alpha^\star = 0.45$ and costs $1.19$~HM.

\begin{table}[t]
\centering
\small
\caption{Dev-set reshuffle robustness for CoOp+DMC. For each of the $\binom{11}{2}=55$ ways to pick a 2-dataset dev subset, we choose the fixed $\alpha$ that minimizes the mean gap-to-oracle on the dev pair, then measure the mean gap-to-oracle on the remaining 9 held-out datasets. The paper's dev pair is Flowers102+Caltech101. Gaps are to the oracle of the seed-averaged $\alpha$-curve (Table~\ref{tab:alpha_oracle_per_dataset}).}
\label{tab:alpha_devset_reshuffle}
\begin{tabular}{lc}
\toprule
Statistic over 55 dev pairs & Value \\
\midrule
Pairs that pick $\alpha^\star \in [0.15,0.25]$ & 47/55 (85\%) \\
Pairs that pick $\alpha^\star \le 0.10$ or $\ge 0.35$ & 2/55 (4\%) \\
Pairs with held-out mean gap $\le 0.30$ HM & 47/55 (85\%) \\
Pairs with held-out mean gap $\le 0.50$ HM & 54/55 (98\%) \\
\midrule
Mean held-out gap across 55 pairs & 0.17 HM \\
Median held-out gap across 55 pairs & 0.15 HM \\
\midrule
Best pair: EuroSAT+SUN397 ($\alpha^\star=0.20$) & 0.08 HM \\
Worst pair: Food101+ImageNet ($\alpha^\star=0.45$) & 1.19 HM \\
\bottomrule
\end{tabular}
\end{table}

\FloatBarrier

\subsection{Drift-Penalty \texorpdfstring{$\mu$}{mu} sweep on the development set}
\label{app:alpha_dev_sweep}

Table~\ref{table:alpha_select} sweeps the drift-penalty weight $\mu$
on the two development datasets (Flowers102, Caltech101; 3 seeds
each, $\lambda_{\text{gen}} = 8.0$, $\beta = 4.0$), reporting both
mean and worst-case oracle gap; these runs are separate from those of
Table~\ref{tab:alpha_selection_full}, so the numbers differ. Accuracy-based criteria fail:
validation accuracy selects $\alpha = 1.0$ on Flowers102 (pure
specialization), losing $-7.19$~HM versus oracle, because base-class
validation accuracy increases monotonically with $\alpha$. Proxy-HM (random
base-class splits used as proxy novel classes) selects the same
$\alpha$ as Val-Acc in every case, indicating that base-class
splits cannot simulate the base-novel boundary. The drift-penalty
criterion balances validation accuracy against cosine drift from
$f_{w_1}$; with $\mu = 4$, the gap shrinks to $-0.13$ on Flowers102 and
$-0.02$ on Caltech101. The fixed $\alpha = 0.20$ achieves comparable
performance with no hyperparameters.

\begin{table}[!htbp]
\centering
\caption{Drift-Penalty $\mu$ sweep on the development set
(Flowers102, Caltech101; 3 seeds, $\lambda_{\text{gen}} = 8.0$,
$\beta = 4.0$). Mean and worst-case HM gap to oracle.}
\label{table:alpha_select}
\setlength{\tabcolsep}{6pt}
\begin{tabular}{lcc|cc}
\toprule
& \multicolumn{2}{c}{\textbf{Flowers102}} & \multicolumn{2}{c}{\textbf{Caltech101}} \\
\cmidrule(lr){2-3}\cmidrule(lr){4-5}
\textbf{Criterion} & Mean $\Delta$ & Worst $\Delta$ & Mean $\Delta$ & Worst $\Delta$ \\
\midrule
Val-Acc & $-7.19$ & $-12.69$ & $-0.29$ & $-0.48$ \\
Proxy-HM & $-7.19$ & $-12.69$ & $-0.29$ & $-0.48$ \\
Gram-Align & $-1.47$ & $-1.91$ & $-0.24$ & $-0.45$ \\
ValAcc$\times$Gram & $-0.78$ & $-1.22$ & --- & --- \\
Drift-Pen ($\mu{=}1$) & $-0.78$ & $-1.22$ & $-0.16$ & $-0.48$ \\
Drift-Pen ($\mu{=}2$) & $-0.16$ & $-0.36$ & $-0.16$ & $-0.48$ \\
\textbf{Drift-Pen ($\boldsymbol{\mu{=}4}$)} & $\boldsymbol{-0.13}$ & $\boldsymbol{-0.29}$ & $\boldsymbol{-0.02}$ & $\boldsymbol{-0.06}$ \\
\bottomrule
\end{tabular}
\end{table}

\subsection{Cross-validated fixed \texorpdfstring{$\alpha$}{alpha}}

We select $\alpha$ on two development datasets (Flowers102,
Caltech101) and evaluate on 8 held-out datasets.
On the dev set, $\alpha \in \{0.20, 0.25\}$ are tied (mean HM
$= 90.44$); we choose $\alpha = 0.20$ as the more conservative
candidate. Table~\ref{tab:alpha_crossval} reports the oracle gap
on held-out datasets.

\begin{table}[!htbp]
\centering
\caption{Generalization of fixed $\alpha = 0.20$ (selected on
dev set) to 8 held-out datasets. Mean $\Delta$ to oracle HM.}
\label{tab:alpha_crossval}
\small
\setlength{\tabcolsep}{4pt}
\begin{tabular}{lccccccccc}
\toprule
$\alpha$ & Pets & Food & Cars & Aircraft & SUN & DTD & EuroSAT & UCF & \textbf{Avg} \\
\midrule
0.10 & $-0.20$ & $-0.30$ & $-0.30$ & $-0.66$ & $-0.26$ & $-1.33$ & $-0.14$ & $-0.33$ & $-0.44$ \\
0.15 & $-0.19$ & $-0.22$ & $-0.20$ & $-0.42$ & $-0.18$ & $-0.46$ & $-0.18$ & $-0.06$ & $-0.24$ \\
\textbf{0.20} & $\boldsymbol{-0.17}$ & $\boldsymbol{-0.13}$ & $\boldsymbol{-0.14}$ & $\boldsymbol{-0.22}$ & $\boldsymbol{-0.07}$ & $\boldsymbol{-0.23}$ & $\boldsymbol{-0.28}$ & $\boldsymbol{-0.21}$ & $\boldsymbol{-0.18}$ \\
0.25 & $-0.12$ & $-0.09$ & $-0.01$ & $-0.44$ & $-0.01$ & $-0.43$ & $-0.53$ & $-0.41$ & $-0.26$ \\
0.30 & $-0.15$ & $-0.08$ & $-0.15$ & $-0.93$ & $-0.02$ & $-0.54$ & $-0.77$ & $-0.86$ & $-0.44$ \\
\bottomrule
\end{tabular}
\end{table}

\begin{figure*}[!htbp]
\centering
\includegraphics[width=\textwidth]{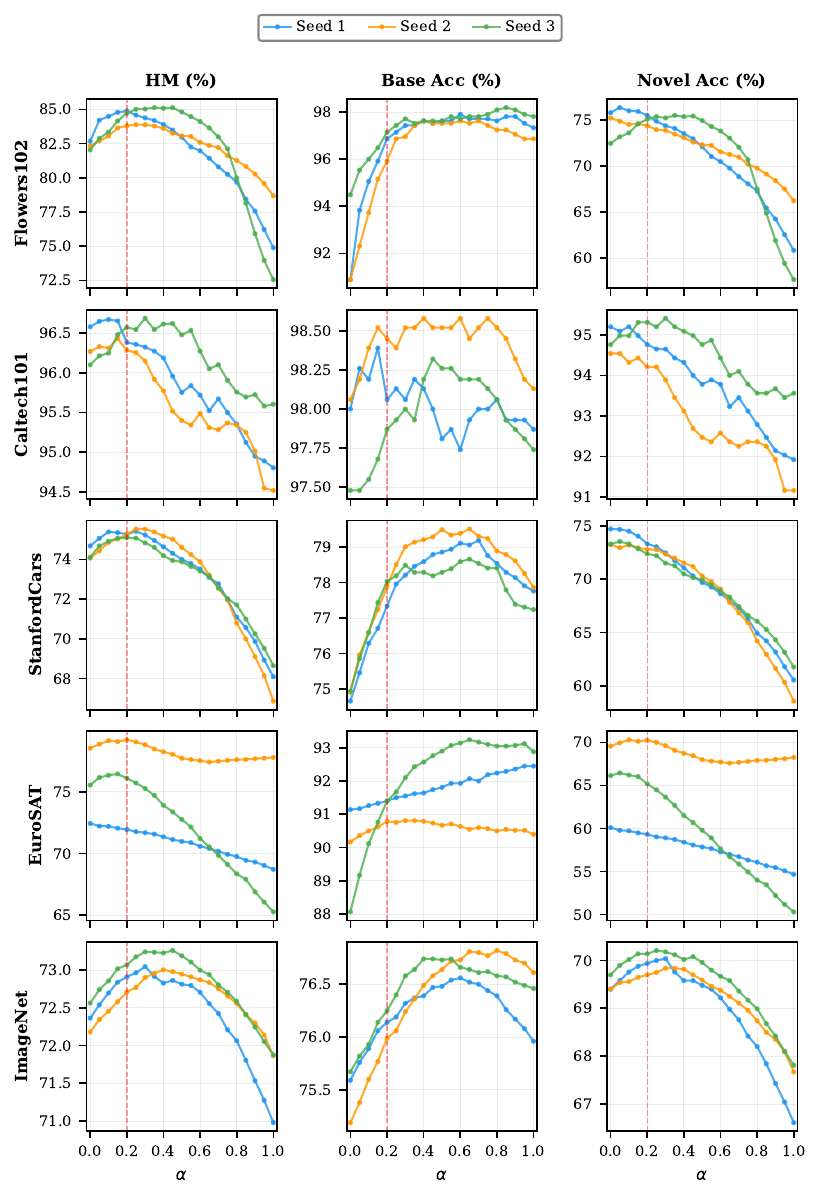}
\caption{\textbf{HM, Base accuracy, and Novel accuracy vs $\alpha$}
on four datasets. Each row is one dataset; columns
show HM, Base, and Novel accuracy. The dashed red line marks
$\alpha = 0.20$. Base accuracy increases monotonically with
$\alpha$ (the specialization prompt dominates), while Novel accuracy
decreases (generalization degrades). The HM peak reflects the
trade-off between these opposing trends and lies in
$\alpha \in [0.10, 0.35]$ across all datasets.}
\label{fig:alpha_hm}
\end{figure*}

Fixed $\alpha = 0.20$ achieves the best average gap ($-0.18$) on the eight held-out datasets. Any $\alpha \in [0.15, 0.25]$ yields
an average gap of at most $0.26$~HM. This stability is consistent with the
consistently shaped interpolation curves of DMC, whose HM peak
lies in a narrow band across datasets (Figure~\ref{fig:alpha_hm}).

\begin{figure*}[!htbp]
\centering
\includegraphics[width=\textwidth]{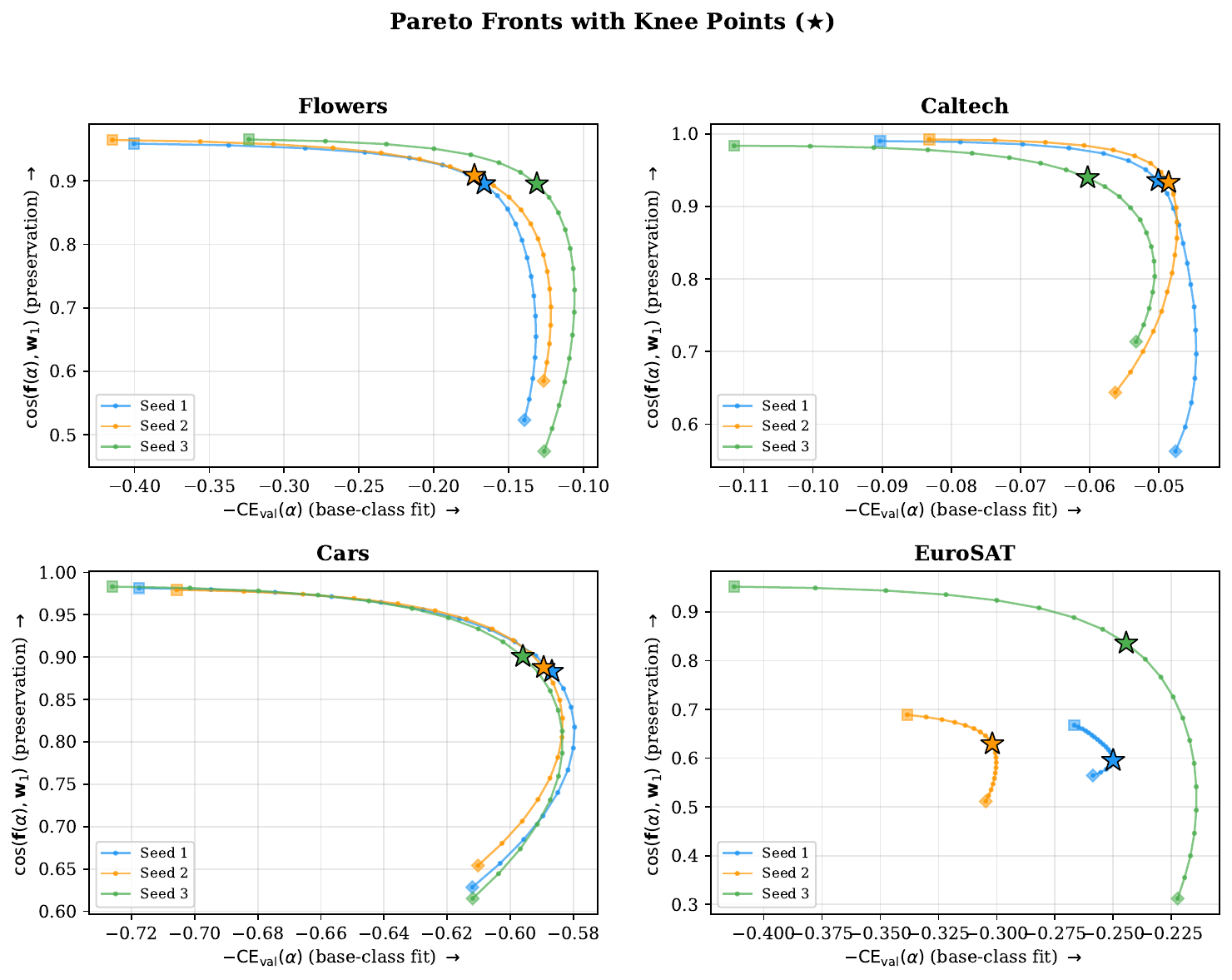}
\caption{\textbf{Pareto fronts} for $\alpha$ selection on four
datasets. Stars mark Pareto knee points. As $\alpha$
increases, features move right (better base fit) and down (more
drift from $f_{w_1}$).}
\label{fig:alpha_pareto}
\end{figure*}

\FloatBarrier

\FloatBarrier

\section{Hyperparameter Sensitivity}
\label{app:hyperparam}

DMC's two main scalar weights are the LMC path weight $\beta$ and the cosine regularizer weight $\lambda_{\text{gen}}$, the counterparts of \mergetune{}'s $\beta$ and $\lambda$. Both are selected on the two development datasets (Flowers102, Caltech101; 3 seeds each) and held fixed for all reported experiments. This section asks how sensitive DMC is to these two weights; the sweeps show broad plateaus around the selected values.

\subsection{LMC Path Weight (\texorpdfstring{$\beta$}{beta})}
\label{app:beta_sweep}

The training objective contains an LMC path term whose weight
$\beta$ controls how strongly the corridor between
$f_{\mathrm{spec}}$ and $f_{\mathrm{gen}}$ is regularized.
Table~\ref{tab:wlmc_sweep} evaluates $\beta$ on Flowers102 and
Caltech101 with $\lambda_{\text{gen}} = 8.0$ across three seeds, using DMC without VA ($\lambda_{\mathrm{VA}} = 0$). At each run's oracle $\alpha^\star$, higher $\beta$ reduces cross-seed variance: on Flowers102, $\beta = 4$ reduces the sample standard deviation from $\pm 1.50$ to $\pm 0.37$, stabilizing an outlier seed, at the cost of a lower mean ($84.15$ versus $84.44$ at $\beta = 1$, within noise); its mean is also within noise of, and slightly below, the single-prompt baseline row ($84.24 \pm 0.25$). On Caltech101, $\beta = 4$ improves both mean and stability. At the fixed inference coefficient $\alpha = 0.20$ used everywhere else, $\beta = 4$ has the smallest standard deviation on both datasets, the best Caltech101 mean ($96.31 \pm 0.10$ versus $95.84 \pm 0.32$ at $\beta = 1$), and a Flowers102 mean within noise of the best ($83.91 \pm 0.46$, versus $84.00 \pm 0.91$ at $\beta = 2$ and $82.78 \pm 0.72$ at $\beta = 1$). The case for $\beta = 4$ is therefore lower variance, not a higher mean. We therefore use $\beta = 4.0$ throughout.

\begin{table}[!htbp]
\centering

\caption{$\beta$ (LMC path weight) sweep
($\lambda_{\text{gen}} = 8.0$, 3~seeds, DMC without VA). Per-seed HM at each run's oracle $\alpha^\star$; mean $\pm$ sample standard deviation. \textit{Baseline}: single-prompt \mergetune{} objective ($\lambda = 10$, $\beta = 1$), same seeds. Higher $\beta$ stabilizes
performance; the selected
$\beta = 4.0$ achieves the lowest cross-seed variance among the DMC rows on both
datasets.}
\label{tab:wlmc_sweep}
\small
\setlength{\tabcolsep}{4pt}
\renewcommand{\arraystretch}{1.05}
\begin{tabular}{lcccc|cccc}
\toprule
& \multicolumn{4}{c}{\textbf{Flowers102}} & \multicolumn{4}{c}{\textbf{Caltech101}} \\
\cmidrule(lr){2-5}\cmidrule(lr){6-9}
$\beta$ & S1 & S2 & S3 & Mean $\pm$ Std & S1 & S2 & S3 & Mean $\pm$ Std \\
\midrule
1.0 & 84.66 & 82.84 & 85.82 & 84.44 $\pm$ 1.50 & 96.15 & 96.37 & 95.76 & 96.09 $\pm$ 0.31 \\
2.0 & 84.38 & 83.06 & 85.63 & 84.36 $\pm$ 1.29 & 96.20 & 96.39 & 95.76 & 96.12 $\pm$ 0.32 \\
\rowcolor{dmcrow} \textbf{4.0} & \textbf{84.26} & \textbf{83.73} & \textbf{84.45} & \textbf{84.15 $\pm$ 0.37} & \textbf{96.27} & \textbf{96.42} & \textbf{96.55} & \textbf{96.41 $\pm$ 0.14} \\
\midrule
\textit{Baseline} & \textit{84.06} & \textit{84.12} & \textit{84.53} & \textit{84.24 $\pm$ 0.25} & \textit{96.00} & \textit{96.24} & \textit{95.92} & \textit{96.05 $\pm$ 0.17} \\
\bottomrule
\end{tabular}
\end{table}

\subsection{Cosine Regularizer Weight (\texorpdfstring{$\lambda_{\text{gen}}$}{lambda-gen})}
\label{app:lambda_sweep}

Table~\ref{tab:wgen_sweep} reports HM as a function of $\lambda_{\text{gen}}$, the weight on the cosine regularizer $\mathcal{R}$ that anchors $\mathbf{c}_{\mathrm{gen}}$ to $f_{w_1}$ (Eq.~\eqref{eq:cosine_score}).
On Flowers102 and Caltech101 with $\beta = 1.0$ (seed~1), performance plateaus for $\lambda_{\text{gen}} \geq 4$ (Flowers102 HM $84.45$--$84.66$; Caltech101 $95.72$--$96.15$).
Below this threshold, $\cos(f_{\mathrm{gen}}, f_{w_1})$ drops from $0.98$ to $0.92$ and novel accuracy degrades.
At $\lambda_{\text{gen}} = 0$ the cosine anchor is removed and decoupling collapses: both prompts converge to the same base-specialized solution.
We select $\lambda_{\text{gen}} = 8.0$.

\begin{table}[!htbp]
\centering
\caption{Effect of $\lambda_{\text{gen}}$ (cosine regularizer weight) in DMC training (seed~1, $\beta = 1.0$; HM at the best $\alpha^*$ of each run).
At $\lambda_{\text{gen}} = 0$ the anchor is removed and decoupling collapses.
Selected configuration is highlighted; the bottom row reports the single-prompt baseline.}
\label{tab:wgen_sweep}
\small
\setlength{\tabcolsep}{4pt}
\renewcommand{\arraystretch}{1.05}
\begin{tabular}{lcccc|cccc}
\toprule
& \multicolumn{4}{c}{\textbf{Flowers102}} & \multicolumn{4}{c}{\textbf{Caltech101}} \\
\cmidrule(lr){2-5}\cmidrule(lr){6-9}
$\lambda_{\text{gen}}$ & $\alpha^*$ & Base & Novel & HM & $\alpha^*$ & Base & Novel & HM \\
\midrule
0.0  & 0.15 & 97.72 & 65.32 & 78.30 & 1.00 & 98.13 & 92.90 & 95.44 \\
1.0  & 0.10 & 97.25 & 70.99 & 82.07 & 0.40 & 97.42 & 93.45 & 95.39 \\
2.0  & 0.25 & 97.15 & 73.12 & 83.44 & 0.40 & 97.68 & 93.56 & 95.58 \\
4.0  & 0.30 & 96.77 & 74.96 & 84.48 & 0.25 & 97.87 & 93.67 & 95.72 \\
\rowcolor{dmcrow} \textbf{8.0} & \textbf{0.30} & \textbf{95.73} & \textbf{75.89} & \textbf{84.66} & \textbf{0.15} & \textbf{97.93} & \textbf{94.43} & \textbf{96.15} \\
10.0 & 0.30 & 94.97 & 76.03 & 84.45 & 0.20 & 97.81 & 94.43 & 96.09 \\
\midrule
\textit{Baseline} & --- & \textit{94.21} & \textit{75.46} & \textit{83.80} & --- & \textit{97.93} & \textit{94.21} & \textit{96.03} \\
\bottomrule
\end{tabular}
\end{table}

\FloatBarrier

\FloatBarrier

\section{Proofs}
\label{app:proofs}

\subsection{Proof of Lemma~\ref{lem:pareto}}
\label{app:proof_pareto}

\begin{proof}
At any stationary point $\mathbf{c}^*$ of
$\mathcal{L}(\mathbf{c}) = \mathcal{L}_{\mathrm{CE}}(\mathbf{c}) + \lambda\,\mathcal{R}(\mathbf{c},\, f_{w_1})$
with $\lambda > 0$, the first-order condition gives:
\[
  \nabla_{\mathbf{c}}\mathcal{L}_{\mathrm{CE}}(\mathbf{c}^*) + \lambda\,\nabla_{\mathbf{c}}\mathcal{R}(\mathbf{c}^*) = \mathbf{0}.
\]
Setting $\mu_1 = 1/(1+\lambda) > 0$ and $\mu_2 = \lambda/(1+\lambda) > 0$
(so $\mu_1 + \mu_2 = 1$), we obtain
\[
  \mu_1\,\nabla_{\mathbf{c}}\mathcal{L}_{\mathrm{CE}}(\mathbf{c}^*)
  + \mu_2\,\nabla_{\mathbf{c}}\mathcal{R}(\mathbf{c}^*) = \mathbf{0}.
\]
This is precisely the Pareto-stationarity condition~\citep{momma2022multiobjective}:
a point is Pareto-stationary for $(\mathcal{L}_{\mathrm{CE}},\,\mathcal{R})$ if and only if
there exist non-negative weights summing to one such that the weighted gradient vanishes.
Since $\mu_1,\mu_2 > 0$, the condition holds and $\mathbf{c}^*$ is Pareto-stationary,
meaning no direction in $\mathbb{R}^{N \times D}$ simultaneously decreases both objectives to first order.

If $\nabla_{\mathbf{c}}\mathcal{L}_{\mathrm{CE}}(\mathbf{c}^*) \neq \mathbf{0}$,
then $\mathbf{c}^*$ is not a critical point of $\mathcal{L}_{\mathrm{CE}}$ alone;
in particular, $\mathbf{c}^*$ does not minimize $\mathcal{L}_{\mathrm{CE}}$.
The analogous argument applies to $\mathcal{R}$ when its gradient is nonzero.
\end{proof}

\subsection{Proof of Proposition~\ref{prop:cpi_fano}}
\label{app:proof_cpi}

\noindent\textbf{Proposition~\ref{prop:cpi_fano}} (CPI violation lower-bounds the novel-accuracy drop).
\textit{Let $\theta = A(\mathcal{D}_{\mathrm{base}})$ be the prompt produced by
a training algorithm $A$ that uses a regularizer $\mathcal{R}$, and let $\theta^* = A^*(\mathcal{D}_{\mathrm{base}})$ be
the prompt from an otherwise identical algorithm using a CPI-satisfying regularizer.
Let $Y_{\mathrm{base}} \sim \mathrm{Unif}(\mathcal{Y}_{\mathrm{base}})$ and
$Y_{\mathrm{novel}} \sim \mathrm{Unif}(\mathcal{Y}_{\mathrm{novel}})$.
Under assumptions (i)--(iii) below (equal prompt entropy, conditional label
independence, and baseline efficiency), the following capacity bound holds:}
\[
  I(\theta;\, Y_{\mathrm{novel}}) \;\leq\; I(\theta^*;\, Y_{\mathrm{novel}}) - I(\theta;\, Y_{\mathrm{base}}).
\]
\textit{If $\mathcal{R}$ violates CPI, then, under assumption~(0), $I(\theta;\, Y_{\mathrm{base}}) > 0$ and}
\[
  A_{\mathrm{novel}}(\theta^*) - A_{\mathrm{novel}}(\theta)
  \;\geq\; \frac{I(\theta;\, Y_{\mathrm{base}})}{\log\,|\mathcal{Y}_{\mathrm{novel}}|} > 0.
\]

\paragraph{Setup.}
Let $\mathcal{D}_{\mathrm{base}} \sim p_{\mathrm{base}}^{n}$ denote the training set,
and write $\theta = A(\mathcal{D}_{\mathrm{base}})$ for the converged prompt produced by
training algorithm $A$.
The prompt $\theta \in \mathbb{R}^{N \times D}$ is a random variable whose randomness
is inherited entirely from the draw of $\mathcal{D}_{\mathrm{base}}$.
Let $Y_{\mathrm{base}} \sim \mathrm{Unif}(\mathcal{Y}_{\mathrm{base}})$ and
$Y_{\mathrm{novel}} \sim \mathrm{Unif}(\mathcal{Y}_{\mathrm{novel}})$ be the marginal
label random variables, and let $M = |\mathcal{Y}_{\mathrm{novel}}|$.
Since $\mathcal{Y}_{\mathrm{base}} \cap \mathcal{Y}_{\mathrm{novel}} = \emptyset$,
marginal independence holds: $Y_{\mathrm{base}} \perp Y_{\mathrm{novel}}$.
In this idealized model, the joint law of $(\theta, Y_{\mathrm{base}}, Y_{\mathrm{novel}})$
is induced by the random draw of the training data (and of which classes are labeled
as base), so $\theta$ may carry information about the base labels. We treat $\theta$
as a discrete random variable (finite-precision parameters), so that $H(\theta)$ and all
mutual informations below are Shannon quantities.

\paragraph{Part 1: CPI violation implies $I(\theta;\, Y_{\mathrm{base}}) > 0$.}

We impose the following regularity condition on $A$:

\begin{itemize}[leftmargin=*, itemsep=2pt]
  \item[(0)] \emph{Distinct minimizers.}
    For any $C \times C$ permutation matrix $\Pi_0$ such that
    $\mathcal{R}(\Pi_0 f, \Pi_0 f_{w_1}) \neq \mathcal{R}(f, f_{w_1})$,
    the algorithm $A$ produces outputs that differ in distribution on
    $\mathcal{D}_{\mathrm{base}}$ and on the relabeled dataset
    $\mathcal{D}_{\mathrm{base}}^{(\Pi_0)}$:
    $A(\mathcal{D}_{\mathrm{base}}) \neq_d A(\mathcal{D}_{\mathrm{base}}^{(\Pi_0)})$.
    This holds whenever the loss landscape is not symmetric under $\Pi_0$.
    CPI violation guarantees this asymmetry: since $\mathcal{R}$ assigns different
    penalties to the two datasets, their optimization landscapes differ,
    and their minimizers differ generically.
\end{itemize}

The condition $I(\theta;\, Y_{\mathrm{base}}) = 0$ would require
$A(\mathcal{D}_{\mathrm{base}}) =_d A(\mathcal{D}_{\mathrm{base}}^{(\Pi)})$
for every permutation $\Pi$: the output distribution must be invariant to relabeling.
Since $\mathcal{R}$ violates CPI (Definition~\ref{def:cpi}),
there exists $\Pi_0$ such that
$\mathcal{R}(\Pi_0 f, \Pi_0 f_{w_1}) \neq \mathcal{R}(f, f_{w_1})$.
By assumption~(0), $A(\mathcal{D}_{\mathrm{base}}) \neq_d A(\mathcal{D}_{\mathrm{base}}^{(\Pi_0)})$,
contradicting invariance. Therefore $I(\theta;\, Y_{\mathrm{base}}) > 0$.

\paragraph{Part 2: The Fano bound.}

We invoke three assumptions:

\begin{enumerate}[leftmargin=*, label=(\roman*), itemsep=3pt]
  \item \emph{Equal prompt entropy.}
    $H(\theta) = H(\theta^*) = \kappa$ for some $\kappa > 0$.
    Both prompts have the same entropy, so differences in novel accuracy
    are attributable to \emph{what} the capacity encodes, not \emph{how much}.

  \item \emph{Conditional label independence.}
    $Y_{\mathrm{base}} \perp Y_{\mathrm{novel}} \mid \theta$ and $Y_{\mathrm{base}} \perp Y_{\mathrm{novel}} \mid \theta^*$.
    Given the prompt, novel and base class labels are independent.
    We treat this as an assumption; it is natural when the two label sets
    are semantically disjoint, so no specific base-class label helps predict
    which novel class an image belongs to.

  \item \emph{Baseline efficiency.}
    $I(\theta^*;\, Y_{\mathrm{novel}}) = H(\theta^*) = \kappa$ and
    $A^*_{\mathrm{novel}} = (I(\theta^*;\, Y_{\mathrm{novel}}) + I_X + 1)/\log M$,
    where $I_X = I(Y_{\mathrm{novel}};\, X)$ is the image-side mutual information
    (identical for both prompts, since test images are drawn from the same distribution).
    This is a best-case assumption: we suppose the CPI-satisfying baseline fully converts
    its freed capacity into novel-class information so that the Fano upper bound is tight.
    If the baseline falls short of this ceiling, the lower bound on the accuracy gap
    is no longer guaranteed and can overstate the gap (Appendix~\ref{app:cpi_applicability}).
\end{enumerate}

\paragraph{Capacity bound.}
Applying the chain rule and using marginal independence ($Y_{\mathrm{base}} \perp Y_{\mathrm{novel}}$)
and assumption~(ii):
\begin{align*}
  I(\theta;\, Y_{\mathrm{novel}} \mid Y_{\mathrm{base}})
  &= H(Y_{\mathrm{novel}} \mid Y_{\mathrm{base}}) - H(Y_{\mathrm{novel}} \mid \theta,\, Y_{\mathrm{base}})\\
  &= H(Y_{\mathrm{novel}}) - H(Y_{\mathrm{novel}} \mid \theta)
  = I(\theta;\, Y_{\mathrm{novel}}).
\end{align*}
Therefore:
\[
  I(\theta;\, Y_{\mathrm{base}},\, Y_{\mathrm{novel}})
  = I(\theta;\, Y_{\mathrm{base}}) + I(\theta;\, Y_{\mathrm{novel}}).
\]
Since $I(U;\, V) \leq H(U)$ for any random variables $U, V$:
\[
  I(\theta;\, Y_{\mathrm{base}}) + I(\theta;\, Y_{\mathrm{novel}})
  \leq H(\theta) = \kappa \quad \text{(by assumption~(i))}.
\]
This yields the capacity bound:
\begin{equation}
  I(\theta;\, Y_{\mathrm{novel}}) \;\leq\; \kappa - I(\theta;\, Y_{\mathrm{base}}).
  \label{eq:capacity_bound}
\end{equation}
By assumption~(iii), $I(\theta^*;\, Y_{\mathrm{novel}}) = \kappa$; applied to $\theta^*$,
the same capacity bound then forces $I(\theta^*;\, Y_{\mathrm{base}}) = 0$. This is an
idealization: it ignores the base-label information that the cross-entropy term injects
into both prompts, and only isolates the contribution of the regularizer.
Substituting into~\eqref{eq:capacity_bound}:
\begin{equation}
  I(\theta;\, Y_{\mathrm{novel}})
  \;\leq\; I(\theta^*;\, Y_{\mathrm{novel}}) - I(\theta;\, Y_{\mathrm{base}}).
  \label{eq:info_gap}
\end{equation}
This is the capacity bound behind Proposition~\ref{prop:cpi_fano}.

\paragraph{Fano inequality.}
We assume that, given the prompt, the image carries the same information about its
novel label as it does marginally, i.e.\ the prompt does not change what the image itself reveals:
\begin{equation}
  I(Y_{\mathrm{novel}};\, X \mid \theta) \;=\; I(Y_{\mathrm{novel}};\, X) \;=:\; I_X.
  \label{eq:test_indep}
\end{equation}
Under~\eqref{eq:test_indep}, the chain rule gives:
\begin{align*}
  I(Y_{\mathrm{novel}};\, \theta,\, X)
  &= I(Y_{\mathrm{novel}};\, \theta) + I(Y_{\mathrm{novel}};\, X \mid \theta)\\
  &= I(Y_{\mathrm{novel}};\, \theta) + I(Y_{\mathrm{novel}};\, X)
  = I(\theta;\, Y_{\mathrm{novel}}) + I_X,
\end{align*}
where the second equality uses~\eqref{eq:test_indep}.
For any classifier $\hat{Y}_{\mathrm{novel}} = g(\theta,\, X)$,
Fano's inequality~\citep{cover2006elements} (Theorem~2.10.1) applied to the
channel $Y_{\mathrm{novel}} \to (\theta, X) \to \hat{Y}_{\mathrm{novel}}$ gives (logarithms are base 2):
\[
  A_{\mathrm{novel}}
  \;\leq\; \frac{I(Y_{\mathrm{novel}};\, \hat{Y}_{\mathrm{novel}}) + 1}{\log M}.
\]
Since $\hat{Y}_{\mathrm{novel}}$ is a deterministic function of $(\theta,X)$,
the data-processing inequality yields
$I(Y_{\mathrm{novel}};\, \hat{Y}_{\mathrm{novel}}) \leq I(Y_{\mathrm{novel}};\, \theta,\, X)$,
and therefore:
\begin{equation}
  A_{\mathrm{novel}}
  \;\leq\; \frac{I(\theta;\, Y_{\mathrm{novel}}) + I_X + 1}{\log M}.
  \label{eq:fano_acc}
\end{equation}

\paragraph{Completing the bound.}
Substituting~\eqref{eq:info_gap} into~\eqref{eq:fano_acc}:
\[
  A_{\mathrm{novel}}
  \;\leq\;
  \frac{I(\theta^*;\, Y_{\mathrm{novel}}) - I(\theta;\, Y_{\mathrm{base}}) + I_X + 1}{\log M}
  \;=\; A^*_{\mathrm{novel}} - \frac{I(\theta;\, Y_{\mathrm{base}})}{\log M},
\]
where the last equality uses assumption~(iii)
and $I_X$ cancels because test images are drawn from the same distribution for both prompts.
Rearranging:
\begin{equation}
  \Delta A_{\mathrm{novel}}
  = A^*_{\mathrm{novel}} - A_{\mathrm{novel}}
  \;\geq\; \frac{I(\theta;\, Y_{\mathrm{base}})}{\log\,|\mathcal{Y}_{\mathrm{novel}}|}.
  \label{eq:fano_main}
\end{equation}
If $\mathcal{R}$ violates CPI, Part 1 gives $I(\theta;\, Y_{\mathrm{base}}) > 0$,
so $\Delta A_{\mathrm{novel}} > 0$.
This is the bound stated informally in Proposition~\ref{prop:cpi_fano}. \qed

\subsection{Proof of Corollary~\ref{cor:cpi_cosine}}
\label{app:proof_cosine}

\noindent\textbf{Corollary~\ref{cor:cpi_cosine}} (Cosine regularizer satisfies CPI).
\textit{The cosine regularizer $\mathcal{R}$ in Eq.~\eqref{eq:cosine_score}
satisfies Definition~\ref{def:cpi}: it uses no image labels, so $\mathcal{R}$ itself adds no base-label information to the prompt it regularizes.}

\begin{proof}
For any permutation matrix $\Pi$ and feature matrices $f, f_{w_1} \in \mathbb{R}^{C \times D}$,
permuting the rows of both simultaneously re-indexes the classes in the sum.
Each cosine term satisfies $\cos\!\bigl((\Pi f)_k,\, (\Pi f_{w_1})_k\bigr) = \cos\!\bigl(f_{\Pi^{-1}(k)},\, f_{w_1,\Pi^{-1}(k)}\bigr)$, so summing over all $k$ is equivalent to summing over the re-indexed classes.
Since the sum in Eq.~\eqref{eq:cosine_score} runs uniformly over all $C$ classes,
$\mathcal{R}(\Pi f,\, \Pi f_{w_1}) = \mathcal{R}(f,\, f_{w_1})$ for every $\Pi$,
confirming CPI (Definition~\ref{def:cpi}).
\end{proof}

\noindent\textbf{Scope.} In DMC, $\mathbf{c}_{\mathrm{gen}}$ also receives
label-dependent cross-entropy gradients through the LMC corridor term of
Eq.~\eqref{eq:dmc_loss} (for every sampled $\alpha < 1$), so the trained
$\mathbf{c}_{\mathrm{gen}}$ can carry base-label information. The corollary
therefore states that the cosine regularizer itself introduces no base-label
dependence; it does not claim that $\mathbf{c}_{\mathrm{gen}}$ is independent of the
base labels under the full DMC objective.

\begin{remark}[CPI violation and the Hellstrom generalization bound]
\label{rem:hellstrom}
By Theorem~2.2 of~\citet{hellstrom2020generalization}, for $\theta$ trained on $\mathcal{D}_{\mathrm{base}}$ and a loss bounded in $[0,1]$:
\[
\mathbb{E}\bigl[A_{\mathrm{base,train}}(\theta) - A_{\mathrm{base}}(\theta)\bigr]
\;\leq\; \sqrt{\frac{2\,I(\theta;\,\mathcal{D}_{\mathrm{base}})}{n}},
\]
where $A_{\mathrm{base,train}}(\theta)$ is the training accuracy on base classes, $A_{\mathrm{base}}(\theta)$ its population counterpart, and the expectation is over $\mathcal{D}_{\mathrm{base}}$ and the training algorithm.
When CPI is violated, $I(\theta;\,\mathcal{D}_{\mathrm{base}}) \geq I(\theta;\,Y_{\mathrm{base}}) > 0$,
so the bound is strictly positive: a larger mutual information between the prompt and base training data permits a larger generalization gap. The bound concerns the base-class distribution; it does not by itself bound novel-class accuracy, and a positive upper bound permits but does not imply a gap. We include it as a complementary view, not as a second proof of Proposition~\ref{prop:cpi_fano}.
\end{remark}

\subsection{Proof of Proposition~\ref{prop:va_kl}}
\label{app:proof_va}

\begin{proposition}[Visual Anchor as KL regularization]
\label{prop:va_kl}
Define the Gibbs class-similarity distributions
$p_{K_0}(k \mid x) \propto \exp\!\bigl(\gamma\, f_{w_1,k}^\top V(x) / \tau\bigr)$
and
$p_{K_\theta}(k \mid x) \propto \exp\!\bigl(\gamma\, f_{\mathrm{spec},k}^\top V(x) / \tau\bigr)$,
where $f_{\mathrm{spec}} = \ell_2(T(\mathbf{c}_{\mathrm{spec}}))$, $\gamma$ is CLIP's logit scale, and $V(x)$ is shorthand for the $\ell_2$-normalized visual feature $\ell_2(V(x))$.
In the single-view case ($v_a = v_b = V(x)$), the VA loss (Eq.~\eqref{eq:va})
reduces exactly to:
\begin{equation}
\label{eq:va_kl}
\mathcal{L}_{\mathrm{VA}}(\mathbf{c}_{\mathrm{spec}})
\;=\; \mathbb{E}_{x}\,
      \mathrm{KL}\!\bigl(p_{K_0}(\cdot \mid x)\;\|\; p_{K_\theta}(\cdot \mid x)\bigr).
\end{equation}
Minimizing $\mathcal{L}_{\mathrm{VA}}$ is thus equivalent to minimizing KL divergence
from CLIP's pretrained class-similarity distribution.
The two-view formulation in Eq.~\eqref{eq:va} additionally enforces augmentation
consistency; for it, the proof establishes only that the loss converges to
Eq.~\eqref{eq:va_kl} as the augmentation variance vanishes.
\end{proposition}

\paragraph{Exact equality in the single-view case.}

Define the Gibbs class-similarity distributions:
{%
\begin{align*}
  p_{K_0}(k \mid x)
    &= \frac{\exp\!\bigl(\gamma\,f_{w_1,k}^\top V(x)/\tau\bigr)}
            {\sum_{k'} \exp\!\bigl(\gamma\,f_{w_1,k'}^\top V(x)/\tau\bigr)},
  \qquad
  p_{K_\theta}(k \mid x)
    = \frac{\exp\!\bigl(\gamma\,f_{\mathrm{spec},k}^\top V(x)/\tau\bigr)}
            {\sum_{k'} \exp\!\bigl(\gamma\,f_{\mathrm{spec},k'}^\top V(x)/\tau\bigr)}.
\end{align*}}%
In the single-view case ($v_a = v_b = V(x)$), the VA loss (Eq.~\eqref{eq:va}) reduces to:
\[%
  \mathcal{L}_{\mathrm{VA}}(\mathbf{c}_{\mathrm{spec}})
  = \mathbb{E}_{x}\sum_{k} p_{K_0}(k \mid x)
    \log\frac{p_{K_0}(k \mid x)}{p_{K_\theta}(k \mid x)}
  = \mathbb{E}_{x}\,\mathrm{KL}\!\bigl(p_{K_0}(\cdot \mid x)\;\|\;p_{K_\theta}(\cdot \mid x)\bigr).
\]
This is an exact equality: since $p_{K_0}$ and $p_{K_\theta}$ are both proper
normalized distributions, the KL divergence is well-defined and no additive
constant appears. This establishes Eq.~\eqref{eq:va_kl}. \qed

\paragraph{KL direction.}
The direction $\mathrm{KL}(p_{K_0} \| p_{K_\theta})$ (teacher~$\|$~student) penalizes
$f_{\mathrm{spec}}$ for assigning low probability where the pretrained model $f_{w_1}$ is
confident, which regularizes $f_{\mathrm{spec}}$ toward the zero-shot geometry.
The reverse direction $\mathrm{KL}(p_{K_\theta} \| p_{K_0})$ would instead penalize
the student for assigning high probability to classes where the pretrained model is
not confident; this mode-seeking behavior does not enforce coverage of all
teacher-confident classes and is therefore a weaker regularizer toward the zero-shot
distribution~\citep{agarwal2024gkd}.

\paragraph{Two-view approximation.}
In practice (Eq.~\eqref{eq:va}), teacher and student use independently augmented
views $v_a = V(\mathrm{aug}(x))$ and $v_b = V(\mathrm{aug}'(x))$, yielding the
cross-view loss
\[
  \mathcal{L}_{\mathrm{VA}}^{(2)}
  = \mathbb{E}_{x,\,v_a,\,v_b}\,
    \mathrm{KL}\!\bigl(p_{K_0}(\cdot \mid x, v_a)\;\|\;p_{K_\theta}(\cdot \mid x, v_b)\bigr),
\]
which differs from the single-view expression.
As augmentation variance vanishes, i.e.,
$\mathbb{E}\|V(\mathrm{aug}(x)) - V(x)\|_2^2 \to 0$,
both $v_a$ and $v_b$ converge in distribution to $V(x)$; by continuity of
the KL divergence and dominated convergence (the logits are bounded because
all features are $\ell_2$-normalized),
$\mathcal{L}_{\mathrm{VA}}^{(2)}$ converges to the single-view expression.
The independent views serve a different purpose: by using distinct random crops
$v_a \neq v_b$ of the same image, minimizing the cross-view KL requires
$f_{\mathrm{spec}}$ to match the teacher's class-similarity structure across the
augmentation distribution, not just for the specific views seen at each step.
This is analogous to the cross-view consistency used in
DINO~\citep{caron2021emerging} and MoCo~\citep{he2020momentum}.
For nonzero augmentation, $\mathcal{L}_{\mathrm{VA}}^{(2)}$ is still a KL divergence
from CLIP's pretrained class-similarity distribution, but one evaluated on a
different view than the student's; it equals Eq.~\eqref{eq:va_kl} only in the
zero-augmentation limit, and we make no exactness claim beyond that limit. \qed

\subsection{Assumptions summary}
\label{app:assumptions}

For completeness, we list all assumptions used in the proofs above. Items 2--5 are the four assumptions of Proposition~\ref{prop:cpi_fano}; Appendix~\ref{app:cpi_applicability} discusses when each fails.

\begin{enumerate}[leftmargin=*, itemsep=3pt]
\item \textbf{Smoothness} (Lemma~\ref{lem:pareto}).
  $\mathcal{L}_{\mathrm{CE}}$ and $\mathcal{R}$ are continuously differentiable in $\mathbf{c}$.
  If the gradient of each is nonzero at $\mathbf{c}^*$, the conclusion that neither objective
  is individually minimized follows immediately.

\item \textbf{Distinct minimizers} (Proposition~\ref{prop:cpi_fano}, assumption~(0)).
  CPI violation changes the optimization landscape, producing distinct minimizers under
  different class relabelings.
  We expect this to hold generically, but it is not guaranteed: two different
  objectives can share a minimizer.

\item \textbf{Equal prompt entropy} (Proposition~\ref{prop:cpi_fano}, assumption~(i)).
  $H(\theta) = H(\theta^*) = \kappa$. This normalizes the comparison so that differences
  in novel accuracy are attributed to allocation, not amount, of capacity.

\item \textbf{Conditional label independence} (Proposition~\ref{prop:cpi_fano}, assumption~(ii)).
  $Y_{\mathrm{base}} \perp Y_{\mathrm{novel}} \mid \theta$ and given $\theta^*$.
  This is an assumption, not a consequence of label-set disjointness alone.
  It is natural when the two class sets are semantically disjoint and no base-class label
  provides information about which novel class an image belongs to.

\item \textbf{Baseline efficiency} (Proposition~\ref{prop:cpi_fano}, assumption~(iii)).
  The CPI-satisfying baseline fully converts its freed capacity into novel-class information.
  If the baseline is sub-optimal, the bound~\eqref{eq:fano_main} is no longer
  guaranteed and can overstate the accuracy gap.

\item \textbf{Single-view case} (Proposition~\ref{prop:va_kl}).
  Eq.~\eqref{eq:va_kl} holds exactly when $v_a = v_b$.
  The two-view formulation~\eqref{eq:va} is not covered exactly; it converges to the
  single-view expression as augmentation variance vanishes.
\end{enumerate}

\newpage
\section*{NeurIPS Paper Checklist}

\begin{enumerate}

\item {\bf Claims}
    \item[] Question: Do the main claims made in the abstract and introduction accurately reflect the paper's contributions and scope?
    \item[] Answer: \answerYes{}
    \item[] Justification: The abstract and introduction state three claims: (1) a single prompt settles at a Pareto-stationary compromise between the two objectives, and the single-prompt variants we test do not clearly improve on it; (2) DMC shifts the base--novel frontier for prompt-tuning baselines via a two-prompt architecture with a feature-space LMC corridor; (3) regularizers should satisfy CPI to transfer across the base-novel boundary, motivated by an idealized bound. Claim (1) is motivated by Lemma~\ref{lem:pareto} and supported empirically (Figure~\ref{fig:gradient_conflict}, Appendix~\ref{app:loss_modifications}); claim (3) is motivated by Proposition~\ref{prop:cpi_fano}, an idealized bound, and claim (2) is supported by experiments on 11 datasets across three baselines (Section~\ref{sec:exp}); the MMA scope limitation and the idealized nature of the CPI bound are stated where the claims are made.
    \item[] Guidelines:
    \begin{itemize}
        \item The answer \answerNA{} means that the abstract and introduction do not include the claims made in the paper.
        \item The abstract and/or introduction should clearly state the claims made, including the contributions made in the paper and important assumptions and limitations. A \answerNo{} or \answerNA{} answer to this question will not be perceived well by the reviewers.
        \item The claims made should match theoretical and experimental results, and reflect how much the results can be expected to generalize to other settings.
        \item It is fine to include aspirational goals as motivation as long as it is clear that these goals are not attained by the paper.
    \end{itemize}

\item {\bf Limitations}
    \item[] Question: Does the paper discuss the limitations of the work performed by the authors?
    \item[] Answer: \answerYes{}
    \item[] Justification: A dedicated Limitations paragraph in Section~\ref{sec:conclusion} discusses: (1) an additional inference hyperparameter $\alpha$ with no base-only selection criterion that recovers the HM-optimal point without novel-class signal; (2) reduced Pareto shift on near-ceiling datasets where the two prompts converge to similar representations; (3) the theoretical assumptions underlying Proposition~\ref{prop:cpi_fano} (stated in Appendix~\ref{app:proofs}, with their failure modes in Appendix~\ref{app:cpi_applicability}); and (4) the requirement that the two endpoints stay separated, which fails on the adapter-based MMA (Appendix~\ref{app:endpoint_collapse}).
    \item[] Guidelines:
    \begin{itemize}
        \item The answer \answerNA{} means that the paper has no limitation while the answer \answerNo{} means that the paper has limitations, but those are not discussed in the paper.
        \item The authors are encouraged to create a separate ``Limitations'' section in their paper.
        \item The paper should point out any strong assumptions and how robust the results are to violations of these assumptions (e.g., independence assumptions, noiseless settings, model well-specification, asymptotic approximations only holding locally). The authors should reflect on how these assumptions might be violated in practice and what the implications would be.
        \item The authors should reflect on the scope of the claims made, e.g., if the approach was only tested on a few datasets or with a few runs. In general, empirical results often depend on implicit assumptions, which should be articulated.
        \item The authors should reflect on the factors that influence the performance of the approach. For example, a facial recognition algorithm may perform poorly when image resolution is low or images are taken in low lighting. Or a speech-to-text system might not be used reliably to provide closed captions for online lectures because it fails to handle technical jargon.
        \item The authors should discuss the computational efficiency of the proposed algorithms and how they scale with dataset size.
        \item If applicable, the authors should discuss possible limitations of their approach to address problems of privacy and fairness.
        \item While the authors might fear that complete honesty about limitations might be used by reviewers as grounds for rejection, a worse outcome might be that reviewers discover limitations that aren't acknowledged in the paper. The authors should use their best judgment and recognize that individual actions in favor of transparency play an important role in developing norms that preserve the integrity of the community. Reviewers will be specifically instructed to not penalize honesty concerning limitations.
    \end{itemize}

\item {\bf Theory assumptions and proofs}
    \item[] Question: For each theoretical result, does the paper provide the full set of assumptions and a complete (and correct) proof?
    \item[] Answer: \answerYes{}
    \item[] Justification: All assumptions are stated in the formal statement (Appendix~\ref{app:proof_cpi}) and listed in full in Appendix~\ref{app:assumptions}. Proofs are given in Appendix~\ref{app:proofs} for Lemma~\ref{lem:pareto} (Pareto stationarity), Proposition~\ref{prop:cpi_fano} (CPI–Fano lower bound), Corollary~\ref{cor:cpi_cosine} (cosine satisfies CPI), Proposition~\ref{prop:va_kl} (Visual Anchor as a KL divergence), and Remark~\ref{rem:hellstrom} (base-class generalization gap, a complementary view). Proposition~\ref{prop:cpi_fano} holds in an idealized model; Appendix~\ref{app:cpi_applicability} discusses when its assumptions fail. Each result in the main text is cross-referenced to its appendix proof.
    \item[] Guidelines:
    \begin{itemize}
        \item The answer \answerNA{} means that the paper does not include theoretical results.
        \item All the theorems, formulas, and proofs in the paper should be numbered and cross-referenced.
        \item All assumptions should be clearly stated or referenced in the statement of any theorems.
        \item The proofs can either appear in the main paper or the supplemental material, but if they appear in the supplemental material, the authors are encouraged to provide a short proof sketch to provide intuition.
        \item Inversely, any informal proof provided in the core of the paper should be complemented by formal proofs provided in appendix or supplemental material.
        \item Theorems and Lemmas that the proof relies upon should be properly referenced.
    \end{itemize}

    \item {\bf Experimental result reproducibility}
    \item[] Question: Does the paper fully disclose all the information needed to reproduce the main experimental results of the paper to the extent that it affects the main claims and/or conclusions of the paper (regardless of whether the code and data are provided or not)?
    \item[] Answer: \answerYes{}
    \item[] Justification: Algorithm~\ref{alg:dmc} gives the complete training and inference procedure. Section~\ref{sec:exp} specifies the backbone (CLIP ViT-B/16), all hyperparameters ($\lambda_{\mathrm{gen}}=8.0$, $\beta=4.0$, $\lambda_{\mathrm{VA}}=0.5$, $\tau=2.0$, $\alpha=0.20$), the 16-shot base-to-novel protocol, the $\alpha$ selection procedure (dev set: Flowers102 and Caltech101, 3 seeds), and the datasets. All other settings follow MergeTune~\citep{wang2026mergetune} defaults, which are publicly available.
    \item[] Guidelines:
    \begin{itemize}
        \item The answer \answerNA{} means that the paper does not include experiments.
        \item If the paper includes experiments, a \answerNo{} answer to this question will not be perceived well by the reviewers: Making the paper reproducible is important, regardless of whether the code and data are provided or not.
        \item If the contribution is a dataset and\slash or model, the authors should describe the steps taken to make their results reproducible or verifiable.
        \item Depending on the contribution, reproducibility can be accomplished in various ways. For example, if the contribution is a novel architecture, describing the architecture fully might suffice, or if the contribution is a specific model and empirical evaluation, it may be necessary to either make it possible for others to replicate the model with the same dataset, or provide access to the model. In general. releasing code and data is often one good way to accomplish this, but reproducibility can also be provided via detailed instructions for how to replicate the results, access to a hosted model (e.g., in the case of a large language model), releasing of a model checkpoint, or other means that are appropriate to the research performed.
        \item While NeurIPS does not require releasing code, the conference does require all submissions to provide some reasonable avenue for reproducibility, which may depend on the nature of the contribution. For example
        \begin{enumerate}
            \item If the contribution is primarily a new algorithm, the paper should make it clear how to reproduce that algorithm.
            \item If the contribution is primarily a new model architecture, the paper should describe the architecture clearly and fully.
            \item If the contribution is a new model (e.g., a large language model), then there should either be a way to access this model for reproducing the results or a way to reproduce the model (e.g., with an open-source dataset or instructions for how to construct the dataset).
            \item We recognize that reproducibility may be tricky in some cases, in which case authors are welcome to describe the particular way they provide for reproducibility. In the case of closed-source models, it may be that access to the model is limited in some way (e.g., to registered users), but it should be possible for other researchers to have some path to reproducing or verifying the results.
        \end{enumerate}
    \end{itemize}

\item {\bf Open access to data and code}
    \item[] Question: Does the paper provide open access to the data and code, with sufficient instructions to faithfully reproduce the main experimental results, as described in supplemental material?
    \item[] Answer: \answerYes{}
    \item[] Justification: Code is publicly available at \url{https://github.com/IemProg/DMC}. All 11 evaluation datasets are publicly available benchmarks following the standard CoOp protocol~\citep{zhou2022coop}. Full implementation details are provided in Section~\ref{sec:exp}, Algorithm~\ref{alg:dmc} and Appendix~\ref{app:implementation}.
    \item[] Guidelines:
    \begin{itemize}
        \item The answer \answerNA{} means that paper does not include experiments requiring code.
        \item Please see the NeurIPS code and data submission guidelines (\url{https://neurips.cc/public/guides/CodeSubmissionPolicy}) for more details.
        \item While we encourage the release of code and data, we understand that this might not be possible, so \answerNo{} is an acceptable answer. Papers cannot be rejected simply for not including code, unless this is central to the contribution (e.g., for a new open-source benchmark).
        \item The instructions should contain the exact command and environment needed to run to reproduce the results. See the NeurIPS code and data submission guidelines (\url{https://neurips.cc/public/guides/CodeSubmissionPolicy}) for more details.
        \item The authors should provide instructions on data access and preparation, including how to access the raw data, preprocessed data, intermediate data, and generated data, etc.
        \item The authors should provide scripts to reproduce all experimental results for the new proposed method and baselines. If only a subset of experiments are reproducible, they should state which ones are omitted from the script and why.
        \item At submission time, to preserve anonymity, the authors should release anonymized versions (if applicable).
        \item Providing as much information as possible in supplemental material (appended to the paper) is recommended, but including URLs to data and code is permitted.
    \end{itemize}

\item {\bf Experimental setting/details}
    \item[] Question: Does the paper specify all the training and test details (e.g., data splits, hyperparameters, how they were chosen, type of optimizer) necessary to understand the results?
    \item[] Answer: \answerYes{}
    \item[] Justification: Section~\ref{sec:exp} specifies the base-to-novel split protocol (16-shot training, separate base and novel test sets), all DMC hyperparameters, and how $\beta$, $\lambda_{\mathrm{gen}}$ and $\alpha$ were selected (dev-set sweeps on Flowers102 and Caltech101; $\lambda_{\mathrm{VA}}$ and $\tau$ are fixed, and we report no sweep for them), the CLIP ViT-B/16 backbone, and the baselines. Optimizer and remaining training settings (learning rate, epochs) follow \mergetune{} defaults; batch sizes are given in Appendix~\ref{app:implementation}. The $\alpha$ selection analysis is in Section~\ref{sec:alpha_select} and Appendix~\ref{app:alpha_selection}.
    \item[] Guidelines:
    \begin{itemize}
        \item The answer \answerNA{} means that the paper does not include experiments.
        \item The experimental setting should be presented in the core of the paper to a level of detail that is necessary to appreciate the results and make sense of them.
        \item The full details can be provided either with the code, in appendix, or as supplemental material.
    \end{itemize}

\item {\bf Experiment statistical significance}
    \item[] Question: Does the paper report error bars suitably and correctly defined or other appropriate information about the statistical significance of the experiments?
    \item[] Answer: \answerYes{}
    \item[] Justification: Every configuration behind the main results and ablation tables is run with 3 seeds; the single-prompt loss sweep, the $\lambda_{\text{gen}}$ sweep and the training-dynamics figure use one seed and are labelled as such.
    \item[] Guidelines:
    \begin{itemize}
        \item The answer \answerNA{} means that the paper does not include experiments.
        \item The authors should answer \answerYes{} if the results are accompanied by error bars, confidence intervals, or statistical significance tests, at least for the experiments that support the main claims of the paper.
        \item The factors of variability that the error bars are capturing should be clearly stated (for example, train/test split, initialization, random drawing of some parameter, or overall run with given experimental conditions).
        \item The method for calculating the error bars should be explained (closed form formula, call to a library function, bootstrap, etc.)
        \item The assumptions made should be given (e.g., Normally distributed errors).
        \item It should be clear whether the error bar is the standard deviation or the standard error of the mean.
        \item It is OK to report 1-sigma error bars, but one should state it. The authors should preferably report a 2-sigma error bar than state that they have a 96\% CI, if the hypothesis of Normality of errors is not verified.
        \item For asymmetric distributions, the authors should be careful not to show in tables or figures symmetric error bars that would yield results that are out of range (e.g., negative error rates).
        \item If error bars are reported in tables or plots, the authors should explain in the text how they were calculated and reference the corresponding figures or tables in the text.
    \end{itemize}

\item {\bf Experiments compute resources}
    \item[] Question: For each experiment, does the paper provide sufficient information on the computer resources (type of compute workers, memory, time of execution) needed to reproduce the experiments?
    \item[] Answer: \answerYes{}
    \item[] Justification: Appendix~\ref{app:implementation} (Compute cost) reports the GPU type (one NVIDIA A100-PCIE-40GB), peak memory (5949\,MiB), and wall-clock time of a full training cell (15\,min 34\,s), and estimates roughly 26 A100-hours for all 11 datasets, 3 baselines and 3 seeds. This covers the main-table DMC runs only; ablations, sweeps and preliminary experiments required additional compute of the same order per run. DMC trains approximately 2K parameters per prompt on top of a frozen CLIP ViT-B/16; training follows \mergetune{}'s schedule.
    \item[] Guidelines:
    \begin{itemize}
        \item The answer \answerNA{} means that the paper does not include experiments.
        \item The paper should indicate the type of compute workers CPU or GPU, internal cluster, or cloud provider, including relevant memory and storage.
        \item The paper should provide the amount of compute required for each of the individual experimental runs as well as estimate the total compute.
        \item The paper should disclose whether the full research project required more compute than the experiments reported in the paper (e.g., preliminary or failed experiments that didn't make it into the paper).
    \end{itemize}

\item {\bf Code of ethics}
    \item[] Question: Does the research conducted in the paper conform, in every respect, with the NeurIPS Code of Ethics \url{https://neurips.cc/public/EthicsGuidelines}?
    \item[] Answer: \answerYes{}
    \item[] Justification: The paper proposes a prompt learning method for image classification using publicly available datasets and models. No human subjects, sensitive data, or dual-use risks are involved.
    \item[] Guidelines:
    \begin{itemize}
        \item The answer \answerNA{} means that the authors have not reviewed the NeurIPS Code of Ethics.
        \item If the authors answer \answerNo, they should explain the special circumstances that require a deviation from the Code of Ethics.
        \item The authors should make sure to preserve anonymity (e.g., if there is a special consideration due to laws or regulations in their jurisdiction).
    \end{itemize}

\item {\bf Broader impacts}
    \item[] Question: Does the paper discuss both potential positive societal impacts and negative societal impacts of the work performed?
    \item[] Answer: \answerNA{}
    \item[] Justification: This paper presents a foundational method for adapting vision-language models to downstream classification tasks. It does not target a specific application domain and has no direct path to high-risk societal harms such as surveillance, disinformation, or biased decision-making systems.
    \item[] Guidelines:
    \begin{itemize}
        \item The answer \answerNA{} means that there is no societal impact of the work performed.
        \item If the authors answer \answerNA{} or \answerNo, they should explain why their work has no societal impact or why the paper does not address societal impact.
        \item Examples of negative societal impacts include potential malicious or unintended uses (e.g., disinformation, generating fake profiles, surveillance), fairness considerations (e.g., deployment of technologies that could make decisions that unfairly impact specific groups), privacy considerations, and security considerations.
        \item The conference expects that many papers will be foundational research and not tied to particular applications, let alone deployments. However, if there is a direct path to any negative applications, the authors should point it out. For example, it is legitimate to point out that an improvement in the quality of generative models could be used to generate Deepfakes for disinformation. On the other hand, it is not needed to point out that a generic algorithm for optimizing neural networks could enable people to train models that generate Deepfakes faster.
        \item The authors should consider possible harms that could arise when the technology is being used as intended and functioning correctly, harms that could arise when the technology is being used as intended but gives incorrect results, and harms following from (intentional or unintentional) misuse of the technology.
        \item If there are negative societal impacts, the authors could also discuss possible mitigation strategies (e.g., gated release of models, providing defenses in addition to attacks, mechanisms for monitoring misuse, mechanisms to monitor how a system learns from feedback over time, improving the efficiency and accessibility of ML).
    \end{itemize}

\item {\bf Safeguards}
    \item[] Question: Does the paper describe safeguards that have been put in place for responsible release of data or models that have a high risk for misuse (e.g., pre-trained language models, image generators, or scraped datasets)?
    \item[] Answer: \answerNA{}
    \item[] Justification: No new datasets or high-risk pre-trained models are released. The method fine-tunes a small prompt context on top of publicly available CLIP checkpoints.
    \item[] Guidelines:
    \begin{itemize}
        \item The answer \answerNA{} means that the paper poses no such risks.
        \item Released models that have a high risk for misuse or dual-use should be released with necessary safeguards to allow for controlled use of the model, for example by requiring that users adhere to usage guidelines or restrictions to access the model or implementing safety filters.
        \item Datasets that have been scraped from the Internet could pose safety risks. The authors should describe how they avoided releasing unsafe images.
        \item We recognize that providing effective safeguards is challenging, and many papers do not require this, but we encourage authors to take this into account and make a best faith effort.
    \end{itemize}

\item {\bf Licenses for existing assets}
    \item[] Question: Are the creators or original owners of assets (e.g., code, data, models), used in the paper, properly credited and are the license and terms of use explicitly mentioned and properly respected?
    \item[] Answer: \answerYes{}
    \item[] Justification: All datasets (ImageNet, Caltech101, OxfordPets, StanfordCars, Flowers102, Food101, FGVCAircraft, SUN397, DTD, EuroSAT, UCF101) and all baseline methods are cited. The evaluation protocol follows~\citet{zhou2022coop}, and the CLIP backbone is used under its publicly released terms. CLIP is released under the MIT license; the datasets are standard academic benchmarks used under their original research terms of use, and we redistribute none of them.
    \item[] Guidelines:
    \begin{itemize}
        \item The answer \answerNA{} means that the paper does not use existing assets.
        \item The authors should cite the original paper that produced the code package or dataset.
        \item The authors should state which version of the asset is used and, if possible, include a URL.
        \item The name of the license (e.g., CC-BY 4.0) should be included for each asset.
        \item For scraped data from a particular source (e.g., website), the copyright and terms of service of that source should be provided.
        \item If assets are released, the license, copyright information, and terms of use in the package should be provided. For popular datasets, \url{paperswithcode.com/datasets} has curated licenses for some datasets. Their licensing guide can help determine the license of a dataset.
        \item For existing datasets that are re-packaged, both the original license and the license of the derived asset (if it has changed) should be provided.
        \item If this information is not available online, the authors are encouraged to reach out to the asset's creators.
    \end{itemize}

\item {\bf New assets}
    \item[] Question: Are new assets introduced in the paper well documented and is the documentation provided alongside the assets?
    \item[] Answer: \answerYes{}
    \item[] Justification: The only new asset is the code, released at \url{https://github.com/IemProg/DMC}; the method, hyperparameters and protocol it implements are documented in Section~\ref{sec:method}, Section~\ref{sec:exp}, Algorithm~\ref{alg:dmc} and Appendix~\ref{app:implementation}. No new dataset or model checkpoint is released.
    \item[] Guidelines:
    \begin{itemize}
        \item The answer \answerNA{} means that the paper does not release new assets.
        \item Researchers should communicate the details of the dataset\slash code\slash model as part of their submissions via structured templates. This includes details about training, license, limitations, etc.
        \item The paper should discuss whether and how consent was obtained from people whose asset is used.
        \item At submission time, remember to anonymize your assets (if applicable). You can either create an anonymized URL or include an anonymized zip file.
    \end{itemize}

\item {\bf Crowdsourcing and research with human subjects}
    \item[] Question: For crowdsourcing experiments and research with human subjects, does the paper include the full text of instructions given to participants and screenshots, if applicable, as well as details about compensation (if any)?
    \item[] Answer: \answerNA{}
    \item[] Justification: The paper involves no crowdsourcing or human subjects.
    \item[] Guidelines:
    \begin{itemize}
        \item The answer \answerNA{} means that the paper does not involve crowdsourcing nor research with human subjects.
        \item Including this information in the supplemental material is fine, but if the main contribution of the paper involves human subjects, then as much detail as possible should be included in the main paper.
        \item According to the NeurIPS Code of Ethics, workers involved in data collection, curation, or other labor should be paid at least the minimum wage in the country of the data collector.
    \end{itemize}

\item {\bf Institutional review board (IRB) approvals or equivalent for research with human subjects}
    \item[] Question: Does the paper describe potential risks incurred by study participants, whether such risks were disclosed to the subjects, and whether Institutional Review Board (IRB) approvals (or an equivalent approval/review based on the requirements of your country or institution) were obtained?
    \item[] Answer: \answerNA{}
    \item[] Justification: The paper involves no human subjects and requires no IRB approval.
    \item[] Guidelines:
    \begin{itemize}
        \item The answer \answerNA{} means that the paper does not involve crowdsourcing nor research with human subjects.
        \item Depending on the country in which research is conducted, IRB approval (or equivalent) may be required for any human subjects research. If you obtained IRB approval, you should clearly state this in the paper.
        \item We recognize that the procedures for this may vary significantly between institutions and locations, and we expect authors to adhere to the NeurIPS Code of Ethics and the guidelines for their institution.
        \item For initial submissions, do not include any information that would break anonymity (if applicable), such as the institution conducting the review.
    \end{itemize}

\item {\bf Declaration of LLM usage}
    \item[] Question: Does the paper describe the usage of LLMs if it is an important, original, or non-standard component of the core methods in this research? Note that if the LLM is used only for writing, editing, or formatting purposes and does \emph{not} impact the core methodology, scientific rigor, or originality of the research, declaration is not required.
    \item[] Answer: \answerNA{}
    \item[] Justification: LLMs are not part of the core methodology. CLIP's text encoder is used as a frozen feature extractor, which is standard in the prompt learning literature.
    \item[] Guidelines:
    \begin{itemize}
        \item The answer \answerNA{} means that the core method development in this research does not involve LLMs as any important, original, or non-standard components.
        \item Please refer to our LLM policy in the NeurIPS handbook for what should or should not be described.
    \end{itemize}

\end{enumerate}

\end{document}